\documentclass[11pt]{article}

\usepackage{theapa}
\usepackage{jair}
\usepackage{times}
\usepackage{amssymb}
\usepackage{curves}
\usepackage{graphicx}

\input epsf %% at top of file
\newcommand{\scalefig}[3]{%%  scale, figure, caption
\begin{figure}[htb]
\begin{center}
  \makebox{{\def\epsfsize##1##2{#1##1}\epsfbox{#2}}}
\end{center}
\vspace*{-0.3cm}
\caption{\label{#2}#3}
\end{figure}}

\newcommand{\scaleonlyfig}[2]{%%  scale, figure, caption
  \makebox{{\def\epsfsize##1##2{#1##1}\epsfbox{#2}}}
}

{\hfill \rule[0.3ex]{1ex}{1ex} \par \addvspace{\bigskipamount}}
{\hfill \rule[0.3ex]{1ex}{1ex} \par \addvspace{\bigskipamount}}

\newcommand{\np}{NP}

\newcommand{\pspace}{PSPACE}

\newcommand{\commentout}[1]{}

\jairheading{26}{2006}{453-541}{12/05}{08/06}
\ShortHeadings{Engineering Benchmarks for Planning}
{Hoffmann, Edelkamp, Thi\'ebaux, Englert, Liporace
  \& Tr\"ug}
\firstpageno{453}

\begin{document}

\title{Engineering Benchmarks for Planning: the Domains Used in the
  Deterministic Part of IPC-4}

\author{\name J\"org Hoffmann
  \email hoffmann@mpi-sb.mpg.de\\
  \addr Max Planck Institute for Computer Science,\\
  Saarbr\"ucken, Germany\\
\name Stefan Edelkamp
  \email stefan.edelkamp@cs.uni-dortmund.de\\
  \addr Fachbereich Informatik,\\
  Universit\"at Dortmund, Germany\\
  \name Sylvie Thi\'ebaux
%Sylvie
%  \email thiebaux@cslws1.anu.edu.au\\
  \email Sylvie.Thiebaux@anu.edu.au\\
  \addr National ICT Australia \& Computer Sciences Laboratory,\\
  The Australian National University, Canberra, Australia\\
  \name Roman Englert
  \email Roman.Englert@telekom.de\\
  \addr Deutsche Telekom Laboratories,\\
  Berlin, Germany\\
  \name Frederico dos Santos Liporace
  \email liporace@inf.puc-rio.br\\
  \addr Departamento de Inform\'atica, PUC-Rio,\\
  Rio de Janeiro, Brazil\\
  \name Sebastian Tr\"ug
  \email trueg@informatik.uni-freiburg.de\\
  \addr Institut f\"ur Informatik,\\
  Universit\"at Freiburg, Germany}

\maketitle

\begin{abstract}
  In a field of research about general reasoning mechanisms, it is
  essential to have appropriate benchmarks. Ideally, the benchmarks
  should reflect possible applications of the developed technology.
  In AI Planning, researchers more and more tend to draw their testing
  examples from the benchmark collections used in the International
  Planning Competition (IPC). In the organization of (the
  deterministic part of) the fourth IPC, {\em IPC-4}, the authors
  therefore invested significant effort to create a useful set of
  benchmarks. They come from five different (potential) real-world
  applications of planning: airport ground traffic control, oil
  derivative transportation in pipeline networks, model-checking
  safety properties, power supply restoration, and UMTS call setup.
  Adapting and preparing such an application for use as a benchmark in
  the IPC involves, at the time, inevitable (often drastic)
  simplifications, as well as careful choice between, and engineering
  of, domain encodings. For the first time in the IPC, we used
  compilations to formulate complex domain features in simple
  languages such as STRIPS, rather than just dropping the more
  interesting problem constraints in the simpler language subsets. The
  article explains and discusses the five application domains and
  their adaptation to form the PDDL test suites used in IPC-4. We
  summarize known theoretical results on structural properties of the
  domains, regarding their computational complexity and provable
  properties of their topology under the $h^+$ function (an idealized
  version of the relaxed plan heuristic). We present new (empirical)
  results illuminating properties such as the quality of the most
  wide-spread heuristic functions (planning graph, serial planning
  graph, and relaxed plan), the growth of propositional
  representations over instance size, and the number of actions
  available to achieve each fact; we discuss these data in conjunction
  with the best results achieved by the different kinds of planners
  participating in IPC-4.
\end{abstract}

\section{Introduction}
\label{introduction}

Today, to a large extent the research discipline of AI planning is
concerned with improving the performance of domain independent
generative planning systems. A domain independent generative planning
system ({\em planner}) must be able to fully automatically find {\em
  plans}: solution sequences in declaratively specified transition
systems.  The simplest planning formalism is {\em deterministic
  planning}.  There, a planner is given as input a set of state
variables (often just Booleans, called {\em facts}), an {\em initial
  state} (a value assignment to the variables), a {\em goal} (a
formula), and a set of {\em actions} (with a precondition formula
describing applicability, and with an effect specifying how the action
changes the state). A plan is a time-stamped sequence of actions that
maps the initial state into a state that satisfies the goal. This sort
of formalism is called deterministic since the initial state is fully
specified and the effects of the actions are non-ambiguous. Both
restrictions may be weakened to obtain {\em non-deterministic} and
{\em probabilistic} planning.

Performance of planners is measured by testing them on {\em benchmark}
example instances of the planning problem. The ``best'' algorithm at
any point in time is, generally, considered to be the one that solves
these examples most efficiently.  In particular, this is the idea in
the International Planning Competition ({\em IPC}), a biennial event
aimed at showcasing the capabilities of current planning systems.

The first IPC took place in 1998, so at the time of writing there were
four such events. Providing details about the IPC is beyond the scope
of this paper, and we refer the reader to the overview articles
written by the organizers of the respective IPC editions
\cite{mcdermott:aim-00,bacchus:aim-01,long:fox:jair-03,hoffmann:edelkamp:jair-05}.
In particular, \citeA{hoffmann:edelkamp:jair-05} provide details about
the 4th IPC, such as overall organization, different tracks,
evaluation, participating planners, and results. Basic information is
included in this paper, so the reader should be able to follow the
main discussion without a detailed background. The language used to
describe planning problems in the IPC is called {\em PDDL}: Planning
Domain Definition Language. It was introduced by \citeA{pddl-handbook}
for the first IPC, IPC-1, in 1998. A subset of the language was
selected by \citeA{pddl-2000-subset} for IPC-2 in 2000. The language
was extended with temporal and numerical constructs by
\citeA{fox:long:jair-03} to form the language {\em PDDL2.1} for IPC-3
in 2002. It was further extended with two additional constructs,
``timed initial literals'' and ``derived predicates'', by
\citeA{hoffmann:edelkamp:jair-05} to form the language {\em PDDL2.2}
for IPC-4 in 2004.

Since, even in its simplest forms, AI planning is a computationally
hard problem, no system can work efficiently in {\em all} problem
instances \cite{bylander:ai-94,helmert:ai-03}.  Thus, it is of crucial
importance {\em what kinds of examples are used for testing}.  Today,
more and more, AI Planning researchers draw their testing examples
from the collections used in the IPC. This makes the IPC benchmarks a
very important instrument for the field. In the organization of the
deterministic part of the 4th IPC (there was also a probabilistic
part, see \citeR{ProbabilisticPart}), the authors therefore invested
considerable effort into creating a set of ``useful'' benchmarks for
planning.

The very first question to answer is what precisely is meant here by
the word ``useful''. This is not an easy question. There is no widely
accepted mathematical definition for deciding whether a set of
benchmarks should be considered useful. There are, however, widely
accepted {\em intuitions} of when this is the case. Benchmarks should
be:

\begin{enumerate}
\item {\bf Oriented at applications} -- a benchmark should reflect an
  application of the technology developed in the field.
\item {\bf Diverse in structure} -- a set of benchmarks should cover
  different kinds of structure, rather than re-state very similar
  tasks.
\end{enumerate}

The first of these is usually considered particularly important --
indeed, AI planning has frequently been criticized for its ``obsession
with toy examples''. In recent years, the performance of
state-of-the-art systems has improved dramatically, and with that more
realistic examples have come within reach. We made another step in
this direction by orienting most of the IPC-4 benchmarks at
application domains. While traditionally planning benchmarks were more
or less fantasy products created having some ``real'' scenario in
mind,\footnote{Of course, there are exceptions to this rule. One
  important one, in our context here, is the Satellite domain, used in
  IPC-3, that we further refined for use in IPC-4. More on this
  later.} we took actual (possible) applications of planning
technology, and turned them into something suitable for the
competition. We considered five different application domains: airport
ground traffic control ({\em Airport}), oil derivative transportation
in pipeline networks ({\em Pipesworld}), model checking safety
properties ({\em Promela}), power supply restoration ({\em PSR}), and
setup of mobile communication in UMTS ({\em UMTS}). Of course, in the
adaptation of an application for use in the IPC, simplifications need
to be made. We will get back to this below.

Diverse structure of benchmarks has traditionally been given less
attention than realism, but we believe that it is no less important.
The structure underlying a testing example determines the performance
of the applied solving mechanism. This is particularly true for
solving mechanisms whose performance rises and falls with the quality
of a heuristic they use.  Hoffmann's
\citeyear{hoffmann:ijcai-01,hoffmann:aips-02,hoffmann:jair-05} results
suggest that much of the spectacular performance of modern heuristic
search planners is due to structural similarities between most of the
traditional planning benchmarks. While this does not imply that modern
heuristic search planners aren't useful, it certainly shows that in
the creation of benchmarks there is a risk of introducing a bias
towards one specific way of solving them. In selecting the benchmark
domains for IPC-4, we tried to cover a range of intuitively very
different kinds of problem structure. We will get back to this below.

On the one hand, a creator of planning benchmarks has the noble goal
of realistic, and structurally diverse, benchmark domains. On the
other hand, he/she has the more pragmatic goal to come up with a
version/representation of the benchmarks that can be attacked with
existing planning systems.  Given the still quite restricted
capabilities of systems, obviously the two goals are in conflict. To
make matters worse, there isn't an arbitrarily large supply of
planning applications that are publicly available, and/or whose
developers agree to have their application used as the basis of a
benchmark. For the IPC organizer, on top of all this, the final
benchmarks must be {\em accessible} for a large enough number of
competing systems, which means they must be formulated in a language
understood by those systems. Further, the benchmarks must show
differences between the scalability of planners, i.e., they must not
be too easy or too hard, thus straddling the boundary of current
system capabilities.

The solution to the above difficulties, at least our solution in the
organization of IPC-4, involved a slow tedious interleaved process of
contacting application developers, choosing domains, exploring domain
versions, and engineering domain version representations. This article
presents, motivates, and discusses our choice of benchmark domains for
IPC-4; it explains the engineering processes that led to the finally
used domain versions and instances. Further, we report about, and
present some new data determining certain structural properties of the
resulting benchmarks (more details below). The main contribution of
the {\em work} is the set of benchmarks, provided in
IPC-4.\footnote{The benchmarks can be downloaded from the IPC-4 web
  page at http://ipc.icaps-conference.org/} The contributions of this
{\em article} are: first, providing the necessary documentation of
these benchmarks; second, describing the technical processes used in
their creation; third, providing an extensive discussion of the
structural properties of the benchmarks. Apart from these more
technical contributions, we believe that our work has value as an
example of a large-scale attempt at engineering a useful set of
benchmarks for classical planning.

It is difficult to make any formal claim about our created set of
benchmarks, such as that they are in some way better than the previous
benchmarks. When working on this, our intent was to overcome certain
shortcomings of many benchmarks, though one would be hard pressed to
come up with a formal proof that such improvements were indeed made.
After all, judging the quality of a set of benchmarks is a rather
complex matter guided mostly by intuitions, and, worse, personal
opinions.\footnote{Consider for example the Movie domain used in
  IPC-1. All instances of this domain, no matter what their size is,
  share the same space of reachable states; the only thing that
  increases is the connectivity between states, i.e. the number of
  actions that have the same effect. Still one can argue that Movie is
  a useful benchmark, in the sense that it can highlight
  systems/approaches that have/have no difficulties in attacking such
  problem characteristics.} What we did was, do our best to create as
realistic, structurally diverse, and accessible benchmarks as possible
for IPC-4. Our belief is that we succeeded in doing so. The benchmarks
definitely differ in certain ways from most of the previous
benchmarks. We think that most of these differences are advantageous;
we will discuss this at the places where we point out the differences.

Regarding realism of the benchmarks, as pointed out above, the main
step we took was to design benchmarks ``top-down'', i.e., start from
actual possible applications of planning technology, and turn them
into something suitable for the competition -- rather than the more
traditional ``bottom-up'' approach of just artificially creating a
domain with some ``real'' scenario in mind. Of course, for modelling
an application in PDDL, particularly for modelling it in a way making
it suitable for use in the IPC, simplifications need to be made. In
some cases, e.g., airport ground traffic control, the simplifications
were not overly drastic, and preserved the overall properties and
intuitive structure of the domain. But in other cases, e.g., oil
derivative transportation in pipeline networks, the simplifications we
needed to make were so drastic that these domains could just as well
have been created in the traditional bottom-up way.  Still, even if
greatly simplified, a domain generated top-down has a better chance to
capture some structure relevant in a real application. Moreover, a
top-down domain has the advantage that since it is derived from a real
application, it provides a clear guideline towards more realism; the
future challenge is to make planners work on more realistic encodings
of the application. In the previous competitions, the only domains
generated top-down in the above sense were the Elevator domain used in
IPC-2 \cite{koehler:schuster:aips-00,bacchus:aim-01}, and the
Satellite and Rovers domains used in IPC-3 \cite{long:fox:jair-03}.

Regarding diverse structure of the benchmarks, in contrast to the
previous competitions, in the IPC-4 domains there is no common
``theme'' underlying many of the benchmarks. In IPC-1, 5 out of 7
domains were variants of transportation; in IPC-2, 4 out of 7 domains
were variants of transportation; in IPC-3, 3 out of 6 domains were
variants of transportation, and 2 were about gathering data in space.
Some of the ``variants'' are in fact very interesting in their use of
constructs such as locked locations, fuel units, road map graphs,
stackable objects, and complex side constraints. However, there is
certainly an intuitive similarity in the structure and relationships
in the domains. To some extent this similarity is even automatically
detectable \cite{long:fox:aips-00}. Not so in IPC-4: airport ground
traffic control, oil derivative transportation in pipeline networks,
model checking safety properties, power supply restoration, and UMTS
call setup are rather different topics. At most one could claim that
airport ground traffic control and UMTS call setup both have a
scheduling nature. We will see, however, that the IPC-4 version of
airport ground traffic control allows considerably more freedom than
classical scheduling formulations, making it a {\bf PSPACE}-complete
decision problem.  The particulars of the domains will be overviewed
in Section~\ref{overview}.

Approaching ``structure'' from a more formal point of view is more
difficult. It is largely unclear what, precisely, the relevant
structure in a planning domain/instance is, in a general sense.  While
\citeA{hoffmann:ijcai-01,hoffmann:aips-02,hoffmann:jair-05} provides
one possible definition -- search space surface topology under a
certain heuristic function -- there are many other possible options.
In particular, Hoffmann's results are relevant only for heuristic
search planners that generate their heuristic functions based on the
``ignoring delete lists'' relaxation
\cite{mcdermott:aips-96,mcdermott:ai-99,bonet:etal:aaai-97,bonet:geffner:ai-01,hoffmann:nebel:jair-01}.
For lack of a better formal handle, we used Hoffmann's definitions to
qualify the structure of the domains.  The selected domains cover
different regions of Hoffmann's ``planning domain taxonomy'', in
particular they lie in regions that have less coverage in the
traditional benchmarks. Because they are interesting in the context of
the paper at hand, we summarize Hoffmann's \citeyear{hoffmann:jair-05}
results for 30 domains including all domains used in the previous
competitions. We also summarize Helmert's \citeyear{helmert:icaps-06}
results on the computational complexity of satisficing and optimal
planning in the IPC-4 domains.  It turns out that their complexity
covers a wide range -- the widest possible range, for propositional
planning formalisms -- from PSPACE-hard to polynomial. We finally
provide some new data to analyze the structural relationships and
differences between the domains.  Amongst other things, for each
instance, we measure: the number of (parallel and sequential) steps
needed to achieve the goal, estimated by the smallest plan found by
any IPC-4 participant; the same number as estimated by planning graphs
and relaxed plans; and the distribution of the number of possible
achieving actions for each fact. The results are examined in a
comparison between the different domains, taking into account the
runtime performance exhibited by the different kinds of planners in
IPC-4.

Apart from realism and diverse structure, our main quest in the
creation of the IPC-4 benchmarks was to promote their accessibility.
Applications are, typically, if they can be modelled at all in PDDL,
most naturally modelled using rather complex language constructs such
as time, numeric variables, logical formulas, and conditional effects.
Most existing systems handle only subsets of this, in fact more than
half of the systems entered into IPC-4 (precisely, 11 out of 19) could
handle only the simple STRIPS language, or slight extensions of
it.\footnote{STRIPS (Stanford Research Institute Problem Solver) is
  the name of the simplest and at the same time most wide-spread
  planning language. In the form of the language used today, the state
  variables are all Boolean, formulas are conjunctions of positive
  atoms, action effects are either atomic positive (make a fact
  true/{\em add} it) or atomic negative (make a fact false/{\em
    delete} it) \cite{fikes:nilsson:ai-71}. The languages selected for
  IPC-2 \cite{pddl-2000-subset}, from which PDDL2.1 and PDDL2.2 are
  derived, were STRIPS and ADL. ADL is a prominent, more expressive,
  alternative to STRIPS, extending it with arbitrary first-order
  formulas as preconditions and goal, and with conditional effects,
  i.e., effects that occur only if their individual effect condition
  (a first-order formula) is met in the state of execution
  \cite{pednault:kr-89}.} In the previous competitions, as done for
example in the Elevator, Satellite, and Rovers domains, this was
handled simply by dropping the more interesting domain constraints in
the simpler languages, i.e., by removing the respective language
constructs from the domain/instance descriptions. In contrast, for the
first time in the IPC, we {\em compiled} as much of the domain
semantics as possible down into the simpler language formats. Such a
compilation is hard, sometimes impossible, to do. It {\em can} be done
for ADL constructs, as well as for the two new constructs introduced
for the IPC-4 language PDDL2.2, derived predicates and timed initial
literals.  We implemented, and applied, compilation methods for all
these cases.  The compilations serve to preserve more of the original
domain structure, in the simpler language classes. For example, the
STRIPS version of the Elevator domain in IPC-2 is so simplified from
the original ADL version that it bears only marginal similarity to
real elevator control -- in particular, the planner can explicitly
{\em tell} passengers to get into or out of the lift.\footnote{The
  passengers won't get in (out) at floors other than their origin
  (destination); however, with explicit control, the planner can
  choose to {\em not let} someone in (out). The more accurate encoding
  is via conditional effects of the action stopping the lift at a
  floor.} In contrast, our STRIPS formulation of the airport ground
traffic domain is, semantically, identical to our ADL formulation --
it expresses the same things, but in a more awkward fashion.

The compiled domain ``versions'' were offered to the competitors as
alternative domain version ``formulations'', yielding a 2-step
hierarchy for each domain. That is, each domain in IPC-4 could contain
several different domain versions, differing in terms of the number of
domain constraints/properties considered. Within each domain version,
there could be several domain version formulations, differing in terms
of the language used to formulate the (same) semantics.  The
competitors could choose, within each version, whichever formulation
their planners could handle best/handle at all, and the results within
the domain version were then evaluated together. This way, we intended
to make the competition as accessible as possible while at the same
time keeping the number of separation lines in the data -- the number
of distinctions that need to be made when evaluating the data -- at an
acceptable level.

We are, of course, aware that encoding details can have a significant
impact on system performance.\footnote{A very detailed account of such
  matters is provided by \citeA{howe:dahlman:jair-02}.} Particularly,
when compiling ADL to STRIPS, in most cases we had to revert to fully
grounded encodings.  While this certainly isn't desirable, we believe
it to be an acceptable price to pay for the benefit of accessibility.
Most current systems ground the operators out as a pre-process anyway.
In cases where we considered the compiled domain formulations too
different from the original ones to allow for a fair comparison --
typically because plan length increased significantly due to the
compilation -- the compiled formulation was posed to the competitors
as a separate domain version.

The article is organized as follows. The main body of text contains
general information. In Section~\ref{compilations}, we give a detailed
explanation of the compilation methods we used. In
Section~\ref{overview}, we give a summary of the domains, each with a
short application description, our motivation for including the
domain, a brief explanation of the main simplifications made, and a
brief explanation of the different domain versions and formulations.
In Section~\ref{known}, we summarize Hoffmann's
\citeyear{hoffmann:jair-05} and Helmert's \citeyear{helmert:icaps-06}
theoretical results on the structure of the IPC-4 domains.
Section~\ref{new}, we provide our own empirical analysis of structural
properties. Section~\ref{conclusion} discusses what was achieved, and
provides a summary of the main issues left open. For each of the IPC-4
domains, we include a separate section in Appendix~\ref{domains},
providing detailed information on the application, its adaptation for
IPC-4, its domain versions, the example instances used, and future
directions. Although these details are in an appendix, we emphasize
that this is not because they are of secondary importance. On the
contrary, they describe the main body of work we did. The presentation
in an appendix seems more suitable since we expect the reader to,
typically, examine the domains in detail in a selective and
non-chronological manner.

%%% Local Variables: 
%%% mode: latex
%%% TeX-master: t
%%% End: 

\section{PDDL Compilations}
\label{compilations}

We used three kinds of compilation methods:
\begin{itemize}
\item ADL to SIMPLE-ADL (STRIPS with conditional effects) or STRIPS;
\item PDDL with derived predicates to PDDL without them;
\item PDDL with timed initial literals to PDDL without them. 
\end{itemize}

We consider these compilation methods in this order, explaining, for
each, how the compilation works, what the main difficulties and their
possible solutions are, and giving an outline of how we used the
compilation in the competition. Note that ADL, SIMPLE-ADL, and STRIPS
are subsets of PDDL. Each of the compilation methods was published
elsewhere already (see the citations in the text). This section serves
as an overview article, since a coherent summary of the techniques,
and their behavior in practice, has not appeared elsewhere in the
literature.

\subsection{Compilations of ADL to SIMPLE-ADL and STRIPS}
\label{compilations:adl2strips}

ADL constructs can be compiled away with methods first proposed by
\citeA{gazen:knoblock:ecp-97}. Suppose we are given a planning
instance with constant (object) set $C$, initial state $I$, goal $G$,
and operator set $O$. Each operator $o$ has a precondition $pre(o)$,
and conditional effects $e$, taking the form $con(e)$, $add(e)$,
$del(e)$ where $add(e)$ and $del(e)$ are lists of atoms.
Preconditions, effect conditions, and $G$ are first order logic
formulas (effect conditions are $TRUE$ for unconditional effects).
Since the domain of discourse -- the set of constants -- is finite,
the formulas can be equivalently transformed into propositional logic.
\begin{enumerate}
\item[(1)] Quantifiers are turned into conjunctions and disjunctions,
  simply by expanding them with the available objects: $\forall x:
  \phi(x)$ turns into $\bigwedge_{c \in C} \phi(c)$ and $\exists x:
  \phi(x)$ turns into $\bigvee_{c \in C} \phi(c)$. Iterate until no
  more quantifiers are left.
\end{enumerate}
Since STRIPS allows only conjunctions of positive atoms, some further
transformations are necessary.
\begin{enumerate}
\item[(2)] Formulas are brought into negation normal form: $\neg (\phi
  \wedge \psi)$ turns into $\neg \phi \vee \neg \psi$ and $\neg (\phi
  \vee \psi)$ turns into $\neg \phi \wedge \neg \psi$. Iterate until
  negation is in front of atoms only.
\item[(3)] For each $\neg x$ that occurs in a formula: introduce a new
  predicate $not\mbox{-}x$; set $not\mbox{-}x \in I$ iff $x \not \in
  I$; for all effects $e$: set $not\mbox{-}x \in add(e)$ iff $x \in
  del(e)$ and $not\mbox{-}x \in del(e)$ iff $x \in add(e)$; in all
  formulas, replace $\neg x$ with $not\mbox{-}x$. Iterate until no
  more negations are left.
\item[(4)] Transform all formulas into DNF: $(\phi_1 \vee \phi_2)
  \wedge (\psi_1 \vee \psi_2)$ turns into $(\phi_1 \wedge \psi_1) \vee
  (\phi_1 \wedge \psi_2) \vee (\phi_2 \wedge \psi_1) \vee (\phi_2
  \wedge \psi_2)$. Iterate until no more conjunctions occur above
  disjunctions.  If an operator precondition $pre(o)$ has $n>1$
  disjuncts, then create $n$ copies of $o$ each with one disjunct as
  precondition. If an effect condition $con(e)$ has $n>1$ disjuncts,
  then create $n$ copies of $e$ each with one disjunct as condition.
  If $G$ has $n>1$ disjuncts, then introduce a new fact
  $goal\mbox{-}reached$, set $G := goal\mbox{-}reached$, and create
  $n$ new operators each with one disjunct as precondition and a
  single unconditional effect adding $goal\mbox{-}reached$.
\end{enumerate}

\begin{figure}[htb]
{\small
\begin{tabbing}
(\=:action move\\
\> :parameters\\
\> (?a - airplane ?t - airplanetype ?d1 - direction ?s1 ?s2  - segment ?d2 - direction)\\
\> :precondition\\
\>  (and \= (has-type ?a ?t) (is-moving ?a) (not (= ?s1 ?s2)) (facing ?a ?d1) (can-move ?s1 ?s2 ?d1)\\
\> \>       (move-dir ?s1 ?s2 ?d2) (at-segment ?a ?s1)\\
\> \>       (not (exists (?a1 - airplane) (and (not (= ?a1 ?a)) (blocked ?s2 ?a1))))\\
\> \>       (forall (?s - segment)  (imply  (and  (is-blocked ?s ?t ?s2 ?d2) (not (= ?s ?s1))) (not (occupied ?s)))))\\
\> :effect\\
\>  (and \> (occupied ?s2) (blocked ?s2 ?a) (not (occupied ?s1)) (not (at-segment ?a ?s1)) (at-segment ?a ?s2)\\
\> \>       (when   (not (is-blocked ?s1 ?t ?s2 ?d2)) (not (blocked ?s1 ?a)))\\
\> \>       (when   (not (= ?d1 ?d2)) (and (not (facing ?a ?d1)) (facing ?a ?d2)))\\
\> \>       (forall (?s - segment)  (when   (is-blocked ?s ?t ?s2 ?d2) (blocked ?s ?a)))\\
\> \>       (forall \= (?s - segment) (when\\  
\> \> \>    (and  (is-blocked ?s ?t ?s1 ?d1) (not (= ?s ?s2)) (not (is-blocked ?s ?t ?s2 ?d2)))\\
\> \> \>    (not (blocked ?s ?a))))))
\end{tabbing}}
\caption{\label{compilations:airport}An operator from airport ground traffic control.}
\vspace{-0.0cm}
\end{figure}

As an illustrative example, consider the operator description in
Figure~\ref{compilations:airport}, taken from our domain encoding
airport ground traffic control. This operator moves an airplane from
one airport segment to another. Consider specifically the precondition
formula {\small (not (exists (?a1 - airplane) (and (not (= ?a1 ?a))
  (blocked ?s2 ?a1))))}, saying that no airplane different from ``?a''
is allowed to block segment ``?s2'', the segment we are moving into.
Say the set of airplanes is $a_1, \dots, a_n$.  Then step (1) will
turn the formula into {\small (not (or (and (not (= a$_1$ ?a))
  (blocked ?s2 a$_1$)) \dots (and (not (= a$_n$ ?a)) (blocked ?s2
  a$_n$))))}. Step (2) yields {\small (and (or (= a$_1$ ?a) (not
  (blocked ?s2 a$_1$))) \dots (or (= a$_n$ ?a) (not (blocked ?s2
  a$_n$))))}. Step (3) yields {\small (and (or (= a$_1$ ?a)
  (not-blocked ?s2 a$_1$)) \dots (or (= a$_n$ ?a) (not-blocked ?s2
  a$_n$)))}. Step (4), finally, will (naively) transform this into
{\small (or (and (= a$_1$ ?a) \dots (= a$_n$ ?a)) \dots (and
  (not-blocked ?s2 a$_1$) \dots (not-blocked ?s2 a$_n$)))}, i.e., more
mathematically notated:
\[
\bigvee_{\overline{x} \in \{\mbox{\small (= a$_1$ ?a)}, \mbox{\small
    (not-blocked ?s2 a$_1$)}\} \times \dots \times \{\mbox{\small (=
    a$_n$ ?a)}, \mbox{\small (not-blocked ?s2 a$_n$)}\}} \bigwedge
\overline{x}.
\]
In words, transforming the formula into a DNF requires enumerating all
$n$-vectors of atoms where each vector position $i$ is selected from
one of the two possible atoms regarding airplane $a_i$. This yields an
exponential blow-up to a DNF with $2^n$ disjuncts. The DNF is then
split up into its single disjuncts, each one yielding a new copy of
the operator.

The reader will have noticed that an exponential blow-up is also
inherent in compilation step (1), where each quantifier may be
expanded to $|C|$ sub-formulas, and $k$ nested quantifiers will be
expanded to $|C|^k$ sub-formulas. Obviously, in general there is no
way around either of the blow-ups, other than to deal with more
complex formulas than allowed in STRIPS. In practice, however, these
blow-ups can typically be dealt with reasonably well, thanks to the
relative simplicity of operator descriptions, and the frequent
occurrence of {\em static} predicates, explained shortly. If
quantifiers aren't deeply nested, like in
Figure~\ref{compilations:airport}, then the blow-up inherent in step
(1) does not matter. Transformation to DNF is more often a problem --
like in our example here. The key to successful application of the
compilation in practice, at least as far as our personal experience
goes, is the exploitation of static predicates.
% ST I think you should remove this citation below. It must have been
% pretty obvious to everyone that this is a good to do. I was
% definitely doing this in all my planners, since I started in 1991
% ... I think it is just that nobody bothered writing about it.  JH:
% well, it isn't mentioned by gazen and knoblock. reformulatd the
% sentence a bit. 
This idea is described, for example, by
\citeA{koehler:hoffmann:ecai-ws-00}.  Static predicates aren't
affected by any operator effect. Such predicates can be easily found,
and their truth value is fully determined by the initial state {\em as
  soon as they are fully instantiated}. In the above transformation
through step (4), the operator parameters are still variables, and
even if we knew that ``='' is (of course) a static predicate, this
would not help us because we wouldn't know what ``?a'' is. If we
instantiate ``?a'', however, then, in each such instantiation of the
operator, the ``(= ?a1 ?a)'' atoms trivialize to TRUE or FALSE, and
the large DNF collapses to the single conjunction {\small
  $\bigwedge_{\mbox{\small $a \neq$ ?a1 airplane}}$ (not-blocked ?s2
  ?a1)}, where ``$a$'' is our instantiation of ``?a''. Similarly, the
expansion of quantifiers is often made much easier by first
instantiating the operator parameters, and then inserting TRUE or
FALSE for any static predicate as soon as its parameters are grounded.
Inserting TRUE or FALSE often simplifies the formulas significantly
once this information is propagated upwards (e.g., a disjunction with
a TRUE element becomes TRUE itself).

Assuming our compilation succeeded thus far, after steps (1) to (4)
are processed we are down to a STRIPS description with conditional
effects, i.e., the actions still have conditional effects $con(e)$,
$add(e)$, $del(e)$ where $con(e)$ is a conjunction of atoms. This
subset of ADL has been termed ``SIMPLE-ADL'' by Fahiem Bacchus, who
used it for the encoding of one of the versions of the ``Elevator''
domain used in IPC-2 (i.e. the 2000 competition). We can now choose to
leave it in this language, necessitating a planning algorithm that can
deal with conditional effects directly. Several existing planning
systems, for example FF \cite{hoffmann:nebel:jair-01} and IPP
\cite{koehler:etal:ecp-97}, do this. It is a sensible approach since,
as \citeA{nebel:jair-00} proved, conditional effects cannot be
compiled into STRIPS without either an exponential blow-up in the task
description, or a linear increase in plan length. One might suspect
here that, like with steps (1) and (4) above, the ``exponential
blow-up'' can mostly be avoided in practice. The airport move operator
in Figure~\ref{compilations:airport} provides an example of this. All
effect conditions are static and so the conditional effects disappear
completely once we instantiate the parameters -- which is another good
reason for doing instantiation prior to the compilation. However, the
conditional effects do {\em not} disappear in many other, even very
simple, natural domains. Consider the following effect, taken from the
classical Briefcaseworld domain:
\[
\mbox{\small (forall (?o) (when (in ?o) (and (at ?o ?to) (not (at ?o
  ?from)))))}
\]
The effect says that any object ``?o'' that is currently in the
briefcase moves along with the briefcase. Obviously, the effect
condition is {\em not} static, and the outcome of the operator will
truly depend on the contents of the briefcase. Note that the
``forall'' here means that we actually have a {\em set} of (distinct)
conditional effects, one for each object.

There are basically two known methods to compile conditional effects
away, corresponding to the two options left open by Nebel's
\citeyear{nebel:jair-00} result. The first option is to enumerate all
possible {\em combinations} of effect outcomes, which preserves plan
length at the cost of an exponential blow-up in description size --
exponential in the number of different conditional effects of any
single action.  Consider the above Briefcaseworld operator, and say
that the object set is $o_1, \dots, o_n$. For every subset $o_1',
\dots o_k'$ of $o_1, \dots, o_n$, $o_{k+1}', \dots, o_n'$ being the
complement of the subset, we get a distinct operator with a
precondition that contains all of:
\[
\mbox{\small (in $o_1'$) \dots (in $o_k'$) (not-in $o_{k+1}'$) \dots
  (not-in $o_{n}'$)}
%\mbox{\small (when (and $\bigwedge_{?o \in I}$ (in ?o) $\bigwedge_{?o
%    \not \in I}$ (not-in ?o)) (and $\bigwedge_{?o \in I}$ (at ?o ?to)
%  $\bigwedge_{?o \in I}$ (not (at ?o ?from))))}
\]
Where the effect on the objects is:
\[
\mbox{\small (at $o_1'$ ?to) \dots (at $o_k'$ ?to) (not (at $o_1'$
  ?from)) \dots (not (at $o_k'$ ?from))}
\]
In other words, the operator can be applied (only) if exactly $o_1',
\dots o_k'$ are in the briefcase, and it moves exactly these objects.
Since (in deterministic planning as considered here) there never is
uncertainty about what objects are inside the briefcase and what are
not, exactly one of the new operators can be applied whenever the
original operator can be applied. So the compilation method preserves
the size (nodes) and form (edges) of the state space.  However, we
won't be able to do the transformation, or the planner won't be able
to deal with the resulting task, if $n$ grows beyond, say, maximally
$10 \dots 20$.  Often, real-world operators contain more distinct
conditional effects than that.

The alternative method, first proposed by \citeA{nebel:jair-00}, is to
introduce artificial actions and facts that enforce, after each
application of a ``normal'' action, an effect-evaluation phase during
which all conditional effects of the action must be tried, and those
whose condition is satisfied must be applied. For the above
Briefcaseworld example, this would look as follows. First, the
conditional effect gets removed, a new fact ``evaluate-effects'' is
inserted into the add list, and a new fact ``normal'' is inserted into
the precondition and delete list. Then we have $2n$ new operators, two
for each object $o_i$. One means ``move-along-$o_i$'', the other means
``leave-$o_i$''. The former has ``in($o_i$)'' in its precondition, the
latter ``not-in($o_i$)''.  The former has ``(at $o_i$ ?to)'' and
``(not (at $o_i$ ?from)'' in its effect. Both have
``evaluate-effects'' in their precondition, and a new fact
``tried-$o_i$'' as an add effect. There is a final new operator that
stops the evaluation, whose precondition is the conjunction of
``evaluate-effects'' and ``tried-$o_1$'', \dots, ``tried-$o_n$'',
whose add effect is ``normal'', and whose delete effect is
``evaluate-effects''. If the conditional effects of several operators
are compiled away with this method, then the ``evaluate-effects'' and
``tried-$o_i$'' facts are made specific to each operator; ``normal''
can remain a single fact used by all the operators. If an effect has
$k>1$ facts in its condition, then $k$ ``leave-$o_i$'' actions must be
created, each having the negation of one of the facts in its
precondition.

Nebel's \citeyear{nebel:jair-00} method increases plan length by the
number of distinct conditional effects of the operators. Note that
this is not benign if there are, say, more than $20$ such effects. To
a search procedure that recognizes what the new constructs do, the
search space essentially remains the same as before the compilation.
But, while the artificial constructs can easily be deciphered for what
they are by a human, this is not necessarily true (is likely to not be
the case) for a computer that searches with some general-purpose
search procedure.  Just as an example, in a naive forward search space
there is now a choice of how to {\em order} the application of the
conditional effects (which could be avoided by enforcing some order
with yet more artificial constructs). Probably more importantly,
standard search heuristics are unlikely to recognize the nature of the
constructs. For example, without delete lists it suffices to achieve
all of ``tried-$o_1$'', \dots, ``tried-$o_n$'' just once, and later on
apply only those conditional effects that are needed.

We conclude that if it is necessary to eliminate conditional effects,
whenever feasible, one should compile conditional effects away with
the first method, enumerating effect outcomes. We did so in IPC-4. We
took FF's pre-processor, that implements the transformation steps (1)
to (4) above, and extended it with code that compiles conditional
effects away, optionally by either of the two described methods. We
call the resulting tool ``adl2strips''.\footnote{Executables of
  adl2strips can be downloaded from the IPC-4 web page at
  http://ipc.icaps-conference.org. There is also a download of a tool
  named ``Ground'', based on the code of the Mips system
  \cite{Own:Taming}, that takes in the full syntax of PDDL2.2
  \cite{hoffmann:edelkamp:jair-05} and puts out a grounded
  representation (we did not have to use the tool in IPC-4 since the
  temporal and numeric planners all had their own pre-processing steps
  implemented).} In most cases where we had a domain version
formulated in ADL, we used adl2strips to generate a STRIPS formulation
of that domain version. In one case, a version of power supply
restoration, we also generated a SIMPLE-ADL formulation. In all cases
but one, enumerating effect outcomes was feasible. The single
exception was another version of power supply restoration where we
were forced to use Nebel's \citeyear{nebel:jair-00} method.  Details
of this process, and exceptions where we did not use adl2strips but
some more domain-specific method, are described in the sections on the
individual domains in Appendix~\ref{domains}.

\subsection{Compilations of Derived Predicates}
\label{compilations:dp}

There are several proposals in the literature as to how to compile
derived predicates away, under certain restrictions on their form or
their use in the rest of the domain description
\cite{gazen:knoblock:ecp-97,garagnani:00}. A compilation scheme that
works in general has been proposed by
\citeA{thiebaux:etal:ijcai-03,thiebaux:etal:ai-05}.  Thi\'ebaux et al.
also proved that there is no compilation scheme that works in general
and that does not, in the worst case, involve an exponential blow-up
in either the domain description size or in the length of the plans.
Note here that ``exponential'' refers also to the increase in plan
length, not just to the description blow-up, unlike the compilation of
conditional effects discussed above. This makes the compilation of
derived predicates a rather difficult task. In IPC-4, compilation
schemes oriented at the approaches taken by
\citeA{gazen:knoblock:ecp-97}, and
\citeA{thiebaux:etal:ijcai-03,thiebaux:etal:ai-05}, were used. We
detail this below. First, let us explain what derived predicates are,
and how the compilations work.

Derived predicates are predicates that are not affected by any of the
operators, but whose truth value can be derived by a set of {\em
  derivation rules}. These rules take the form $\phi(\overline{x})
\Rightarrow P(\overline{x})$. The basic intuition is that, if
$\phi(\overline{x})$ is satisfied for an instantiation $\overline{c}$
of the variable vector $\overline{x}$, then $P(\overline{c})$ can be
concluded. More formally, the semantics of the derivation rules are
defined by {\em negation as failure}: starting with the empty
extension, instances of $P(\overline{c})$ are derived until a fixpoint
is reached; the instances that lie outside the fixpoint are assumed to
be FALSE. Consider the following example:
\[
\mbox{\small (:derived (trans ?x ?y) (or (edge ?x ?y ) (exists (?z)
  (and (edge ?x ?z) (trans ?z ?y)))))}
\]
This derivation rule defines the transitive closure over the edges in
a graph. This is a very typical application of derived predicates. For
example, ``above'' in the Blocksworld is naturally formalized by such
a predicate; in our power supply restoration domain, transitive
closure models the power flow over the paths in a network of electric
lines. Obviously, the pairs ``?x'' and ``?y'' that are {\em not}
transitively connected are those that do not appear in the fixpoint --
negation as failure.

Matters become interesting when we think about how derived predicates
are allowed to refer to each other, and how they may be used in the
rest of the task description. Some important distinctions are: Can a
derived predicate appear in the antecedent of a derivation rule? Can a
derived predicate appear {\em negated} in the antecedent of a
derivation rule? Can a derived predicate appear negated in an action
precondition or the goal?

If derived predicates do not appear in the antecedents of derivation
rules, then they are merely non-recursive macros, serving as syntactic
sugar. One can simply replace the derived predicates with their
definitions.\footnote{If the derived predicates are recursive but
  cycle-free, they can be replaced with their definitions but that may
  incure an exponential blow-up.} If a derived predicate $P$ appears
negated in the (negation normal form of the) antecedent of a
derivation rule for predicate $Q$, then the fixpoints of $P$ and $Q$
can not be computed in an interleaved way: the extension of $Q$ may
differ depending on the order in which the individual instances are
derived. Say the rule for $P$ is $A(\overline{x}) \Rightarrow
P(\overline{x})$, where $A$ is a basic predicate, and the rule for $Q$
is $\neg P(\overline{x}) \Rightarrow Q(\overline{x})$. Say we have
objects $a$ and $b$, and our current state satisfies (only) $A(a)$.
Computing the derived predicates in an interleaved way, we may derive
$A(a) \Rightarrow P(a), \neg A(b) \Rightarrow Q(b)$, and stop; we may
also derive $\neg P(a) \Rightarrow Q(a), \neg A(b) \Rightarrow Q(b),
A(a) \Rightarrow P(a)$. There is a non-monotonic behavior, making it
non-trivial to define what the extension of $B$ is. To keep things
simple -- after all the extensions of the derived predicates must be
computed in every new world state --
\citeA{thiebaux:etal:ijcai-03,thiebaux:etal:ai-05} propose to simply
{\em order $Q$ after $P$}. That is, we compute $P$'s extension first
and then compute $Q$ based on that. Generalized, one ends up with a
semantics corresponding to that of stratified logic programs
\cite{apt:et:al:88}. In the context of IPC-4, i.e., in PDDL2.2
\cite{hoffmann:edelkamp:jair-05}, for the sake of simplicity the use
of negated derived predicates in the antecedents of derivation rules
was not allowed.

Whether or not derived predicates appear negated in action
preconditions or the goal makes a difference for Gazen and Knoblock's
\citeyear{gazen:knoblock:ecp-97} compilation scheme. The idea in that
scheme is to simply replace derivation rules with actions. Each rule
$\phi(\overline{x}) \Rightarrow P(\overline{x})$ is replaced with a
new operator with parameters $\overline{x}$, precondition
$\phi(\overline{x})$ and (add) effect $P(\overline{x})$.  Actions that
can influence the truth value of $\phi$ -- that affect any of the
atoms mentioned in $\phi$ -- delete all instances of $P$.  In words,
the new actions allow the derivation of $P$, and if a normal action is
applied that may influence the value of $P$, then the extension of $P$
is re-initialized.

If derived predicates are not used negated, then Gazen and Knoblock's
\citeyear{gazen:knoblock:ecp-97} compilation scheme works. However,
say $\neg P(\overline{c})$ is contained in some action precondition.
In the compiled version, the planner can achieve this precondition
simply by not applying the ``derivation rule'' -- the action -- that
adds $P(\overline{c})$. That is, the planner now has a choice of what
predicate instances to derive, which of course is not the same as the
negation as failure semantics. The reader may at this point wonder why
we do not compile the negations away first, and thereafter use Gazen
and Knoblock's \citeyear{gazen:knoblock:ecp-97} compilation. The
problem there would be the need for {\em inverse derivation rules}
that work with the negation as failure semantics. It is not clear how
this should be done. Say, for example, we want to define the negated
version of the ``(trans ?x ?y)'' predicate above. One would be tempted
to just take the negation of the derivation rule antecedent:
\[
\mbox{\small (:derived (not-trans ?x ?y) (and (not-edge ?x ?y) (forall
  (?z) (or (not-edge ?x ?z) (not-trans ?z ?y)))))}
\]
This does not work, however. Say every node in the graph has at least
one adjacent edge. Starting with an empty extension of ``(not-trans ?x
?y)'', not a single instantiation can be derived: given any x and y
between which there is no edge, for those z that have an edge to x we
would have to have (not-trans z y) in the first place.

One possible solution to the above difficulties is to extend Gazen and
Knoblock's \citeyear{gazen:knoblock:ecp-97} compilation with
constructs that force the planner to compute the entire extension of
the derived predicates before resuming normal planning. A full
description of this, dealing with arbitrary derivation rules, is
described by \citeA{thiebaux:etal:ijcai-03,thiebaux:etal:ai-05}. In a
nutshell, the compilation works as follows. One introduces flags
saying if one is in ``normal'' or in ``fixpoint'' mode. Normal actions
invoke the fixpoint mode if they affect any predicates relevant to the
derivation rules.  In fixpoint mode, an action can be applied that has
one conditional effect for each derivation rule: if the effect
condition is true, and the respective derived predicate instance is
false, then that predicate instance is added, plus a flag
``changes-made''. Another action tests whether there has been a
fixpoint: if ``changes-made'' is true, then the action just resets it
to false; if ``changes-made'' is false, then the action switches back
to normal mode. To reduce the domain to STRIPS, after this compilation
of derived predicates, the negations and conditional effects must be
compiled away with the techniques explained earlier.

One would imagine that Thiebaux et al.'s
\citeyear{thiebaux:etal:ijcai-03,thiebaux:etal:ai-05} compilation,
making use of rather complicated constructs, tends to confuse domain
independent search techniques. Indeed,
\citeA{thiebaux:etal:ijcai-03,thiebaux:etal:ai-05} report that even a
completely naive explicit treatment of derived predicates in FF
performs a lot better, in some benchmark domains, than the standard
version of FF applied to the compiled benchmarks. Gazen and Knoblock's
\citeyear{gazen:knoblock:ecp-97} compilation makes use of less
artificial constructs, and is thus preferable whenever it can safely
be applied. Note, however, that both compilations imply a potentially
exponential blow-up in plan length: exponential {\em in the arity of
  the derived predicates}. The worst case is that every action affects
the derivation rules, and every re-computation of the extension of the
derived predicates has to go through all those predicates'
instantiations. In such a situation, between every pair of normal
actions the planner has to apply on the order of $|C|^a$ actions,
where $a$ is the maximum arity of any derived predicate. While $a$ is
typically very small -- power supply restoration is the only domain we
are aware of that features a derived predicate with more than two
(four, namely) arguments -- even a plan length increase linear in the
number of objects can mean a quite significant decrease in planner
performance.

Of the IPC-4 benchmarks, derived predicates occur (only) in power
supply restoration (Appendix~\ref{psr}) and model checking safety
properties (Appendix~\ref{promela}). For the latter, where the derived
predicates do not occur negated, Stefan Edelkamp encoded a domain
version without derived predicates by hand, using a method along the
lines of the one described by \citeA{gazen:knoblock:ecp-97}.  For
power supply restoration, where derived predicates do occur negated,
we used a variation of the method described by
\citeA{thiebaux:etal:ijcai-03,thiebaux:etal:ai-05}. In both cases, due
to the increase in plan length we considered the resulting domain
formulation too different from the original formulation to be directly
compared with it, in terms of planner performance. So the compiled
formulations were posed to the competitors as distinct domain {\em
  versions}, instead of alternative domain version formulations.
Indeed, just as we expected, planner results in IPC-4 were much worse
for the compiled encodings.

\subsection{Compilations of Timed Initial Literals}
\label{compilations:til}

Timed initial literals are literals that are known to become true at
time points pre-specified in the initial state.  Such literals can be
compiled into durational PDDL relatively easily, at the cost of the
plan length and the domain description size blowing up linearly in the
number of timed initial literals. The compilation was proposed and
brought to our attention by \citeA{expressivepddl}.  The idea is to
use a ``wrapper'' action that must be applied before any other action,
and whose duration is the occurrence time of the last timed initial
literal. The planner must also apply a sequence of ``literal'' actions
that achieve all the timed initial literals by order of occurrence,
the durations being the time intervals between the occurrences.  When
the ``wrapper'' action has terminated, the ``literal'' actions can no
longer be applied.  So the planner is forced to apply them all in
direct sequence. This suffices to encode the desired semantics.
Consider the following example:

{\small
\begin{tabbing}
(:init \=\\
\>  (at 9  (have-to-work))\\
\>  (at 19 (not (have-to-work)))\\
\>  (at 19 (bar-open))\\
\>  (at 23 (not (bar-open))))
\end{tabbing}}

\noindent
To encode this in standard durational PDDL, the ``wrapper'' will be:
{\small
\begin{tabbing}
(\=:action wrapper\\
\> :parameters ()\\
\>  :duration (= ?duration 23)\\
\> :condition\\
\>  (at start (no-wrapper))\\
\> :effect\\
\>  (and \=  (at start (not (no-wrapper)))\\
\> \> (at start (wrapper-started))\\
\> \> (at start (wrapper-active))\\
\> \> (at start (literal-1-started))\\
\> \> (at end (not (wrapper-active)))))\\
\end{tabbing}}

\noindent
Here, ``no-wrapper'' ensures only one wrapper action is executed;
``wrapper-started'' is inserted into the precondition of every normal
action and thus ensures that the wrapper is started before any other
action is executed; ``wrapper-active'' will be a precondition of the
``literal'' actions.  Precisely, these will be: {\small
\begin{tabbing}
(\=:action literal-1\\
\> :parameters ()\\
\>  :duration (= ?duration 9)\\
\> :condition\\
\>  (and \= (over all (wrapper-active))\\
\> \>  (over all (literal-1-started)))\\
\> :effect\\
\>  (and \=  (at end (not (literal-1-started)))\\
\> \> (at end (literal-2-started))\\
\> \> (at end (have-to-work))))\\
\\
(\=:action literal-2\\
\> :parameters ()\\
\>  :duration (= ?duration 10)\\
\> :condition\\
\>  (and \= (over all (wrapper-active))\\
\> \>  (over all (literal-2-started)))\\
\> :effect\\
\>  (and \=  (at end (not (literal-2-started)))\\
\> \> (at end (literal-3-started))\\
\> \> (at end (not (have-to-work)))\\
\> \> (at end (bar-open))))\\
\\
(\=:action literal-3\\
\> :parameters ()\\
\>  :duration (= ?duration 4)\\
\> :condition\\
\>  (and \= (over all (wrapper-active))\\
\> \>  (over all (literal-3-started)))\\
\> :effect\\
\>  (and \=  (at end (not (literal-3-started)))\\
\> \> (at end (not (bar-open)))\\
\> \> (at end (literals-done))))\\
\end{tabbing}}

\noindent
The fact ``literals-done'' will be made a goal, so the planner must
actually apply the ``literal'' actions. Note that we need only three
of these actions here, since two of the timed initial literals -- no
longer having to work and the opening of the bar -- are scheduled to
occur at the same time. Note also that, as with Nebel's
\citeyear{nebel:jair-00} compilation of conditional effects and
Thiebaux et al.'s
\citeyear{thiebaux:etal:ijcai-03,thiebaux:etal:ai-05} compilation of
derived predicates, the compiled encoding is likely to be confusing
for domain independent search methods.

Many of the IPC-4 domains made use of timed initial literals (in some
versions) to encode various kinds of time windows (see
Appendix~\ref{domains}). We compiled these domain versions into pure
(durational) PDDL as above, and provided the resulting encodings as
additional domain {\em versions}. Due to the increase in the number of
actions needed for the plans, we figured that the compilation
constructs were too much of a change for direct comparison. Indeed, as
with the derived predicates, planner results in IPC-4 were much worse
for the domain versions compiled in this way.

%%% Local Variables: 
%%% mode: latex
%%% TeX-master: t
%%% End: 

\section{A Summary of the Domains}
\label{overview}

In this section we provide a brief summary of the IPC-4 domains. For
each domain, we provide: a short description of the application; our
motivation for inclusion of the domain; a brief explanation of the
main simplifications made for IPC-4; and a brief explanation of the
different domain versions and formulations used in IPC-4. We proceed
in alphabetical order.

\subsection{Airport}

We had a contact person for this application domain, Wolfgang Hatzack,
who has been working in this application area for several years. The
domain was adapted for IPC-4 by J\"org Hoffmann and Sebastian Tr\"ug

\noindent {\bf Application.} The task here is to control the ground
traffic at an airport. Timed travel routes must be assigned to the
airplanes so that they reach their targets. There is inbound and
outbound traffic; the former are airplanes that must take off, the
latter are airplanes that have just landed and have to park. The main
problem constraint is, of course, to ensure the safety of the
airplanes. This means to avoid collisions, and also to prevent
airplanes from entering the unsafe zones behind large airplanes that
have their engines running. The optimization criterion is to minimize
the summed up travel time (on the surface of the airport) of all
airplanes.\footnote{An alternative criterion would be to minimize the
  summed up squared delay of all airplanes. This is in the interest of
  the airlines; minimizing summed up travel time is in the interest of
  the airport. Neither of the two can be easily modelled in PDDL2.2,
  as we discuss in Simplifications, below.} There usually are {\em
  standard routes}, i.e., routes that any airplane must take when
outbound from a certain parking area, or inbound from a certain
runway. The reason for introducing such routes is to reduce complexity
for human ground controllers, since significant computer support is
not yet available at real airports.  Solving instances optimally (the
corresponding decision problem) is {\bf PSPACE}-hard without standard
routes \cite{helmert:icaps-06} and {\bf NP}-complete if {\em all}
routes are standardized \cite{hatzack:nebel:ecp-01}. In the latter
case, we have a pure scheduling problem. In the former case,
complicated -- but unrealistic -- airport traffic situations can lead
to exponentially long solutions, see Section~\ref{known:complexity}.

\noindent {\bf Motivation.} Our main motivation for including this
domain was that we were able to model the application quite
accurately, and, in particular, to generate quite realistic instances.
In fact, we were able to generate instances based on a real airport.
This was made possible by our contact to Wolfgang Hatzack, who
completed a PhD about this application \cite{hatzack:diss-02}. Apart
from developing domain-specific solutions \cite{hatzack:nebel:ecp-01},
he developed a realistic simulation tool, which he kindly supplied to
us for the purpose of generating the IPC-4 domain versions and test
instances.  Sebastian Tr\"ug implemented options inside the simulator
that allowed it, at any point in time during the simulation of traffic
flow, to output the current traffic situation in PDDL format. The
simulator
%ST Why no \" in Zuerich? Because you would have to write M\"uchen?
included the real airports Frankfurt, Zurich, and Munich. Frankfurt
and Zurich proved too large for our purposes, but we were able to
devise competition instances based on Munich airport.

\noindent {\bf Simplifications.} We had to make two simplifications.
The first amounts to a discretization of space (location) on the
airport, making the domain amenable to PDDL style discrete actions.
With a continuous space representation, one would need actions with a
continuous choice of {\em how far} to move.  While the discretization
loses precision, we believe that it does not distort the nature of the
problem too much. Due to the amount of expected conflicting traffic at
different points in the airport, which is high only at parking
positions, it is relatively easy to choose a discretization -- with
segments of {\em different} length -- that is precise and small enough
at the same time. Our second simplification is more severe: we had to
drop the original optimization criterion, which is very awkward to
express in current PDDL. To model the travel times of the airplanes,
one needs access to the times at which the plans {\em wait}, i.e., do
nothing.\footnote{The same difficulty arises in the modelling of
  delay, for which one must also compute the travel times.} We are not
aware of a way to express this in current PDDL. The IPC-4 committee
voted against the introduction of an additional language construct, a
``look at the clock'', since that didn't seem relevant anywhere else.
Another option would be to introduce explicit waiting actions, which
causes a lot of trouble because, similar to continuous space, there
must be a continuous choice of {\em how long} to wait. In the end, we
decided to just drop the criterion for now, and ask the planners to
optimize standard makespan instead,\footnote{Makespan, in Planning,
  means the amount of time from the start of the plan until the last
  action stops executing.} corresponding to the arrival time of the
last airplane (meaning, arrival at the destination in the airport).
This is not ideal, but a reasonable optimization criterion.  No
planning system participating in IPC-4, with the single exception of
LPG-td \cite{gerevini:etal:jair-06}, was able to take account of
general optimization criteria other than the built-in ones (like
makespan). We did not use full standard routes, thus allowing the
airplanes a choice of where to move. We {\em did} use standards for
some routes, particularly the regions near runways in large airports.
For one thing, this served to keep large airports manageable for the
PDDL encoding and planners; for another thing, it seems a good
compromise of exploiting the capabilities of computers while at the
same time remaining close to existing practice.

\noindent {\bf Versions and Formulations.} We generated four versions
of the airport domain: a non-temporal one; a temporal one; a temporal
one with time windows, where the fact that planes will land in the
future and block certain runways is modeled using timed initial
literals; and the latter version, but with timed initial literals
compiled away.  In all versions, the constraints ensuring airplane
safety are modelled with ADL logical formulas. A compilation of these
into partially grounded STRIPS provides, in each version, an
alternative formulation: each domain version has one ADL formulation
and one STRIPS formulation.

\subsection{Pipesworld}

Frederico Liporace has been working in this application area for
several years; he submitted a paper on an early domain version to the
workshop on the competition at ICAPS'03.  The domain was adapted for
IPC-4 by Frederico Liporace and J\"org Hoffmann.

\noindent {\bf Application.} Here the task is to control the flow of
different oil derivatives through a pipeline network, so that certain
product amounts are transported to their destinations. Pipeline
networks are graphs consisting of areas (nodes) and pipes (edges),
where the pipes can differ in length. The available actions are to
pump liquid into ends of pipes, with the effect that the liquid at the
other end of the pipe gets ejected. The application is rich in
additional constraints, like, constraints on what types of products
may interface within a pipe, restricted tankage space in areas, and
deadlines for arrival of products.

\noindent {\bf Motivation.} Our main motivation for including this
domain was its original structure. If one inserts something into a
pipe at one end, something possibly completely different comes out of
the pipe at its other end. In this way, {\em changing the position of
%ST objects? From Malte's description, I had the feeling it
%could only change the position of one other batch besides the
%pushed batch.
  one object directly results in changing the position of several
  other objects} -- namely, all objects inside the affected pipeline.
This is not the case in any other transportation domain we are aware
of, in fact it is more reminiscent of complicated single-player games
such as Rubik's Cube. Indeed, the strong interaction between objects
can lead to several subtle phenomena. For example, there are instances
where any solution must pump liquid through a ring of pipeline
segments in a cyclic fashion.

\noindent {\bf Simplifications.} We had to severely simplify this
domain in order to be able to solve reasonably complex instances with
current planners. Most importantly, our encoding is heavily based on
assuming a smallest indivisible unit of liquid, a {\em batch}. Every
amount of liquid in the encoding is modelled in terms of a number of
batches. To capture the continuous nature of the real application,
this means that one has to choose batch size in a trade-off between
encoding size and accuracy. The trade-off is less well-behaved than
the one in Airport (choosing ``segments'' sizes) since the unit size
cannot be made flexible: every batch may pass through every pipeline,
and so the smallest batch governs the discretization of all pipelines.
This is in contrast to Airport, where segments may vary in size.  As
another important simplification, we used ``personalized'' goals, i.e.
the goals referred to specific batch objects rather than to product
amounts.  This serves to avoid large disjunctions enumerating all
possible combinations of individual batches. The simplifications are
quite severe and indeed it seems unlikely that a realistic
representation of Pipesworld, in particular with real-valued product
amounts instead of batches, could be solved efficiently by planners
without introducing more specialized language constructs -- a sort of
``queue'' data structure -- into PDDL, see
Appendix~\ref{pipesworld:future}.

\noindent {\bf Versions and Formulations.} We created six different
versions of Pipesworld: four versions with / without temporal actions,
and with/without tankage restrictions, respectively; one temporal
version without tankage restrictions but with arrival deadlines for
the goal batches; one version identical to the last one except that
timed initial literals were compiled away.

\subsection{Promela}

This domain was created for IPC-4 by Stefan Edelkamp.

\noindent {\bf Application.} Here the task is to validate 
properties in systems of communicating processes 
(often communication protocols), encoded in the
Promela  language. Promela (PROcess MEta LAnguage) is the input language 
of the model checker SPIN \cite{Holzmann:NewBook}. 
The language is loosely based on Dijkstra's guarded command
language, borrowing some notation from
Hoare's CSP language. One important property check is to detect
{\em deadlock} states, where none of the processes can apply a
transition. For example, a process may be blocked when trying to read
data from an empty communication channel. \citeA{Own:Promela}
developed an automatic translation from Promela into PDDL, which was
extended to generate the competition examples.

\noindent {\bf Motivation.} Our main motivation for including this
domain was to further promote and make visible the important
connection between Planning and Model Checking. \emph{Model
  Checking}~\cite{Clarke:ModelChecking} itself is an automated formal
method that basically consists of three phases: modeling,
specification and checking. In the first two phases both the system
and the correctness specification are modeled using some formalism.
The last step automatically checks if the model satisfies its
specification. Roughly speaking, this step analyzes the state space of
the model to check the validity of the specification.
% The main limitation of model checking, the \emph{state
%   explosion problem}, consists of a combinatorial explosion of
% the size of the state spacof the model with respect to the number
% of system variables.
 Especially in concurrent systems, where several components interact,
 state spaces grow exponentially in the size of the components of the
 system. There are two main research branches in model checking:
 \emph{explicit-state model checking}, as implemented in SPIN,
 exploits automata theory and stores each explored state individually,
 while \emph{symbolic model checking} describes sets of states and
 their properties using binary decision diagrams (BDDs) or other
 efficient representations for Boolean formulas.

 Checking the validity of a {\em reachability property}, a property
 that asks if a system state with a certain property is reachable, is
 very similar to the question of plan existence. The use of model
 checking approaches to solve planning problems has been explored in
 some depth, e.g.\ by
 \citeA{Cimatti:Universal,Bertoli:ProbPlanning,Lago:Extended,TALPlan,TLPlan,Stoer:BDDPlan,Fourman:PropPlan,Own:Taming,Dierks2,Sylvie:LTL}.
 However, not much has been done in the inverse direction, applying
 planners to model checking problems. Running IPC-4 planners on
 planning encodings of Promela specifications is a first step in doing
 just that.

 The Promela domain also contributes unusual structural properties to
 our domain set; the computational complexity and local search
 topology are quite different as will be discussed in
 Section~\ref{known}.

\noindent {\bf Simplifications.} The main simplification we had to
make was to use very simple example classes of communicating
processes. As PDDL models refer to fixed-length state vectors, we
could not include process construction calls. 
%ST English
%We, therefore, restricted to \emph{active processes}, 
We therefore only considered \emph{active processes}, i.e., processes
that are called only once at initialization time. PDDL also does not
support temporally extended goals, so we had to consider reachability
properties only.  Moreover, by the prototypical nature of our language
compiler, many features of Promela
%ST english
%like 
such as rendezvous communication were not supported. Although we have
limited support of shared variables, during the competition we chose
simple message passing protocols only; and while we experimented with
other reachability properties, the PDDL goals in the competition event
were on deadlock detection only.  Concretely, the IPC-4 instances come
from two toy examples used in the area of Model-Checking: the
well-known ``Dining Philosophers'' problem, and an ``Optical
Telegraph'' problem which can be viewed as a version of Dining
Philosophers where the philosophers have a complex inner life,
exchanging data between the two hands (each of which is a separate
process). In both, the goal is to reach a deadlock state.

\noindent {\bf Versions and Formulations.} We created eight different
versions of the domain. They differ by the Promela example class
encoded (two options), by whether or not they use numeric variables in
the encoding, and by whether or not they use derived predicates in the
encoding. The four encodings of each Promela example class are
semantically equivalent in the sense that there is a 1-to-1
correspondence between plans. We decided to make them different
versions, rather than formulations, because derived predicates make a
large difference in plan length, and numeric variables make a large
difference in 
%ST english
%applicable 
applicability of planning algorithms/systems. The translation
from Promela to PDDL makes use of ADL constructs, so each domain
version contains one ADL formulation and one (fully grounded) compiled
STRIPS formulation.

\subsection{PSR}

Sylvie Thi\'ebaux and others have worked on this application domain.
The domain was adapted for IPC-4 by Sylvie Thi\'ebaux and J\"org
Hoffmann.

\noindent {\bf Application.} The task in PSR (power supply
restoration) is to reconfigure a faulty power distribution network so
as to resupply customers affected by the faults.  The network consists
of electric lines connected by switches and fed via a number of power
sources that are equipped with circuit-breakers. When faults occur,
the circuit-breakers of the sources feeding the faulty lines open to
protect the network, leaving not only these lines but also many
healthy ones un-supplied. The network needs to be reconfigured by
opening and closing switches and circuit-breakers in such a way as to
resupply the healthy portions. Unreliable fault sensors and switches
lead to uncertainty about the state of the network. Furthermore,
breakdown costs that depend on various parameters need to be optimized
under constraints on the capacity of sources and lines.  The
application is a topic of ongoing interest in the field of power
distribution, and has been investigated by the AI community for a long
time, including from an AI planning standpoint
\cite{thiebaux:etal:uai-96,thiebaux:cordier:ecp-01,bertoli:etal:ecai-02,bonet:thiebaux:icaps-03}.

\noindent {\bf Motivation.} Our motivation for including PSR was
twofold.  First, it is a well-researched interesting application
domain.  Second, it has an original structure rarely found in previous
benchmarks. The most natural encoding models the power propagation
using recursive derived predicates that compute the transitive closure
of the connectivity relation in the network.  
%ST trust me (whoever changed it)
%In difference to 
In contrast with most other planning benchmarks, the number of actions
needed in an optimal plan does not necessarily grow with instance
size: the available actions are to alter the position of switches, and
even in a large network altering the position of just a few switches
may suffice for reconfiguration. The difficult question to answer is,
{\em which} switches.

\noindent {\bf Simplifications.} Three major simplifications had to be
made.  First, for deterministic planning we had to assume that the
network state is fully observable, i.e., that the initial state
description is complete, and that the actions always succeed.  Second,
we ignored all numerical and optimization aspects of PSR.  Third, we
used personalized goals in the sense that the lines to be supplied are
named explicitly in the goal.  Note that, even in this simplified
form, the domain exhibits the structure explained above.

\noindent {\bf Versions and Formulations.} We created four domain
versions, differing primarily by size and available formulations. The
most natural domain formulation is in ADL with derived predicates.
Though we experimented with many combinations of PDDL encodings and
compilation strategies, the size of the instances that we could
compile into simpler languages was quite restricted. Precisely, the
versions are: a ``large'' version in ADL plus derived predicates; a
``middle'' version that we could devise also in SIMPLE-ADL plus
derived predicates and in STRIPS plus derived predicates; a
``middle-compiled'' version in ADL, identical to the ``middle''
version except that the derived predicates were compiled away; and a
``small'' version in pure STRIPS. The instances in the latter domain
version had to be particularly small, since it was extremely difficult
to come up with an encoding in pure STRIPS that did not either yield
prohibitively long plans, or prohibitively large PDDL descriptions. In
fact, to obtain the ``small'' version we applied a pre-computation
step \cite{bertoli:etal:ecai-02} that obviates the need for reasoning
about power propagation and, consequently, the need for derived
predicates. In the resulting tasks, opening or closing a switch
directly -- without the detour to power propagation -- affects other
parts of the network. Thus the planner no longer needs to compute the
flow of power through the network, but is left with the issue of how
to configure that flow.

\subsection{Satellite}

This domain was introduced by \citeA{long:fox:jair-03} for IPC-3; it
was adapted for IPC-4 by J\"org Hoffmann. The domain comes from a NASA
space application, where satellites have to take images of spatial
phenomena. Our motivation for inclusion in IPC-4 was that the domain
is application-oriented in a similar sense to the new domains. Also,
we wanted to have some immediate comparison between the performance
achieved at IPC-3, and that achieved at IPC-4.  On top of the 5 domain
versions used in IPC-3, we added 4 new versions, introducing
additional time windows (formulated alternatively with timed initial
literals or their compilation) for the sending of data to earth.

\subsection{Settlers} 

This domain was also introduced by \citeA{long:fox:jair-03} for IPC-3.
The task is to build up an infrastructure in an unsettled area,
involving the building of housing, railway tracks, sawmills, etc. The
distinguishing feature of the domain is that most of the domain
semantics are encoded in numeric variables. This makes the domain an
important benchmark for numeric planning. For that reason, and because
at IPC-3 no participant could solve any but the smallest instances, we
included the domain into IPC-4. No modification was made except that
we compiled away some universally quantified preconditions in order to
improve accessibility.

\subsection{UMTS}

Roman Englert has been working in this application area for several
years.  The domain was adapted for IPC-4 by Stefan Edelkamp and
Roman Englert.

\noindent {\bf Application.}  The third generation of mobile
communication, the so-called UMTS~\cite{Holma:UMTS}, makes available a
broad variety of applications for mobile terminals. With that comes
the challenge to {\em maintain} several applications on one terminal.
First, due to limited resources, radio bearers have restrictions in
the quality of service (QoS) for applications. Second, the cell setup
for the execution of several mobile applications may lead to
unacceptable waiting periods for the user. Third, the QoS may be
insufficient during the call setup in which case the execution of the
mobile application is shut down. Thus arises the call {\em setup}
problem for several mobile applications. The main requirement is, of
course, to do the setup in the minimum possible amount of time. This
is a (pure) scheduling problem that necessitates ordering and
optimizing the execution of the modules needed in the setup. As for
many scheduling problems, finding {\em some}, not necessarily optimal,
solution is trivial; the main challenge is to find good-quality
solutions, optimal ones ideally.

\noindent {\bf Motivation.} Our main motivation for modelling this
pure scheduling problem as a planning domain was that there is a
strong industrial need for {\em flexible} solution procedures for the
UMTS call setup, due to the rapidly evolving nature of the domain,
particularly of the sorts of mobile applications that are available.
The ideal solution would be to just put an automatic planner on the
mobile device, and let it compute the optimized schedules on-the-fly.
In that sense, UMTS call setup is a very natural and promising field
for real-world application of automatic planners. This is also
interesting in the sense that scheduling problems have so far not been
central to competitive AI planning, so our domain serves to advertise
the usefulness of PDDL for addressing certain kinds of scheduling
problems.

\noindent {\bf Simplifications.} The setup model we chose only
considers coarse parts of the network environment that are present
when UMTS applications are invoked.  Action duration is fixed rather
than computed based on the network traffic. The inter-operational
restrictions between different concurrent devices were also neglected.
We considered plausible timings for the instances rather than
real-application data from running certain applications on a UMTS
device. We designed the domain for up to 10 applications on a single
device. This is a challenge for optimal planners computing minimum
makespan solutions, but not so much a challenge for satisficing
planners.

\noindent {\bf Versions and Formulations.} We created six domain
versions; these arise from two groups with three versions each.  The
first group, the standard UMTS domain, comes with or without timing
constraints. The latter can be represented either using timed initial
literals, or their compilation; as before, we separated these two
options into different domain versions (rather than domain version
formulations) due to the increase in plan size. The second group of
domain versions has a similar structure. The only difference is that
each of the three domain versions includes an additional ``flaw''
action. With a single step, that action achieves one needed fact,
where, normally, several steps are required.  However, the action is
useless in reality because it deletes another fact that is needed, and
that cannot be re-achieved. The flaw action was added to see what
happens when we intentionally stressed planners: beside increasing the
branching factor, the flaw action {\em does} look useful from the
perspective of a heuristic function that ignores the delete lists.

%%% Local Variables: 
%%% mode: latex
%%% TeX-master: t
%%% End: 

\section{Known (Theoretical) Results on Domain Structure}
\label{known}

In this section, we start our structural analysis of the IPC-4 domains
by summarizing some known results from the literature.
\citeA{helmert:icaps-06} analyzes the domains from a perspective of
domain-specific computational complexity.  \citeA{hoffmann:jair-05}
analyzes all domains used in the IPCs so far, plus some standard
benchmarks from the literature, identifying topological properties of
the search space surface under the ``relaxed plan heuristic'' that was
introduced with the FF system \cite{hoffmann:nebel:jair-01}, and
variants of which are used in many modern planning systems. Both
studies are exclusively concerned with purely propositional --
non-temporal STRIPS and ADL -- planning. In what follows, by the
domain names we refer to the respective (non-temporal) domain
versions.\footnote{The UMTS domain, which has only temporal versions,
  is not treated in either of the studies. As for computational
  complexity, it is easy to see that deciding plan existence is in
  {\bf P} and deciding bounded plan existence (optimizing makespan) is
  {\bf NP}-complete for UMTS.  Topological properties of the relaxed
  plan heuristic haven't yet been defined for a temporal setting.}

\subsection{Computational Complexity}
\label{known:complexity}

Helmert \citeyear{helmert:icaps-06} has studied the complexity of plan
existence and bounded plan existence for the IPC-4 benchmark problems.
Plan existence asks whether a given planning task is solvable.
Bounded plan existence asks whether a given planning task is solvable
with no more than a given number of actions. Helmert established the
following results.

In Airport, both plan existence and bounded plan existence are
\pspace-complete, even when all aircraft are inbound and just need to
taxi to and park at their goal location, the map is planar and
symmetric, and the safety constraints simply prevent planes from
occupying adjacent segments.  The proof is by reduction from the
Sliding Tokens puzzle, where a set of tokens must reach a goal
assignment to the vertices of a graph, by moving to adjacent vertices
while ensuring that no two tokens ever find themselves on adjacent
vertices. The length of optimal sequential plans can be exponential in
the number of tokens, and so likewise in the airport domain. Even
parallel plans can only be shorter by a linear amount, since each
plane can move at most once per time step. The proof for the Sliding
Tokens puzzle is quite complicated because it involves construction of
instances with exponentially long optimal plans. As one would expect,
the constructions used are more than unlikely to occur on a real
airport; this is in particular true for the necessary density of
conflicting ``traffic'' on the graph structure. We consider this
interesting since it makes Airport a benchmark with an extremely high
worst-case complexity, but with a {\em much} more good-natured typical
case behavior. Typically, there is ample space in an airport for
(comparatively) few airplanes moving across it.

In Pipesworld, whether with or without tankage, both plan existence
and bounded plan existence are \np-hard. It is unknown whether they
are in \np, however. The \np-hardness proof is by reduction from SAT
with at most four literals per clause and where each variable occurs
in at most 3 clauses. Such a SAT instance is reduced to a network in a
way so that parts of the network (variable subnetworks) represent the
choice of an assignment for each of the variables, and other parts
(clause subnetworks) represent the satisfaction of each of the
clauses. The content of areas and pipes are initialized with batches
in a way so that interface restrictions will guarantee that a goal
area is reached by a certain batch in each clause subnetwork iff the
clause is satisfied by the assignment.

For general Promela planning, as defined by Edelkamp
\citeyear{Own:Promela}, both plan existence and bounded plan existence
are \pspace-complete. The \pspace-hardness proof is by reduction from
the halting problem in space-restricted Turing Machines (TM). The
cells of the machine's tape are each mapped onto a process and a queue
of unit capacity, the states of the TM form the set of Promela
messages, the TM's alphabet form the set of Promela states in all
processes, and the Promela transitions encode the TM's transitions. It
can be shown that the TM halts iff the Promela task reaches a
deadlock.

Dining Philosophers, on the other hand, has a particular structure
where there is one process per philosopher, all with the same
transition graph. Optimal plans can be generated in linear time in the
number of philosophers by making a constant number of transitions to
reach the same known state in each of the graphs. Similar
considerations apply to Optical Telegraph.

PSR tasks can also be solved optimally in polynomial time, but this
requires a rather complex algorithm. All plans start with the wait
action which opens all circuit-breakers affected by a fault. In their
simplest form, optimal plans will follow by prescribing a series of
actions opening all switches connecting a feedable line to a faulty
one.  This is necessary but also sufficient to ensure that the network
is in a safe state in which no faulty line can be re-supplied. Then a
minimal set of devices (disjoint from the previous one) must be closed
so as to resupply the rest of the network. This can be achieved by
generating a minimal spanning tree for the healthy part of the
network, which can be done in polynomial time.

\scalefig{0.25}{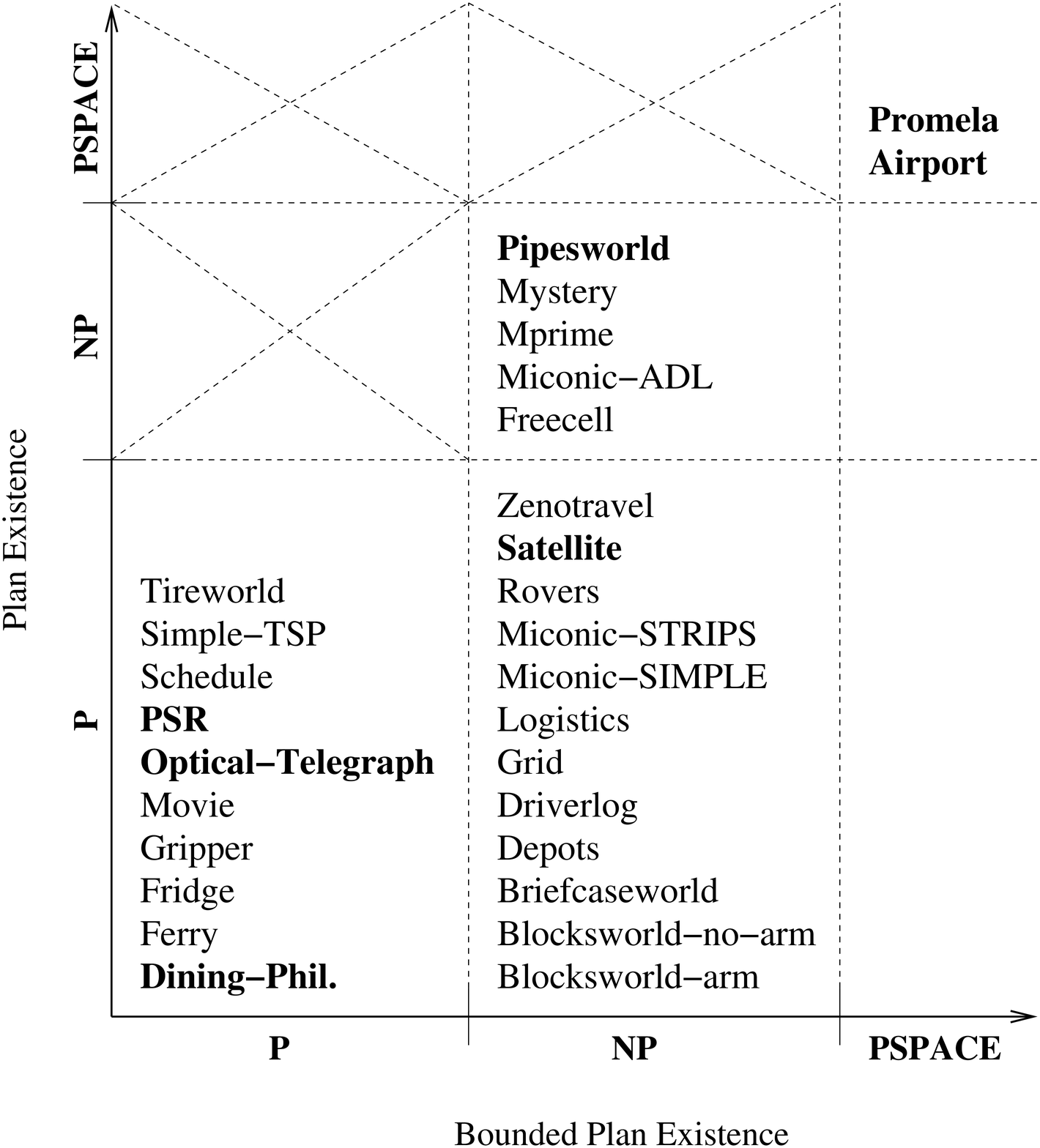}{An overview of Helmert's
  results on the computational complexity of the benchmarks.}

Figure~\ref{XFIG/com_taxonomy.eps} gives an overview of these results
and summarizes Helmert's \citeyear{helmert:ai-03} results for other
standard benchmarks.  The domain set displayed is the same set as
investigated by \citeA{hoffmann:jair-05}, with a few minor differences
explained shortly. Blocksworld-no-arm, Briefcaseworld, Ferry, Fridge,
Simple-TSP, and Tireworld are traditional planning benchmarks that
were never used in an IPC.\footnote{Blocksworld-no-arm is the version
  of Blocksworld where blocks can be moved directly to their
  destination, without referring to a robot arm. Simple-TSP was used
  by \cite{fox:long:ijcai-99} to demonstrate the potential of symmetry
  detection. One simply has to visit $n$ nodes, using a move action
  that can be applied between any two nodes, so that any permutation
  of the nodes is an optimal tour. \citeA{hoffmann:jair-05} also
  investigates the Towers of Hanoi domain.} The IPC-1 benchmarks are
Assembly, Grid, Gripper, Logistics, Movie, Mprime, and Mystery. The
IPC-2 benchmarks are Blocksworld-arm, Freecell, Logistics,
Miconic-ADL, Miconic-SIMPLE, Miconic-STRIPS (``Miconic'' is Schindler
Lift's name for the elevator domain), and Schedule. The IPC-3
benchmarks are Depots, Driverlog, Freecell, Rovers, Satellite, and
Zenotravel.  The IPC-4 benchmarks are displayed in bold face,
including the (hypothetical) general Promela domain.

The table in Figure~\ref{XFIG/com_taxonomy.eps} is organized along two
axes, where the $x$ axis shows the complexity of deciding bounded plan
existence, and the $y$ axis shows the complexity of deciding
(unbounded) plan existence. Membership in a table entry means, for the
\np\ and \pspace\ rows and columns, that the respective problem is
complete for the respective complexity class. An exception is the
Pipesworld domain, for which, as stated above, it is still unknown
whether the two decision problems are also members of \np. The
Assembly domain is not displayed since, there, \citeA{helmert:ai-03}
proved only the existence of exponentially long optimal plans, showing
that plan generation can be quite hard in the domain. The table
sectors above the diagonal are crossed out because unbounded plan
existence can be polynomially reduced to bounded plan existence --
just set the bound to $2^n$, where $n$ is the number of distinct
actions, or, in ADL, the number of distinct conditional effects.

The most striking new feature of IPC-4 is the introduction of
\pspace-complete benchmark domains, filling in the top right corner of
Figure~\ref{XFIG/com_taxonomy.eps}. Thus, the benchmarks cover all
four inhabited sectors of the table. Of the previous IPCs, each of
IPC-1 and IPC-2 cover three sectors -- all inhabited sectors except
the top right corner -- and the IPC-3 benchmarks cover only two
sectors -- namely, bounded plan existence is \np-complete for all
these domains, and all the domains except Freecell have a polynomial
time algorithm deciding unbounded plan existence.

The IPC-4 benchmarks are exceptional in further aspects not visible in
Figure~\ref{XFIG/com_taxonomy.eps}. Most particularly, as explained
above, the polynomial decision algorithm for PSR is highly
non-obvious. Such benchmarks are important since, on the one hand,
they in principle allow planners to provide efficient solutions,
while, on the other hand, necessitating that they employ interesting
techniques for doing so.\footnote{In Helmert's
  \citeyear{helmert:personnal-05} words: ``I think that domains that
  can be solved in polynomial time but where polynomial algorithms are
  not obvious are extraordinarily interesting.  Deterministic PSR
  definitely is a domain of that kind with regard to optimization.
  \np-hard problems cannot be solved without strong reliance on
  search, but polynomial problems can, if the planners capture the
  important concepts.''} Schedule is the only other polynomial
benchmark for which bounded plan generation requires a non-obvious
algorithm. For all the other 20 domains in the left bottom and middle
bottom sectors of the table, the polynomial algorithms -- deciding
bounded or unbounded plan existence -- are completely trivial, mostly
just addressing one subgoal at a time.

As was pointed out already, a final exception lies in the
extraordinarily large difference between worst-case and typical-case
behavior in Airport. As we will see in Section~\ref{new}, even fully
automated methods (the IPC-4 planners) are, at least for unbounded
plan existence (generation), quite efficient in typical instances of
this domain. While large differences between worst-case and
typical-case behavior are not unusual, we believe that the extent of
this phenomenon in Airport really is unusual. For example, planners
tend to find PSR much harder than Airport.

\subsection{The Topology of $h^+$}
\label{known:opth}

\citeA{hoffmann:jair-05} considers the state spaces (the forward
search spaces) of STRIPS and ADL tasks taken from standard benchmark
domains. He defines, given such a task and a world state $s$, $h^+(s)$
to be the length of a shortest possible relaxed plan, or $\infty$ if
there is no relaxed plan. A relaxed plan is a plan that achieves the
goal from $s$ if one assumes that the delete lists are all empty.
Computing $h^+$ (the corresponding decision problem) is {\bf NP}-hard
\cite{bylander:ai-94}. Many modern planners, e.g., HSP
\cite{bonet:geffner:ai-01}, FF \cite{hoffmann:nebel:jair-01}, SGPlan
\cite{wah:chen:ijait-04,chen:etal:ipc4booklet-04}, YAHSP
\cite{vidal:icaps-04}, and Fast-Diagonally-Downward
\cite{helmert:icaps-04,helmert:jair-06}, can be interpreted as doing
some sort of heuristic search with an {\em approximation} of $h^+$,
plus further techniques like problem decomposition
\cite{wah:chen:ijait-04}, lookahead techniques \cite{vidal:icaps-04},
and additional different heuristic functions \cite{helmert:icaps-04}.
In this context, a question of great practical interest is the quality
of the underlying heuristic function in the addressed domains.
Heuristic quality can be measured in terms of topological properties
of the search space surface: How many local minima are there? How
large are they? What about flat regions?  \citeA{hoffmann:jair-05}
investigates these questions for the $h^+$ function, for which
topological properties of the search space surface can be {\em
  proven.}

Hoffmann defines topological phenomena following
\citeA{frank:etal:jair-97}. He identifies several parameters that show
particularly interesting behavior in planning benchmarks. A {\em dead
  end} is a world state that is reachable from the initial state but
from which the goal state cannot be reached. An {\em unrecognized dead
  end} is a dead end $s$ for which $h^+(s) < \infty$. The {\em exit
  distance} from a state $s$ is the length of a shortest path in the
state space leading from $s$ to some other state $s'$, so that $h^+(s)
= h^+(s')$, and $s'$ has a direct neighbor state $s''$ with $h^+(s'')
< h^+(s')$.  That is, the exit distance from $s$ is the number of
steps we need to go from $s$ in order to find a better state ($s''$),
minus $1$ since the distance to $s'$ is measured. Here, $s'$ plays the
role of an ``exit'' state as used by \citeA{frank:etal:jair-97}. A
state lies on a {\em local minimum} if all paths to an exit have a
temporary increase in the heuristic value; otherwise the state lies on
a {\em bench}. The {\em maximal local minimum exit distance} ({\em
  mlmed}), for a state space, is the maximum over the exit distances
of all states lying on local minima in the state space.  Similarly,
the {\em maximal bench exit distance} ({\em mbed}) is the maximum over
the exit distances of all states lying on benches. The core results of
Hoffmann's \citeyear{hoffmann:jair-05} investigation are displayed in
Figure~\ref{XFIG/new_taxonomy.eps}.

\scalefig{0.25}{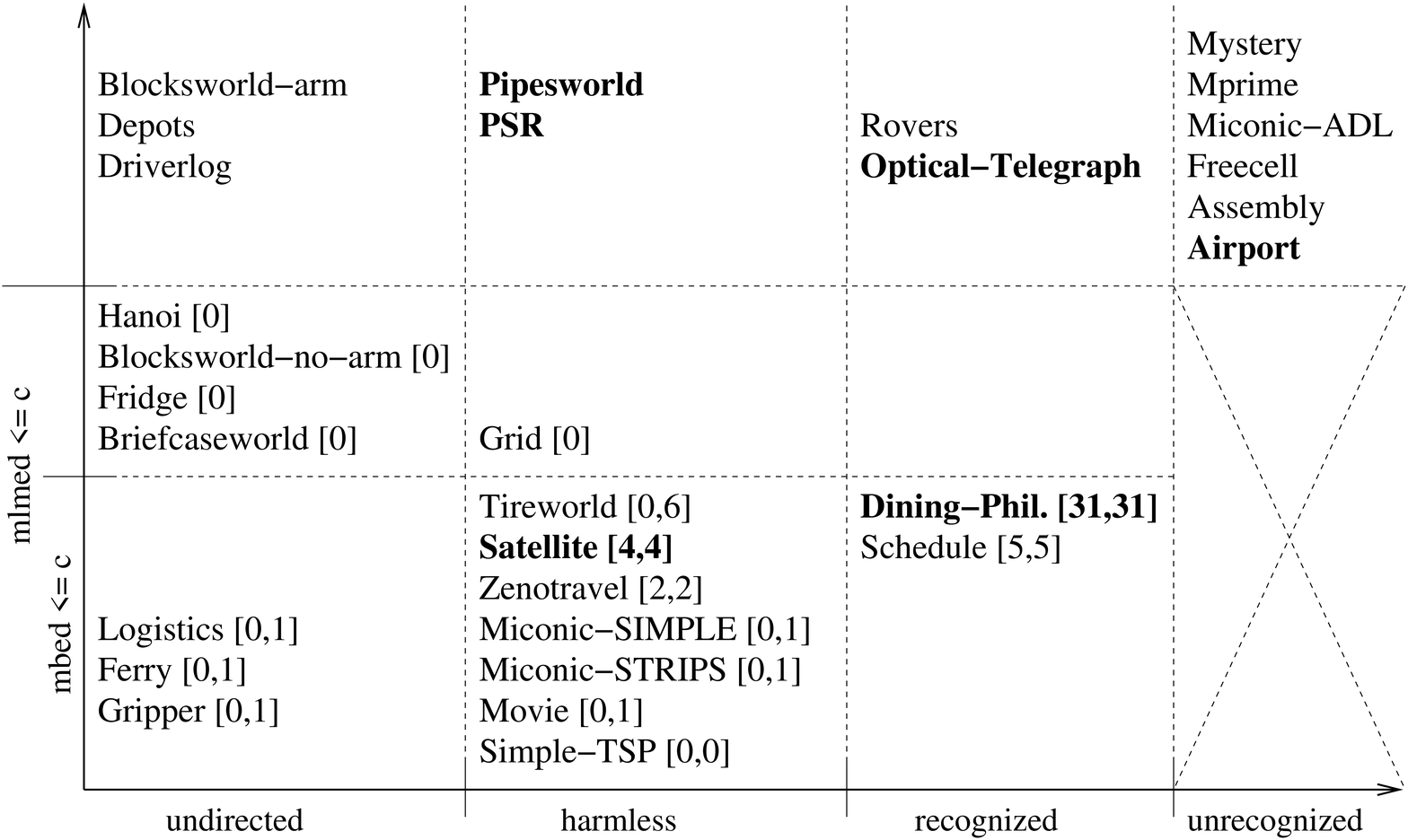}{An overview of Hoffmann's
  results on the topology of $h^+$ in the benchmarks.}

The $x$-axis in Figure~\ref{XFIG/new_taxonomy.eps} corresponds to
properties regarding dead ends. The $y$-axis corresponds to properties
regarding the exit distance from local minima and benches. The domains
are assigned to the appropriate table sectors -- classes of domains --
depending on the worst-case behavior possible in them. In more detail,
the meaning of the table is the following. A state space is
``undirected'' if every transition (action) can be directly inverted;
the state space is ``harmless'' if such an inversion is not possible,
but there are no dead ends anyway; ``recognized'' means that there are
dead ends, but $h^+$ is $\infty$ for all of them; ``unrecognized''
means that there is at least one unrecognized dead end. A domain falls
into the class of its worst-case instance: for example, if there is a
single instance whose state space contains a single unrecognized dead
end, then the domain is considered ``unrecognized''. The results are
proved, i.e., if a domain is, for example, considered ``harmless'',
then this means that provably no instance of the domain contains any
dead ends.

On the $y$-axis in Figure~\ref{XFIG/new_taxonomy.eps}, the distinction
lines correspond to the existence or non-existence of {\em constant}
upper bounds on the maximal local minimum exit distance (upper line)
and on the maximal bench exit distance (lower line). Note that
constant upper bounds on the maximal local minimum exit distance exist
in {\em all} domains below the upper line -- in the domains below the
lower line, both bounds exist.\footnote{This presentation assumes that
  the domains with bounded bench exit distance are a subset of those
  with bounded local minimum exit distance. This is not true in
  general, but does hold in all the considered benchmark domains.} By
``constant'', it is meant here that the bound is valid for every
instance of the domain, regardless of its size. The actual bounds
proved are displayed in brackets; local minimum bound precedes bench
bound in the cases where there are both. The right bottom part of the
table is crossed out since unrecognized dead ends have infinite exit
distance and so these domain classes are empty.\footnote{One could
  skip unrecognized dead ends from the definition of the maximum exit
  distances, but \citeA{hoffmann:jair-05} argues that this is
  un-intuitive, plus making things unnecessarily complicated.}

The obvious intuition behind Figure~\ref{XFIG/new_taxonomy.eps} is
that there is a transition from ``easy'' to ``hard'' -- for planning
systems based on heuristic search approximating $h^+$ -- as one moves
from the left bottom side to the top right side of the table. Indeed,
the table does, in that sense, coincide very well with the empirical
behavior of, at least, the FF system. Note how extreme the topological
behavior is in many domains. If the upper bound on the local minimum
exit distance is $0$ then this means that there are no local minima at
all. This is the case in 13 of the 30 investigated domains. In several
domains, such as the widely used Logistics benchmark, on top of that a
single step suffices to reach an exit from benches.
\citeA{hoffmann:jair-05} shows that FF would be polynomial in the
bottom classes of the table, provided with an oracle computing $h^+$.

Considering the table from a perspective of benchmark development, one
notices that particularly the older benchmarks tend to lie on the left
bottom side; consider for example Ferry, Briefcaseworld, Fridge,
Simple-TSP, and Tireworld. The distribution of the IPC-1 benchmarks --
Gripper, Logistics, Movie, Grid, Assembly, Mystery, and Mprime -- is
somewhat extreme: the first four in our list here belong to the most
simple classes, the last three belong to the hardest class (until
today, the Mystery and Mprime domains are amongst those causing
planners the most trouble). In the IPC-2 benchmarks -- Logistics,
Blocksworld-arm, Miconic-STRIPS, Miconic-SIMPLE, Schedule, Freecell,
and Miconic-ADL -- again, we have many simple and a few very
challenging domains. The most notable exceptions in that respect are
Blocksworld-arm, on the left top side of the table, and Schedule,
which does contain dead ends and local minima.  In the IPC-3
benchmarks, the distribution starts to get more varied.  The domains
-- Zenotravel, Satellite, Depots, Driverlog, Rovers, and Freecell --
span three of the four top classes in the table, plus one of the
bottom classes. The IPC-4 domains, shown in bold face, obviously
continue this development. The only two of them sharing a class are
Pipesworld and PSR.\footnote{Actually, Pipesworld is invertible in the
  sense that every two-step sequence (starting and ending a pumping
  operation) can be directly undone. It is considered ``harmless''
  here since the single actions cannot be inverted.} They continue the
emphasis on spanning the top classes in the table; the only new domain
in one of the bottom classes is Dining Philosophers, and that is
highly exceptional in that is has an exceedingly large bound, making
the bound practically useless for exploitation in
planning.\footnote{Indeed, $h^+$ is a very bad heuristic in Dining
  Philosophers. It basically comes down to counting the number of
  unsatisfied goals.} The Satellite domain adopted from the IPC-3
benchmarks serves to represent (a more interesting instance of) the
easier classes. Note that Satellite is so simple here because we are
talking about the STRIPS version, which drops the more challenging
problem constraints formulated with numeric variables. The Airport
domain is exceptional in the top right class in that, again, its
worst-case -- its place in Figure~\ref{XFIG/new_taxonomy.eps} --
differs a lot from its typical case. A dead end in Airport is a
situation where two airplanes completely block each other's
paths.\footnote{The relaxed plan can use free space in between the
  planes to make them move ``across'' each other.} Of course,
practical airports are designed in a way so that this doesn't usually
happen. As mentioned earlier, there usually are -- non-overlapping, as
far as possible -- standard routes, and the only place where blocking
can occur is in densely populated areas near parking positions.

%%% Local Variables: 
%%% mode: latex
%%% TeX-master: t
%%% End: 

\section{New (Empirical) Results on Domain Structure}
\label{new}

We now provide an empirical analysis of various structural parameters
of the IPC-4 domains. For the sake of readability and conciseness, we
focus on the non-temporal domain versions only. For most types of data
we measure, the results for the temporal domain versions are quite
similar. To some extent, this is visible in the tables showing numbers
of actions and facts, for all domain versions, in the individual
domain descriptions in Appendix~\ref{domains}.

Our empirical analysis is aimed at highlighting further
characteristics of, and differences between, the IPC-4 domains. Apart
from focussing on more practical parameters, the analysis has --
compared to the theoretical results cited in the previous section --
the big advantage that it tells us something about the actual {\em
  instances} run in the competition. Note that the choice of instances
can make a huge difference -- for example, as stated earlier, a
real-world airport is not very likely to have exponentially long
plans, and neither is it likely to provoke many dead-end situations.
Where possible at all, the instances used in IPC-4 were chosen to be
relatively realistic (details in Appendix~\ref{domains}).

The analysis is structured into three sub-sections.
Section~\ref{new:encoding} shows how, in the individual domains, the
size of the grounded encoding grows over instance size.
Section~\ref{new:heuristics} assesses the correspondence between the
quality of standard heuristic functions, and the runtime achieved in
IPC-4.  Section~\ref{new:connectivity}, finally, assesses the ``fact
connectivity'' over instance size, meaning the number of choices one
has to achieve each fact, and the number of actions a fact is required
for.

\subsection{Encoding Size}
\label{new:encoding}

\begin{figure}[htb]
\begin{tabular}{cc}
\scaleonlyfig{0.59}{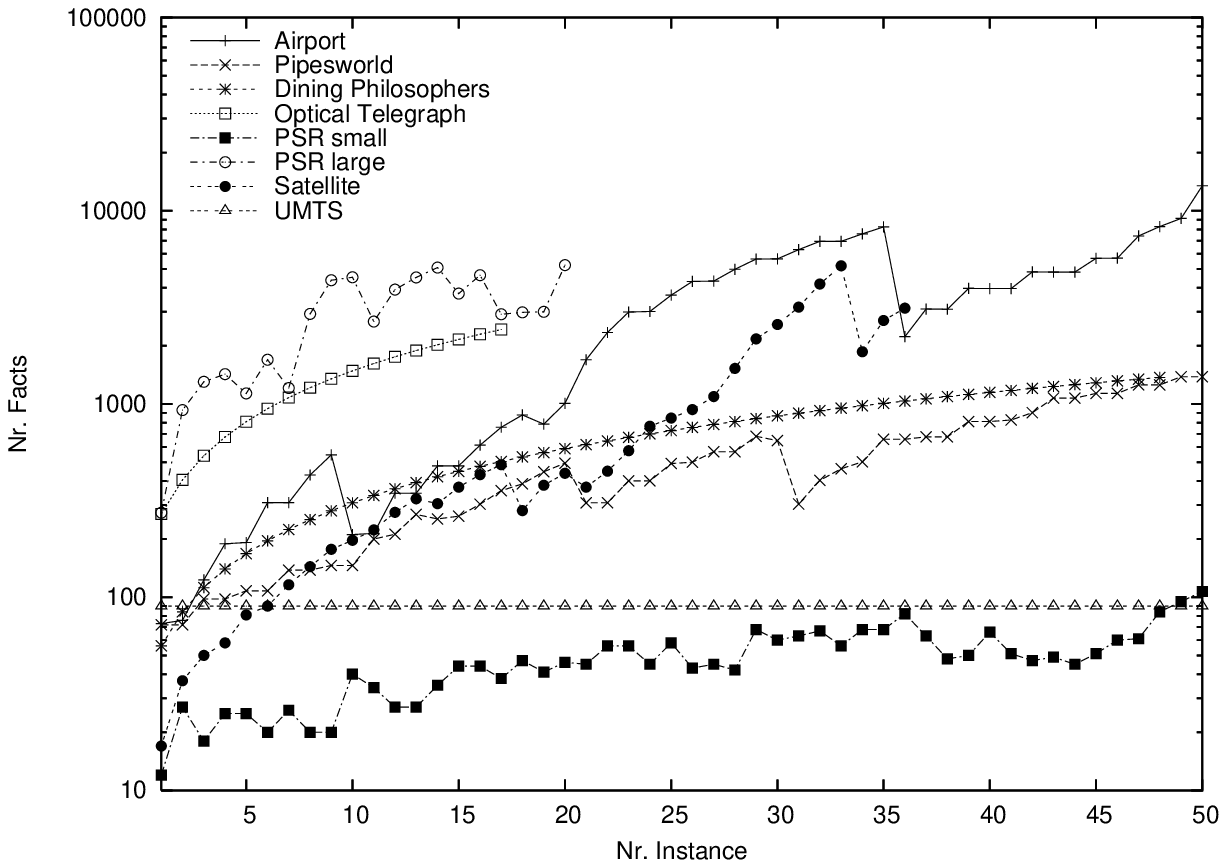} & \hspace{-0.4cm} \scaleonlyfig{0.59}{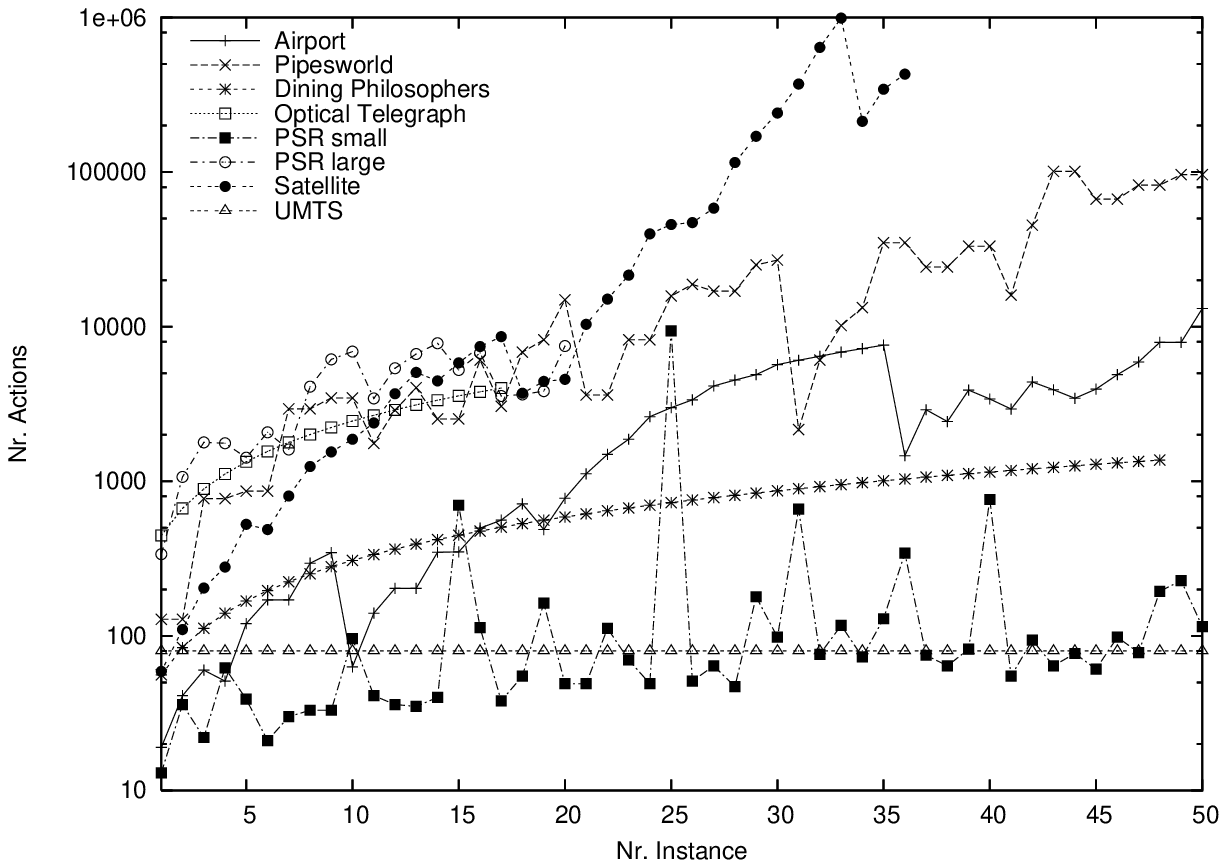}\\
(a) & (b)\\
\vspace{-0.8cm}
\end{tabular}
\caption{\label{new:encoding:data}Numbers of (a) ground facts and (b)
  ground actions, plotted over instance number, in selected versions
  of the IPC-4 domains.}
\vspace{-0.4cm}
\end{figure}

All current STRIPS and ADL planners, as far as the authors are aware,
ground all parameters and variables in a pre-process, ending up with a
task representation consisting of ground facts and ground actions. An
obvious question to ask is how large these grounded encodings are.
Figure~\ref{new:encoding:data} shows our data, numbers of facts and
actions plotted over instance size for (selected versions of) the
different domains. The numbers are measured using FF's pre-processor.
This filters out static facts -- facts that are not added or deleted
by any action -- and ``unreachable'' actions, meaning actions that do
not appear in a relaxed planning graph (a planning graph without mutex
reasoning) for the initial state \cite{hoffmann:nebel:jair-01};
formulas are compiled into simple STRIPS-like conjunctions of facts,
along the lines of \citeA{gazen:knoblock:ecp-97} as outlined in
Section~\ref{compilations}.

\vspace{0.0cm}

In all cases except UMTS (that has only temporal versions), the domain
version selected for Figure~\ref{new:encoding:data} is non-temporal.
Let us consider the domains one by one. In Airport, there is just one
non-temporal version. The plots in Figure~\ref{new:encoding:data} (a)
and (b) show us quite nicely how the instances are scaled, with sharp
drops in the curves corresponding to steps to a new underlying
airport. Precisely, instances 1 to 3, 4 to 9, 10 to 20, 21 to 35, and
36 to 50 are based on growing airports, respectively, and within each
airport the number of travelling airplanes grows from just 1 or 2
until up to 15 (in instance 50). For example, from instance 35 to
instance 36 we step from one half of Munich airport, with 12
airplanes, to the full Munich airport, with just 2 airplanes.

In Pipesworld, there are two non-temporal versions, with and without
tankage restrictions. Figure~\ref{new:encoding:data} shows data for
the former, which is the more challenging one (the IPC-4 planners
fared much worse on it); without tankage restrictions, there are
slightly fewer facts, and about a factor of 5-10 fewer actions. The
Pipesworld instances are scaled in a similar way as the Airport ones:
five growing pipeline networks each feature a growing number of
travelling liquid ``batches''. The networks underlie the instances 1
to 10, 11 to 20, 21 to 30, 31 to 40, and 41 to 50, respectively.
Corresponding drops can be observed when stepping from instance 30 to
31, and, less significantly, when stepping from 20 to 21 or from 40 to
41. A major difference to Airport is visible in the much more crippled
nature (featuring much more variance) of the curve for the number of
actions. This is because, in Airport, few objects move on a big
spacious structure, while, in Pipesworld, many objects move within a
rather dense space.\footnote{How much the objects can or cannot move
  affects also the number of ground actions due to the mentioned
  filtering of ``unreachable'' actions.} This fundamental difference
between Airport and Pipesworld also manifests itself in that the order
of curves is reversed for the numbers of facts and actions: in
Airport, extraordinarily many facts are required to describe the huge
airport structure, while in Pipesworld there are fewer facts for a
smaller structure, but many more actions describing how things move
along that structure. As stated earlier, in Pipesworld, different
objects affect each other's {\em position} when moving.

In the Promela domains, Dining Philosophers and Optical Telegraph, the
data for the domain versions with and without derived predicates are
identical, if a derivation rule deriving a fact is counted as an
action achieving the fact. The main difference to what we have seen
before lies in the extremely smooth scaling. Both domains have just a
single size parameter, and the numbers of ground facts and actions
grow as linear functions in that parameter -- the functions for
Optical Telegraph being about an order of magnitude higher than those
for Dining Philosophers. The curves for Optical Telegraph stop at
instance 17 because after that we were not able to compute the
grounded representation -- too much time and memory were needed in the
simplification of precondition formulas. Note that this is not an
artifact of our data presentation, but rather constitutes a serious
limitation to any planner that tries to perform such pre-processing.

In PSR, the most interesting domain versions are ``small'', since that
could be formulated in STRIPS, and ``large'', since that goes up to
instances of a realistic size (in the largest instances, that is). As
the name ``small'' suggests, the numbers are quite small -- to be able
to compile into STRIPS, as indicated earlier we had to make the
instances {\em very} small.\footnote{The only notable exception is
  instance nr. 25, where the number of actions peaks to 9400. This is
  due to an exceedingly complex goal formula, with 9216 disjuncts in
  its DNF, of which each yields an extra goal-achievement action,
  c.f.\ Section~\ref{compilations}.} Essentially the same compilation
problem is also visible in the curves for ``large'', that have a huge
number of ground facts and actions in relatively early instances
already. The curves stop at instance 20 because beyond that,
simplifying formulas becomes extremely costly.  In both versions, we
note a high degree of variance both in the numbers of facts and
actions, which somewhat corresponds to the huge degree of variance to
be observed for planner performance in this domain (see
Figure~\ref{new:heuristics:psr}). Part of this variance, at least the
pace of the oscillations if not their amplitude, can be explained by
the way the instances are scaled.  For a given number of sources (the
instance size), we generated instances with an increasing minimal
number of switches originally fed by a given source, and for a given
number of switches, we generated instances with an increasing
percentage of faulty lines ranging from 10\% to 70\%. Intuitively, the
larger the number of switches per source, the larger and harder we
expect the instance to be.  Furthermore, the percentage of faulty
lines tends to induce an easy-hard-easy pattern. If most lines are
faulty, only a small part of the network can be resupplied and only a
few devices need to be switched. Similarly, if a very few faulty lines
exist, most of the network can be resupplied with a few switching
operations. With an intermediate percentage, the effects of the
actions become more complex -- they are conditioned on the positions
of many other switches -- and so the instances become critically
constrained and harder to solve.

In Satellite, the main observation to be made is the extremely steep
ascent of the curves after instance 20, particularly the growth to
extremely high numbers of actions. There are two reasons for this.
First, one action in Satellite (take-image) has 4 parameters and is
``reachable'' for almost any combination of objects with the correct
types (most of the time, actions have only 2 or 3 parameters). Second,
the size of the instances themselves grows very sharply beyond
instance 20 -- which, simply, is because instances 21 to 36, as used
in IPC-4, correspond to the 16 instances posed in IPC-3 to challenge
the {\em hand-tailored} planners.

We do not consider Settlers here to ease readability of the graphs,
and since that domain is quite obviously exceptional anyway, in that
it relies almost completely on numeric variables. For UMTS,
Figure~\ref{new:encoding:data} shows data for the plain domain version
without time windows and flaw action. The obvious characteristic is
that the numbers of facts and actions are {\em constants}. This is
true for all domain versions, the numbers vary only slightly. The
reason is that, the way the UMTS instances are scaled, every instance
describes the same applications and requirements; what changes is
(only) the {\em goal}, specifying what applications actually need to
be set up.  Independent of this effect of the particular scaling
method used, we can observe that the numbers of facts and actions are
relatively low -- around only 100 even in the largest instances, where
all the applications must be set up, and the plans contain all the
actions.

\subsection{Quality of Heuristics, and Runtime}
\label{new:heuristics}

In this section, we measure the length of the best (sequential and
parallel) plans found by any planner, the (sequential and parallel)
plan length estimates returned by the most common heuristic functions,
and the runtime taken by the planners. Precisely, for the optimal
planners, we measure:
\begin{itemize}
\item The optimal makespan, as found by the IPC-4 parallel optimal
  planners (planners optimizing makespan).
\item The length of a standard plan graph \cite{blum:furst:ai-97},
  i.e., the index of the first plan graph layer that contains the
  goals without mutexes.
\item The best runtime taken by any parallel optimal planner in IPC-4.
\item The optimal sequential plan length, as found by the IPC-4
  sequential optimal planners.
\item The length of a serialized plan graph, where any pair of
  non-NOOP actions is made mutex.
\item The best runtime taken by any sequential optimal planner in
  IPC-4.
\end{itemize}
For the satisficing planners,  we measure:
\begin{itemize}
\item The best (shortest) plan length, as found by any planner in
  IPC-4.
\item The length of a relaxed plan for the initial state (an action
  sequence that solves the task if one assumes all delete lists are
  empty; computed with FF \cite{hoffmann:nebel:jair-01}).
\item The best runtime taken by any satisficing planner in IPC-4.
\end{itemize}
Our main goal will be to identify characteristic behavior of domains,
and to identify characteristic effects of heuristic quality on
performance. The reader will note that, in our selection of
measurements, we make several simplifying assumptions. Optimal
planners are not exclusively based on plan graph estimates.
Satisficing planners are not exclusively based on relaxed plan
estimates.  Further, some of the satisficing planners minimize
makespan, not sequential plan length. We chose to not take account of
the latter since there is no potentially over-estimating
(non-admissible) heuristic specifically estimating parallel plan
length; to the best of our knowledge, all satisficing planners
minimizing makespan actually use a heuristic estimating the number of
remaining actions, and employ some method to greedily arrange the
chosen actions as a parallel plan.  That said, we do not wish to imply
that our simplifying assumptions are safe in the sense that we do not
lose important information. The simplifying assumptions are necessary
to make the analysis and its presentation feasible. The data we show
definitely do capture many crucial aspects of IPC-4 heuristic quality
and planner runtime. We show data for the individual domains,
proceeding in alphabetical order. The (IPC-4) runtime results were
obtained on a Linux machine running two Pentium-4 CPUs at 3GHz, with 6
GB main memory; time and memory cutoffs were 30 minutes and 1 GB, per
instance.

\vspace{0.0cm}
\begin{figure}[thb]
\begin{tabular}{cc}
\scaleonlyfig{0.58}{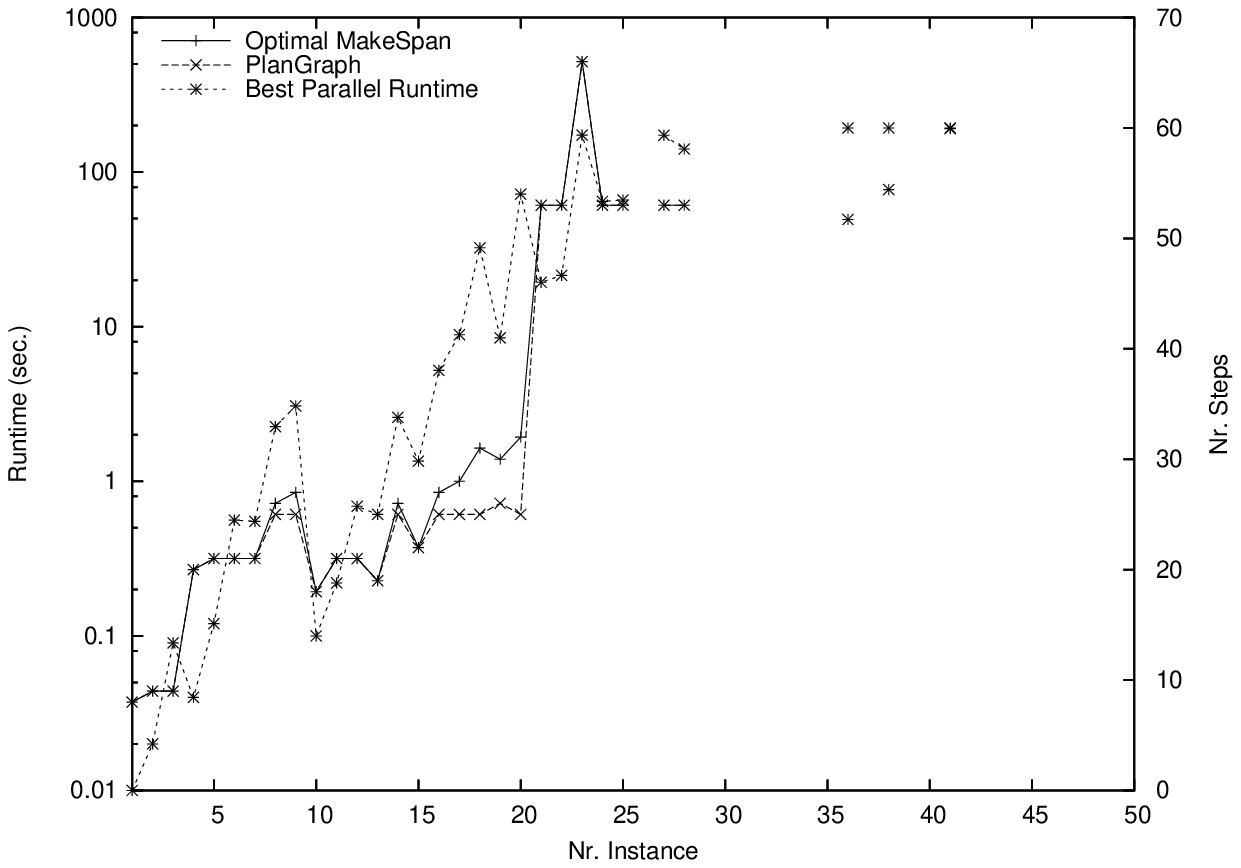} & \hspace{-0.3cm} \scaleonlyfig{0.58}{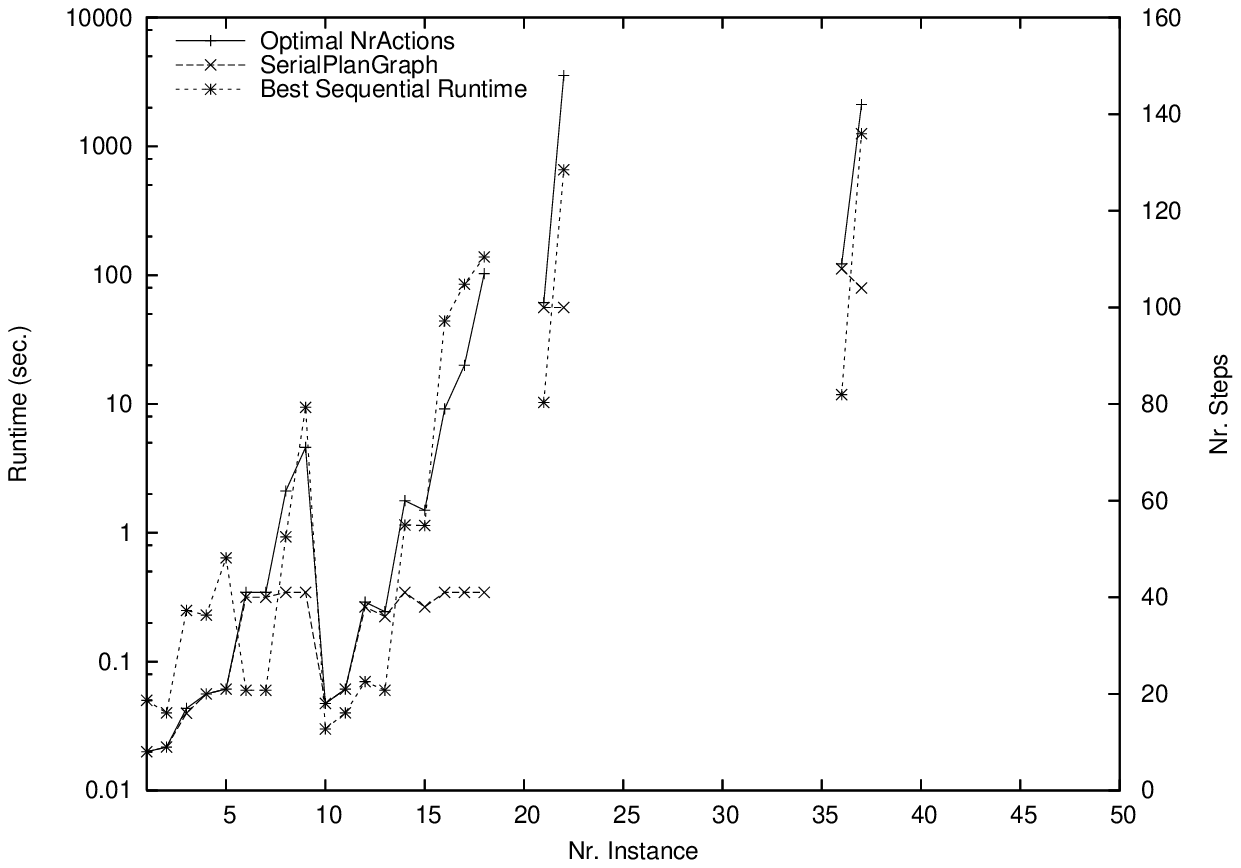}\\
(a) & (b)\\
\end{tabular}
\begin{center}
\scaleonlyfig{0.58}{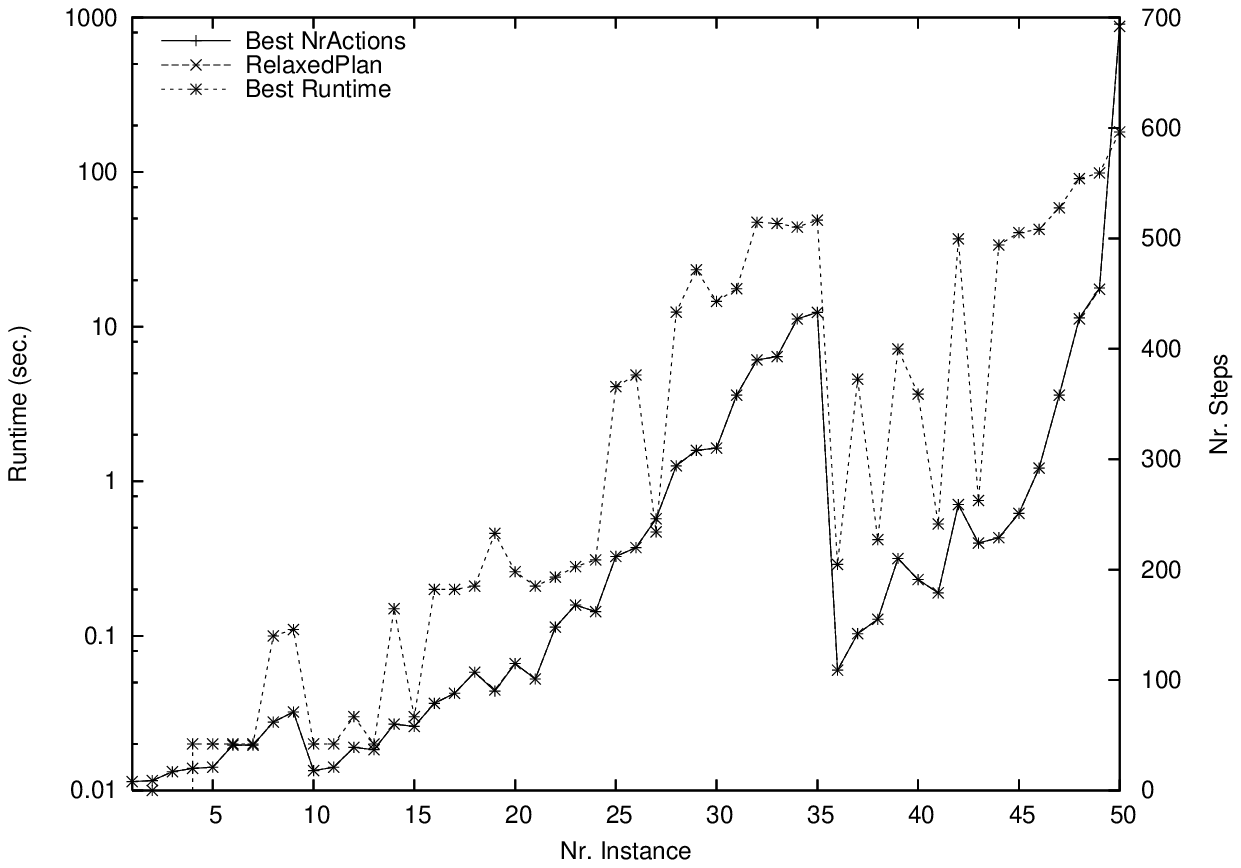}\\
(c)
\end{center}
\vspace{-0.5cm}
\caption{\label{new:heuristics:airport}Airport domain. Plots of
  (parallel) plan length, its heuristic estimation, and runtime, for
  (a) optimal parallel planners, (b) optimal sequential planners, and
  (c) satisficing planners.}
\vspace{-0.4cm}
\end{figure}

Consider Figure~\ref{new:heuristics:airport}, showing data for the
Airport domain. Note that the $y$ axis has two different meanings,
runtime on the left hand side, and number of (parallel or sequential)
plan steps on the right hand side. The same applies to all figures
below in this sub-section. For Airport, we observe a clear correlation
between quality of plan length estimation, and runtime. For the
optimal parallel planners, Figure~\ref{new:heuristics:airport} (a),
this is best observed between instances nr.  15 and 20. There, the
difference between makespan and its estimate by the plan graph grows,
and with it grows the achieved runtime, on an exponential scale. It
may look like a counter example that, for instance nr. 20, where the
plan graph estimate is exact (coincides with the real makespan), the
runtime does not get lower again. Note however, that instance 20 is
based on a much larger airport than the previous instances. From
instance 20 onwards, the only instances solved by any parallel planner
have an exact plan graph estimate. For the optimal sequential
planners, Figure~\ref{new:heuristics:airport} (b), we get a similar
behavior between instances nr.  14 and 18.  The behavior is also very
strong in instances nr. 35 and 36: while the plan length grows a lot
from 35 to 36, the serial plan graph becomes a little shorter;
correspondingly, the runtime goes up by two orders of magnitude. The
same is true for instances 20 and 21.

For the satisficing planners, in Figure~\ref{new:heuristics:airport}
(c), the most striking observation is that the length of the real plan
coincides, in all instances, {\em exactly} with the length of the
relaxed plan (for the respective initial state). This is actually
quite easy to explain: an optimal plan moves the airplanes in a way so
that they never block their paths; the same plan is optimal even when
ignoring the delete lists. Moving the airplanes without blocking is
always possible at the start. The situation changes only when a wrong
decision was made, so that additional moves have become necessary --
in reality, but not without delete lists -- to avoid a blocking
situation. Apart from this, Figure~\ref{new:heuristics:airport} shows
quite nicely that the runtime taken corresponds very closely to the
length of the plan found. Note that the latter is huge, 694 in the
largest instance.

\vspace{0.0cm}
\begin{figure}[htb]
\begin{tabular}{cc}
\scaleonlyfig{0.58}{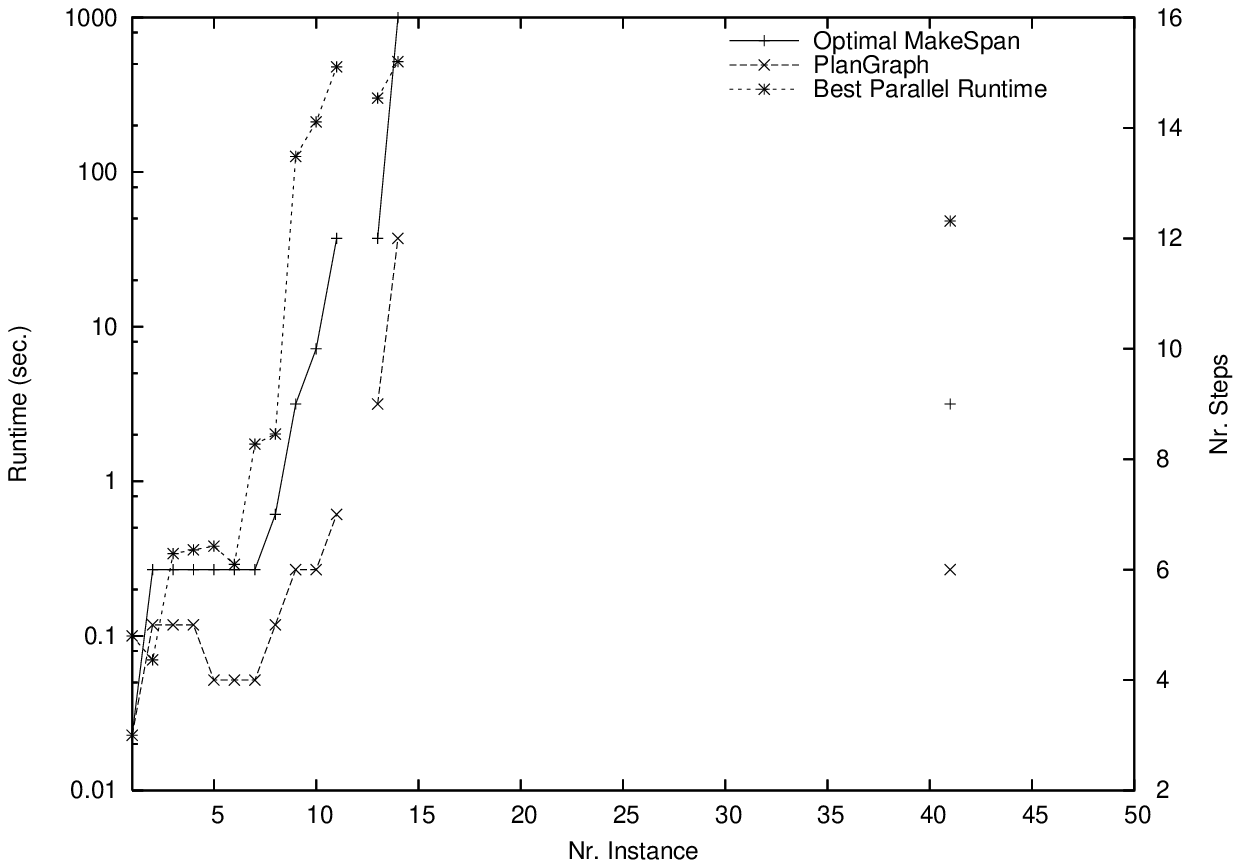} & \hspace{-0.3cm} \scaleonlyfig{0.58}{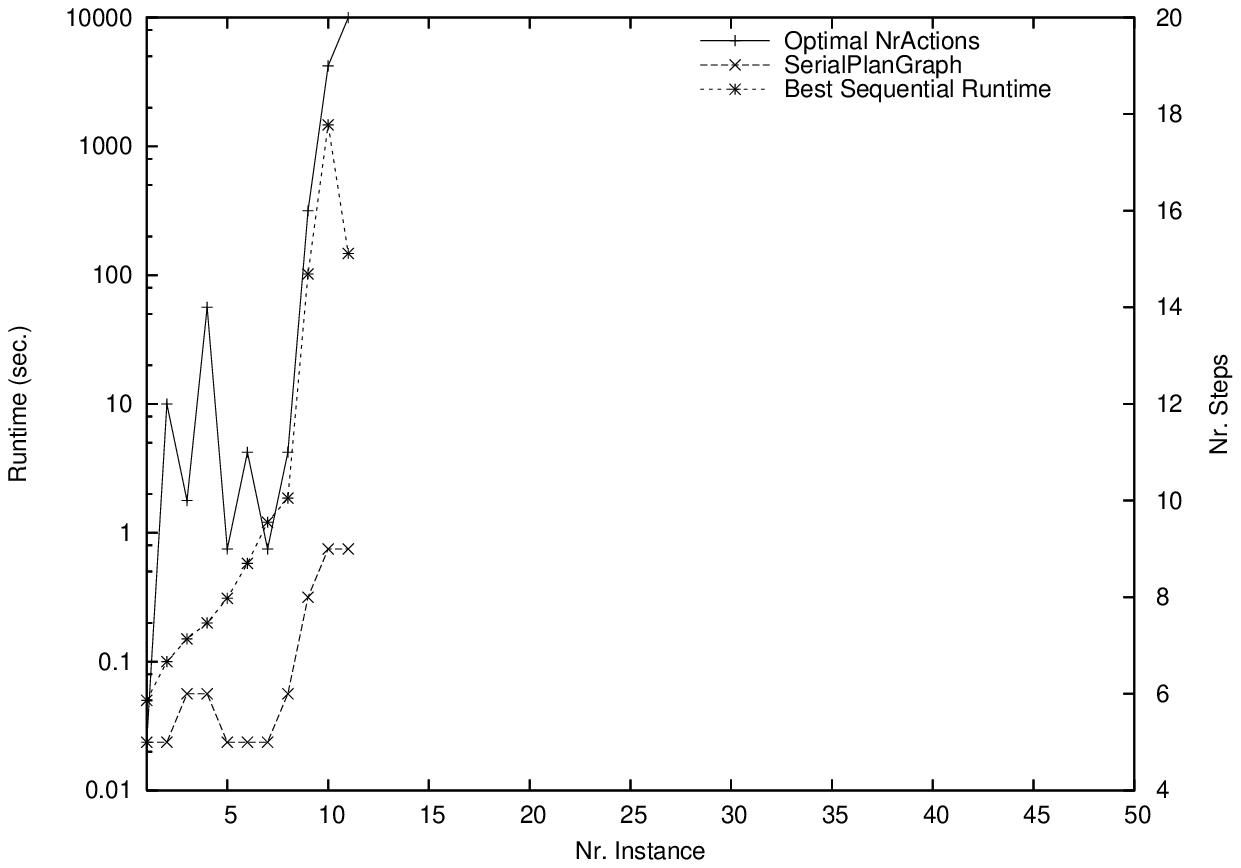}\\
(a) & (b)\\
\end{tabular}
\begin{center}
\scaleonlyfig{0.58}{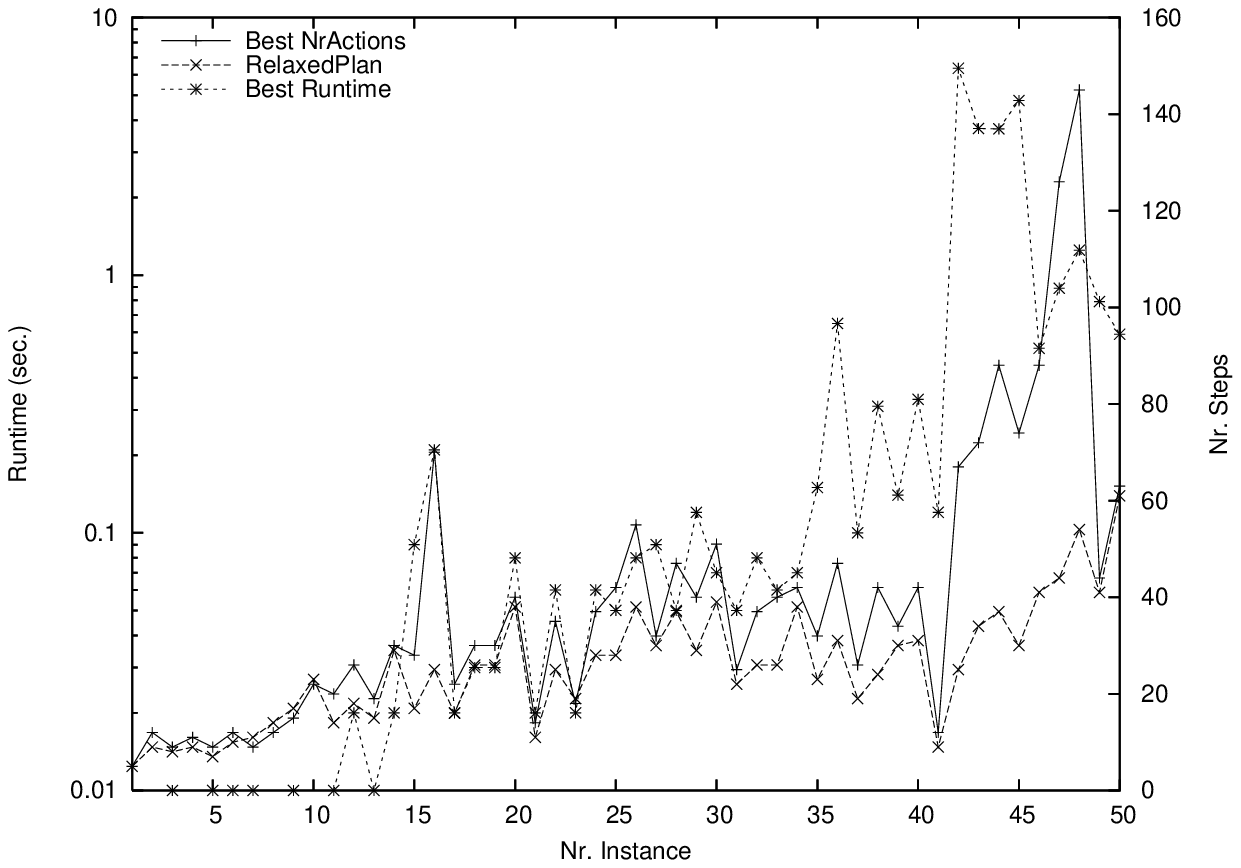}\\
(c)
\end{center}
\vspace{-0.5cm}
\caption{\label{new:heuristics:pipesworld}Pipesworld domain without
  tankage restrictions.  Plots of (parallel) plan length, its
  heuristic estimation, and runtime, for (a) optimal parallel
  planners, (b) optimal sequential planners, and (c) satisficing
  planners.}
\vspace{-0.4cm}
\end{figure}

In the Pipesworld domain, there are two non-temporal domain versions:
with/without tankage restrictions, i.e., restrictions on the amount of
liquid that can be stored in any of the network areas.
Figure~\ref{new:heuristics:pipesworld} shows our data for the version
without such restrictions; the observations to be made in the other
domain version are similar, except that both sorts of planners scale
much worse, thus providing us with less data. For the optimal
planners, Figure~\ref{new:heuristics:pipesworld} (a) and (b), the most
striking difference with the Airport domain in
Figure~\ref{new:heuristics:airport} (a) and (b) is that the quality of
even the parallel plan graph heuristic is very bad: it underestimates
the real makespan to a much larger extent than it does in Airport. The
underestimation grows with instance size, and, naturally, the runtime
grows as well. Note that the planners fail to scale much earlier than
in Figure~\ref{new:heuristics:airport} (a) and (b). There is one
slight exception to the rule that a poorer heuristic estimate leads to
a longer runtime: from instance number 10 to 11, the optimal sequential
plan length grows from 19 to 20, the length of the serial plan graph
remains 9, and the runtime drops from 1400 to 150 secs.

Similarly to the situation for the optimal planners, for the
satisficing planners, Figure~\ref{new:heuristics:pipesworld} (c), the
main difference from Figure~\ref{new:heuristics:airport} (c) is the
much worse quality of the heuristic function: the relaxed plan length
now differs greatly from the length of the real plans found,
particularly for the larger instances. Very curiously, despite the
worse quality of the heuristic, the runtimes are {\em much} lower.
The longest time taken for any instance is below 10 seconds. This goes
to show, first, the shortcomings of our analysis here: we give the
heuristic quality only for the initial state, which may differ a lot
from the situation in the rest of the state space. For example, in
Airport a planner using relaxed plans may get lost in huge dead ends
when a wrong decision was made early on.  Second, of course, other
techniques that the satisficing planners use are also relevant. The
runtime data in Figure~\ref{new:heuristics:airport} (b) are
exclusively due to SGPlan \cite{wah:chen:ijait-04} and YAHSP
\cite{vidal:icaps-04}, whose problem decomposition/greedy lookahead
techniques appear to work extremely well in this domain. All other
satisficing planners perform much worse, failing to solve the largest
instances. We note that in Pipesworld, the overall runtime curves (for
all planners) are characteristically very jagged and show considerable
variance in comparison to, e.g., Airport.  This information gets lost
in the best-of presentation chosen for our figures here. It seems to
be that hardness in this domain comes from interactions too subtle to
be seen with the rather high-level parameters measured here.  We
re-iterate that the domain version {\em with} tankage restrictions is
much more challenging to the planners, the only planner getting
anywhere close to the largest instances being YAHSP.

\vspace{0.0cm}
\begin{figure}[htb]
\begin{tabular}{cc}
\scaleonlyfig{0.58}{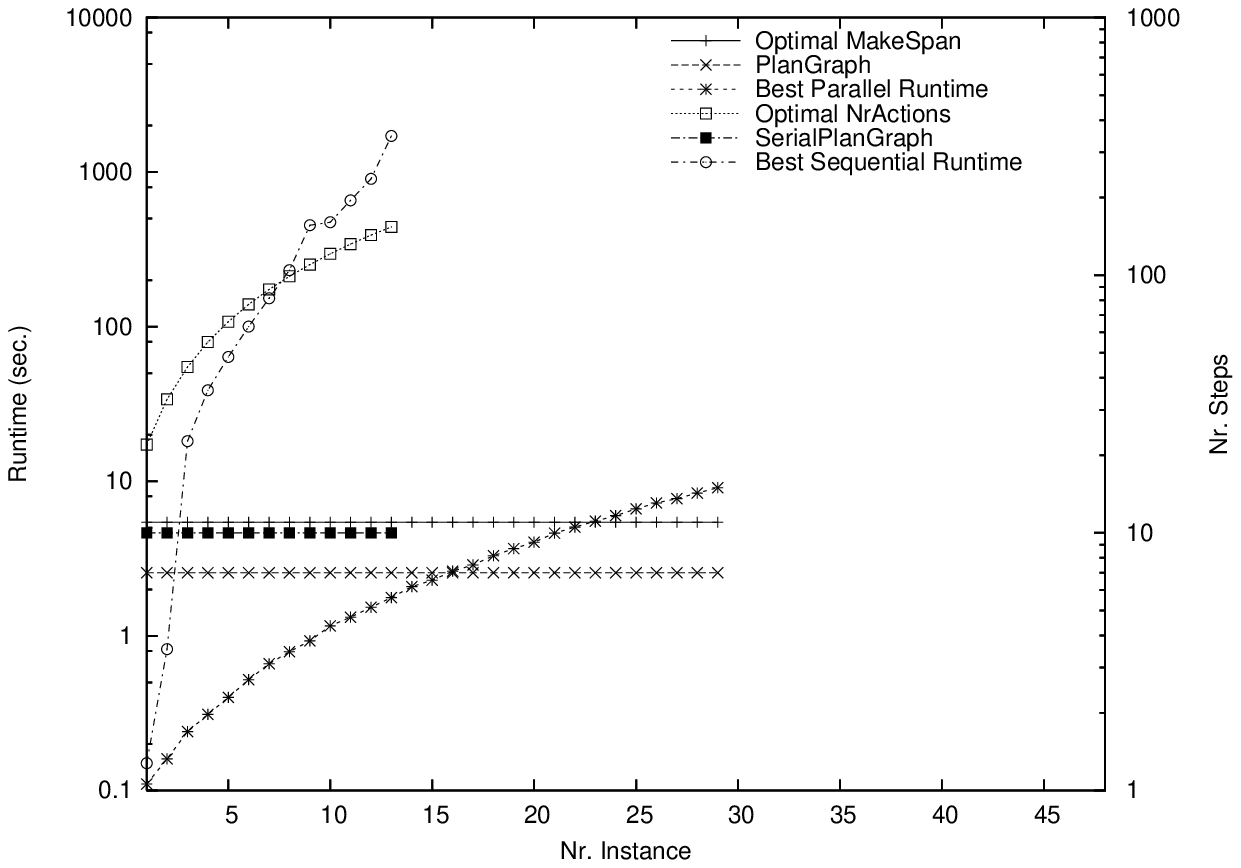} & \hspace{-0.3cm} \scaleonlyfig{0.58}{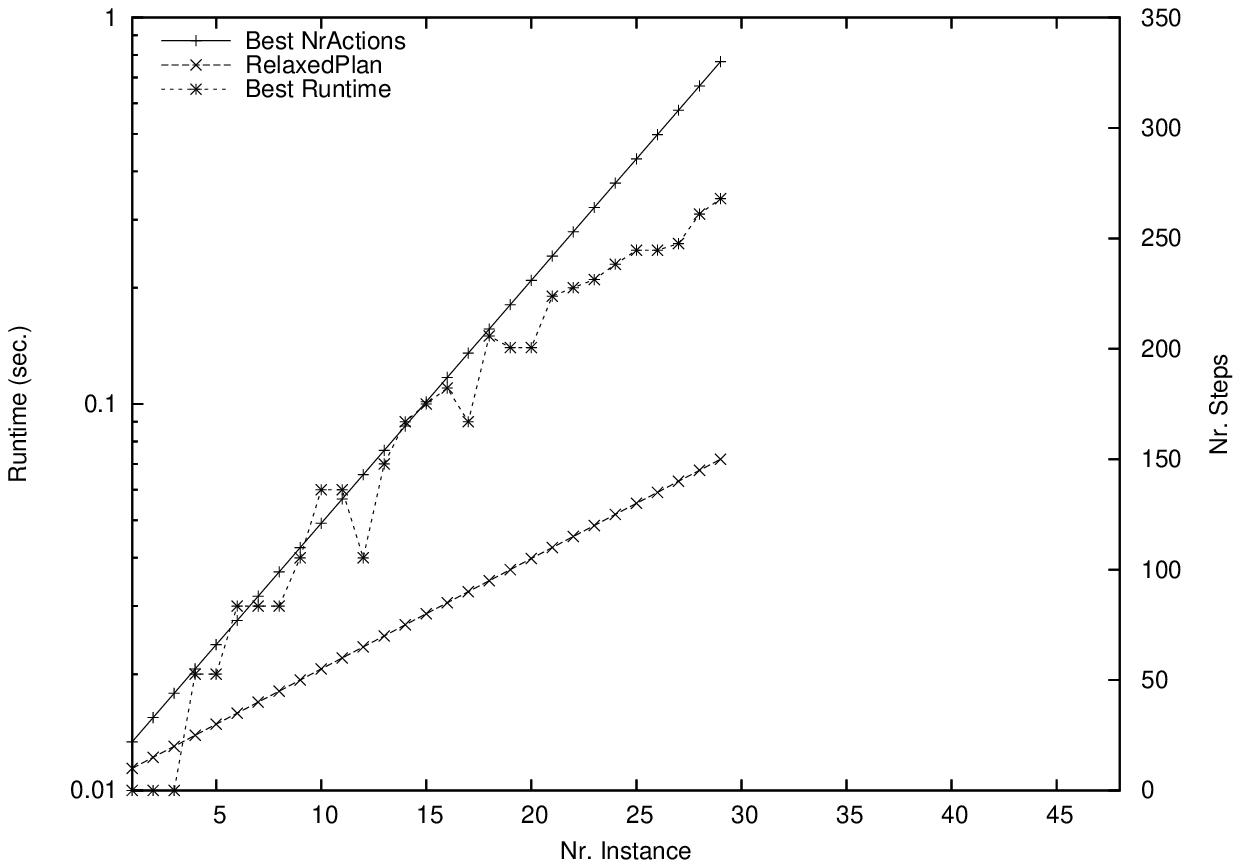}\\
(a) & (b)\\
\vspace{-0.8cm}
\end{tabular}
\caption{\label{new:heuristics:dining}Dining Philosophers domain
  without derived predicates.  Plots of (parallel) plan length, its
  heuristic estimation, and runtime, for (a) optimal planners and (b)
  satisficing planners.}
\vspace{-0.4cm}
\end{figure}

Figure~\ref{new:heuristics:dining} shows our data for Promela/Dining
Philosophers without derived predicates. We do not show two separate
figures for the optimal planners since the curves are quite easy to
read. From even a quick glance, one sees that the domain has a very
characteristic behavior different from the other domains. The optimal
makespan, plan graph length, and serial plan graph length are all
constant across instance size. In contrast, the optimal sequential
plan length grows as a linear function of size; note the logarithmic
scale of the right hand side $y$ axis in
Figure~\ref{new:heuristics:dining} (a), which we had to use to make
the figure (the values of the other plan step measures) readable. The
best plans found by the satisficing planners are optimal, i.e., the
NrActions data are identical on both sides of the figure. In
Figure~\ref{new:heuristics:dining} (a), we once again see the effect
of heuristic quality on search performance: the parallel planners
scale as a linear function in instance size, while the sequential
planners, for whom the heuristic function becomes worse and worse,
scale highly exponentially. The latter might also be true for the
satisficing planners; it is a bit hard to tell since the solved
instances are solved extremely quickly. The reason why no instance
with index higher than 29 is solved is that, for these instances,
similarly to what we discussed above (Section~\ref{new:encoding}),
simplifying precondition formulas became prohibitively costly, so
these instances were available in ADL only. The only two satisficing
planners that scaled well in Dining Philosophers (without derived
predicates) were SGPlan and YAHSP -- neither of which could handle the
ADL formulation of the domain. Similarly, from the optimal planners
only SATPLAN'04 and Optiplan scaled well, and neither could handle the
ADL formulation. Note that the inability of planners to handle
formulas without pre-simplification techniques thus constitutes a
serious limitation.

In Optical Telegraph without derived predicates (no figure shown) the
observations are similar to the ones in
Figure~\ref{new:heuristics:dining}, except that the planners scale
much worse. Most particularly, the optimal sequential planners solve
only the single smallest instance, and the best satisficing runtime is
clearly exponential in instance size, taking over 1500 seconds to
solve instance number 25. In the Promela domain versions {\em with}
derived predicates, there are no results for optimal planners since
none of them could handle derived predicates.  The observations for
the satisficing planners are similar to the above: NrActions grows as
a linear function of instance size, relaxed plan length grows as a
linear function with significantly lower gradient. The planners are
very fast in Dining Philosophers but need a lot of time ($>$ 1000 sec)
to solve the largest Optical Telegraph instances (some of which remain
unsolved).  We omit the results for the Promela domain versions using
numeric variables, since only two planners participated in these
domain versions.

\vspace{0.0cm}
\begin{figure}[htb]
\begin{tabular}{cc}
\scaleonlyfig{0.58}{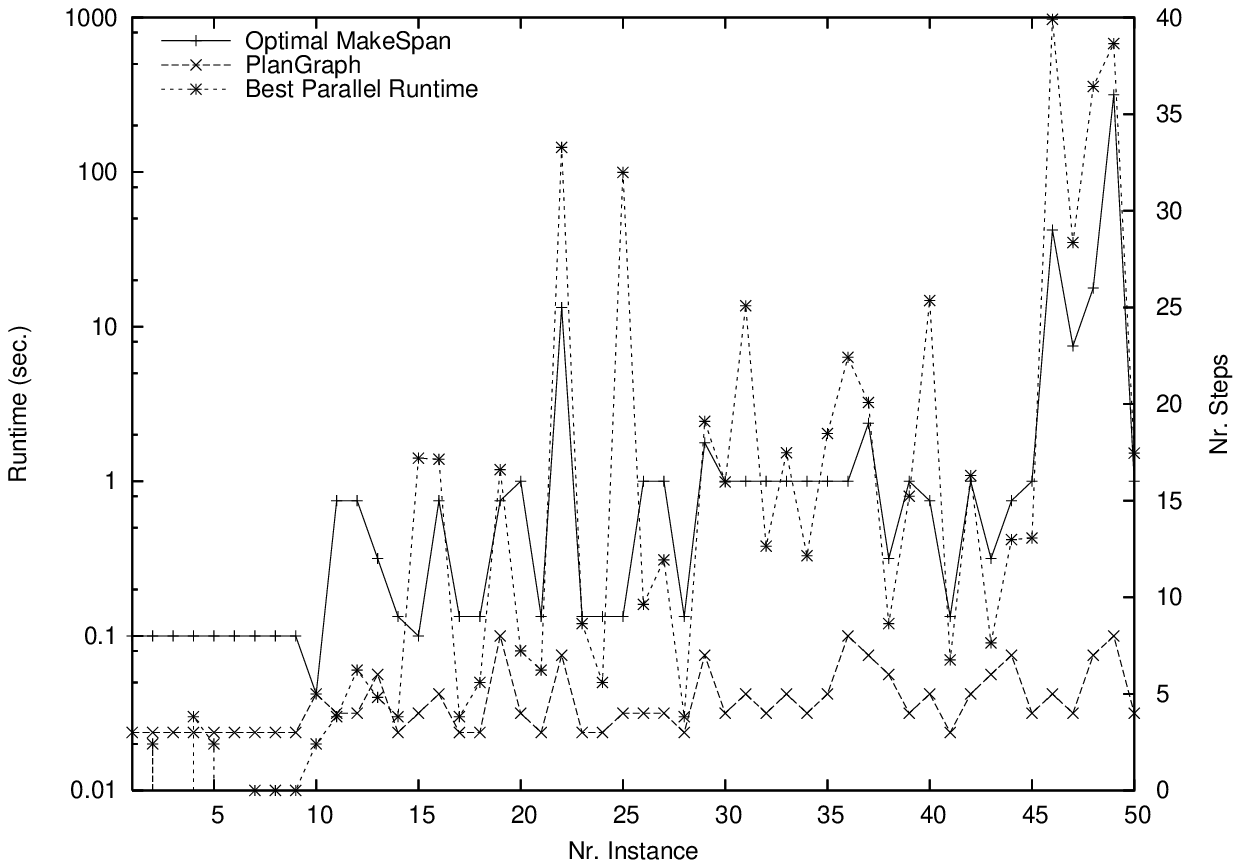} & \hspace{-0.3cm} \scaleonlyfig{0.58}{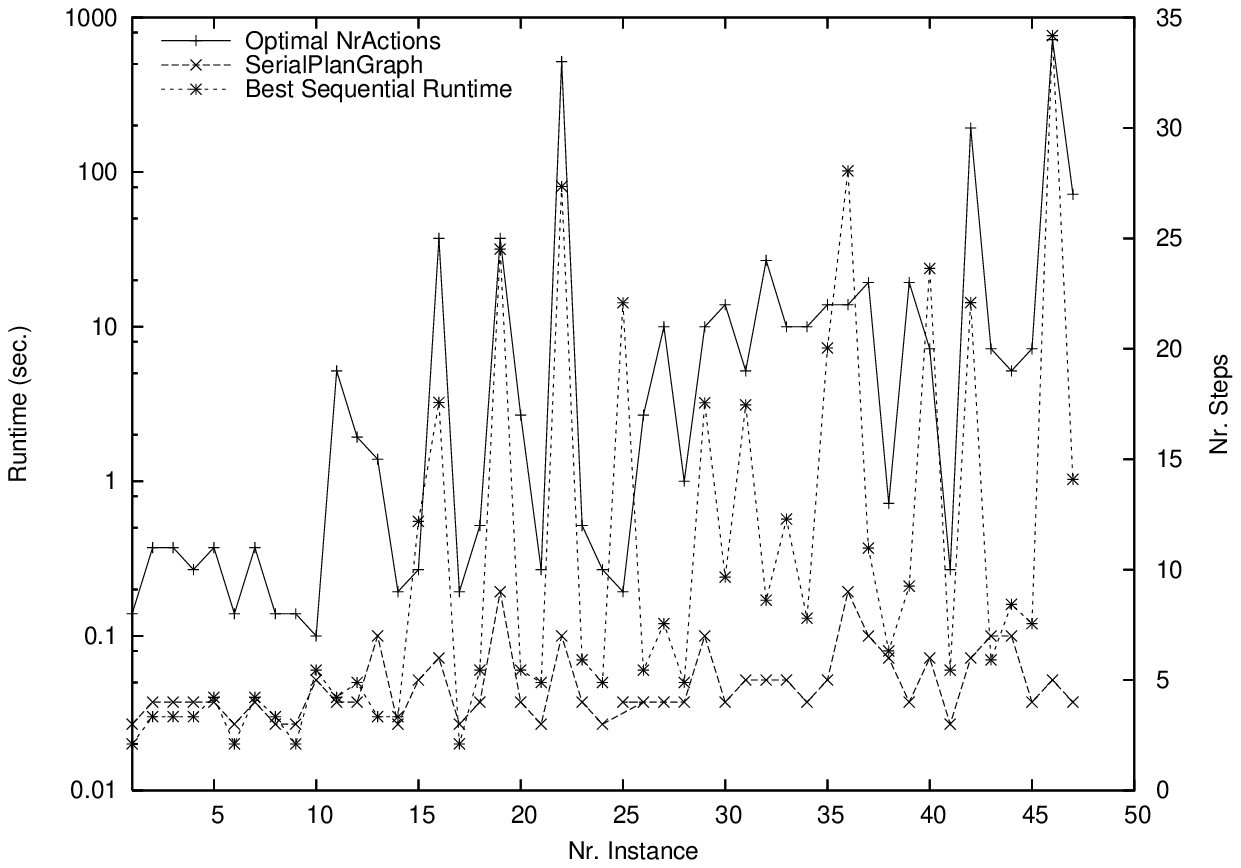}\\
(a) & (b)\\
\scaleonlyfig{0.58}{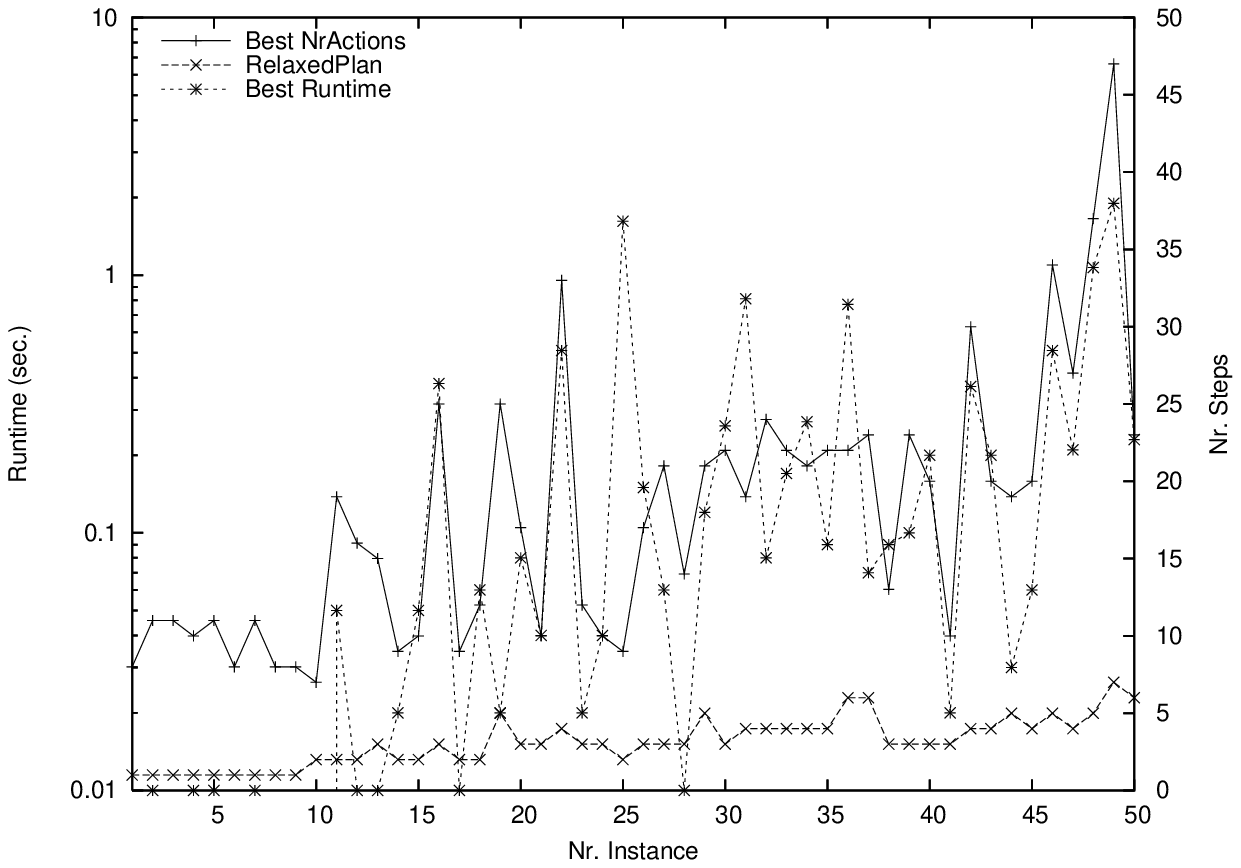} & \hspace{-0.3cm} \scaleonlyfig{0.58}{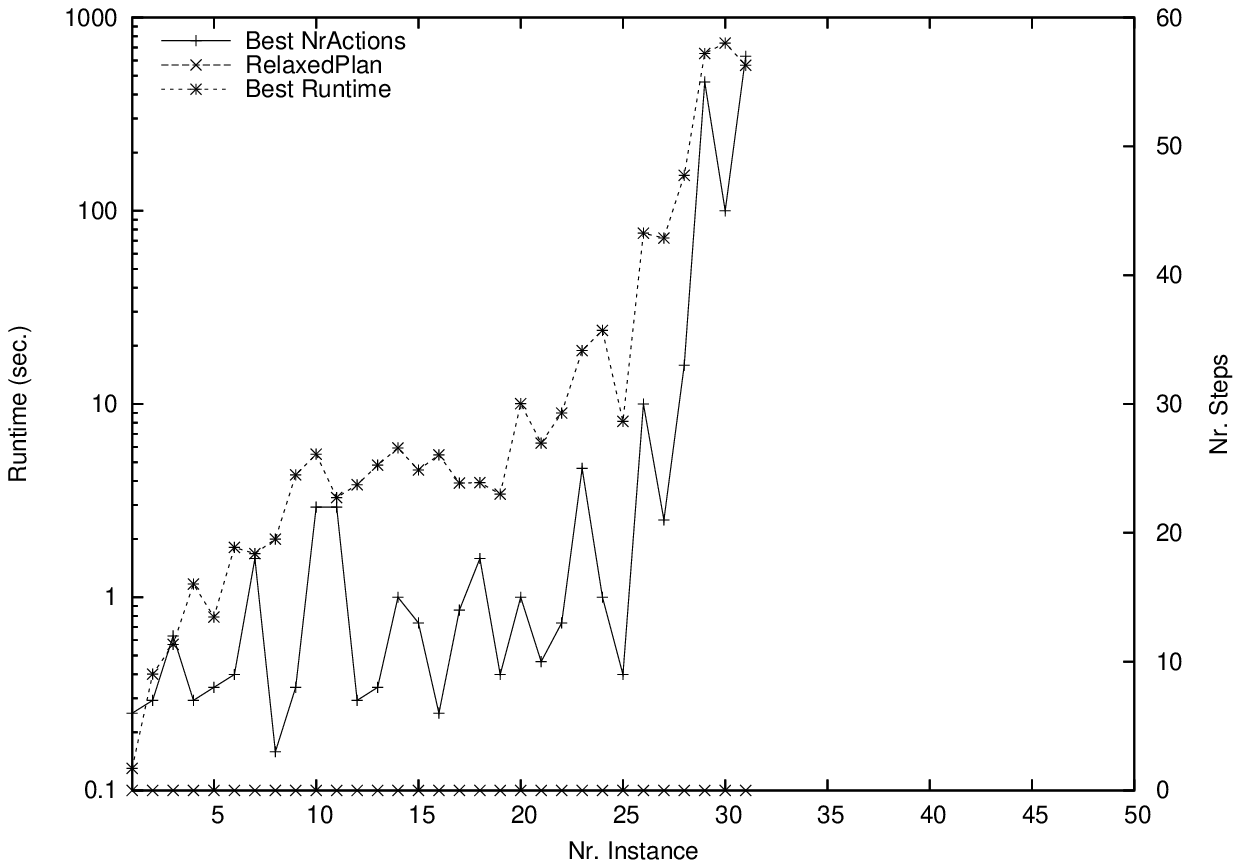}\\
(c) & (d)\\
\vspace{-0.8cm}
\end{tabular}
\caption{\label{new:heuristics:psr}PSR domain.  Plots of (parallel)
  plan length, its heuristic estimation, and runtime, for (a) parallel
  optimal planners in PSR ``small'' (STRIPS version), (b) sequential
  optimal planners in PSR ``small'', (c) satisficing planners in PSR
  ``small'', and (d) satisficing planners in PSR ``large'' (featuring
  ADL and derived predicates).}
\vspace{-0.4cm}
\end{figure}

Figure~\ref{new:heuristics:psr} shows our results for the PSR domain.
Figure~\ref{new:heuristics:psr} (a), (b) and (c) show plots for the
domain version PSR ``small'', which comes in pure STRIPS and was
addressed by all IPC-4 planners; Figure~\ref{new:heuristics:psr} (d)
shows plots for PSR ``large'', which comes in ADL with derived
predicates and was addressed by four satisficing planners only. We do
not show data for PSR ``middle-compiled'' and PSR ``middle'': in the
former, just two satisficing planners participated; in the latter, six
satisficing planners participated, but they all scaled quite well on
these less challenging instances so the results are less interesting
than those for PSR large.

First, note that all curves in PSR ``small'' show a large amount of
zig-zagging, which is quite unusual and which cannot simply be
accounted for by the way the instances are scaled.\footnote{The same
  is true for the runtime curves of the individual planners. In fact,
  the planners even disagree widely about which instances are solved
  easily and which take a lot of time.}  Consider
Figure~\ref{new:heuristics:psr} (a).  The main observation to be made
is that the real optimal makespan is {\em much} larger than its
estimation by a plan graph, particularly in the larger instances.
Still, the optimal parallel planners are quite efficient, at least in
that they can solve all the instances. The runtime data are entirely
due to SATPLAN'04, whose search techniques are apparently quite
efficient in this domain even with a bad plan graph lower bound.  The
other optimal planners are all at least one order of magnitude slower,
and can't solve some of the largest instances; for example, none can
solve instances 48 and 49.  As for the optimal sequential planners in
Figure~\ref{new:heuristics:psr} (b), the results are pretty similar
except that the runtime scaling is somewhat worse. For both kinds of
optimal planners, the runtime is clearly correlated with the length of
the optimal plans, which, since the plan graph bounds are almost
constant, coincides with the difference between the real plan length
and its estimate.

In Figure~\ref{new:heuristics:psr} (c), we observe that the relaxed
plan is a very bad estimator of plan length in PSR ``small'' (at least
for the respective initial states), but that the planners solve all
instances quite efficiently anyway. The runtime data are entirely due
to YAHSP and Fast Downward; particularly Fast Downward is extremely
efficient, showing only a very slight increase of runtime over
instance size, being the only satisficing planner capable of solving
instances 48 and 49. Note that YAHSP \cite{vidal:icaps-04} uses
powerful techniques besides a relaxed plan heuristic, and that Fast
Downward \cite{helmert:icaps-04} uses a more involved (and apparently
more powerful, in this case) heuristic function. Note also that, at
least in terms of solved instances, optimal and satisficing planners
are, unusually, equally good (or bad) in this domain: exactly one of
each group solves all instances, all other planners cannot solve
instances 48 and 49. The difficulty the planners are experiencing in
this domain is also remarkable since the instances, or at least their
grounded encodings, are actually very small when compared to the
instances of the other domains, c.f.  Figure~\ref{new:encoding:data}.
This indicates that the domain has some fundamental characteristic
that is not yet captured very well by the search heuristics/techniques
of (most of) the planners -- which nicely complements what we said
about the non-obvious polynomial algorithm for PSR in
Section~\ref{known:complexity}.

In Figure~\ref{new:heuristics:psr} (d), we see that the relaxed plan
(computed with the version of FF handling derived predicates, see
\citeR{thiebaux:etal:ijcai-03,thiebaux:etal:ai-05}) is a rather
useless estimator in the PSR domain when expressed in the most natural
way using ADL and derived predicates. The relaxed plan constantly
contains $0$ steps, meaning that the over-approximation of the
semantics of derived predicates makes the initial state look like a
goal state; the same happens in PSR middle. While the situation may be
different in other parts of the state space -- the heuristic value is
not constantly $0$ -- this, apparently, causes serious trouble for all
satisficing planners except Fast Downward. No planner except Fast
Downward can solve an instance higher than number 16. Fast Downward
seems to profit, again, from its more involved heuristic function,
reaching its scaling limit at instance number 31.

\vspace{0.0cm}
\begin{figure}[htb]
\begin{tabular}{cc}
\scaleonlyfig{0.58}{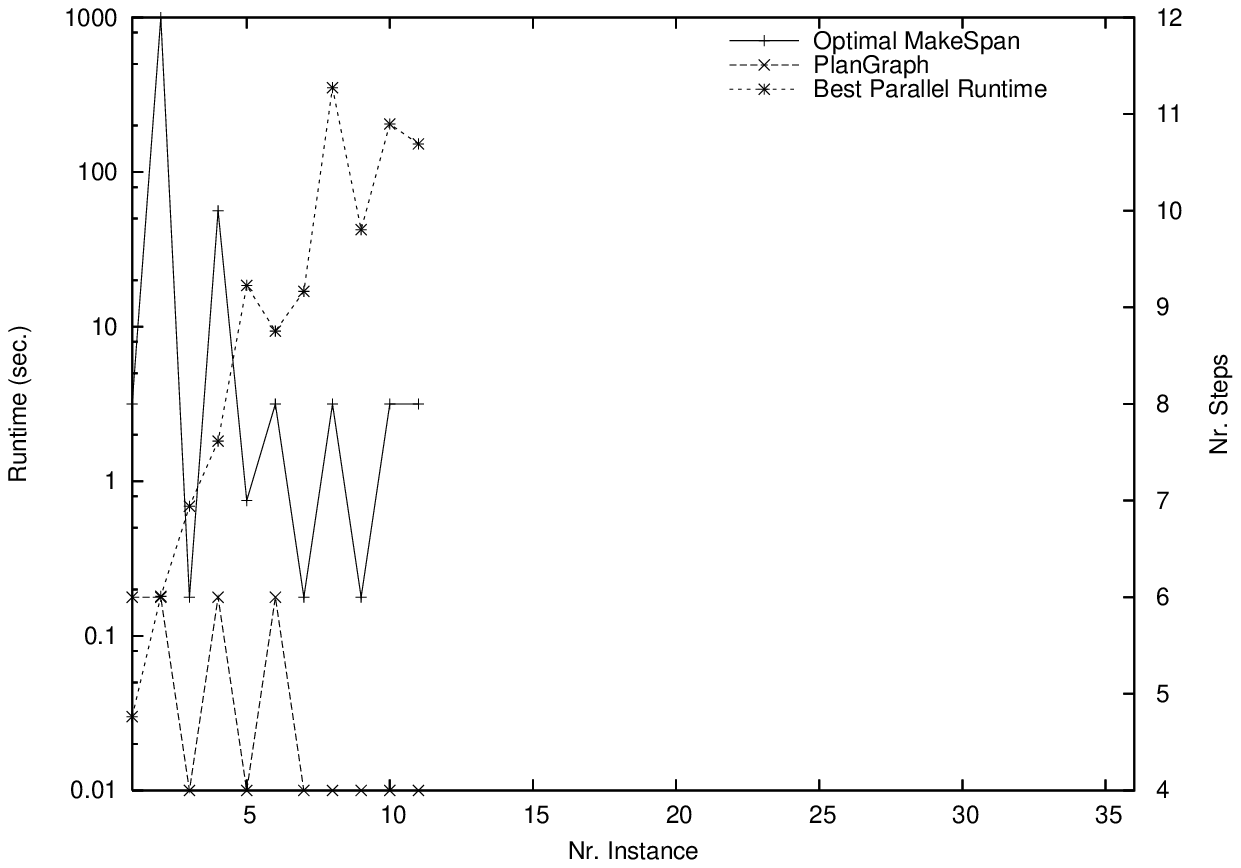} & \hspace{-0.3cm} \scaleonlyfig{0.58}{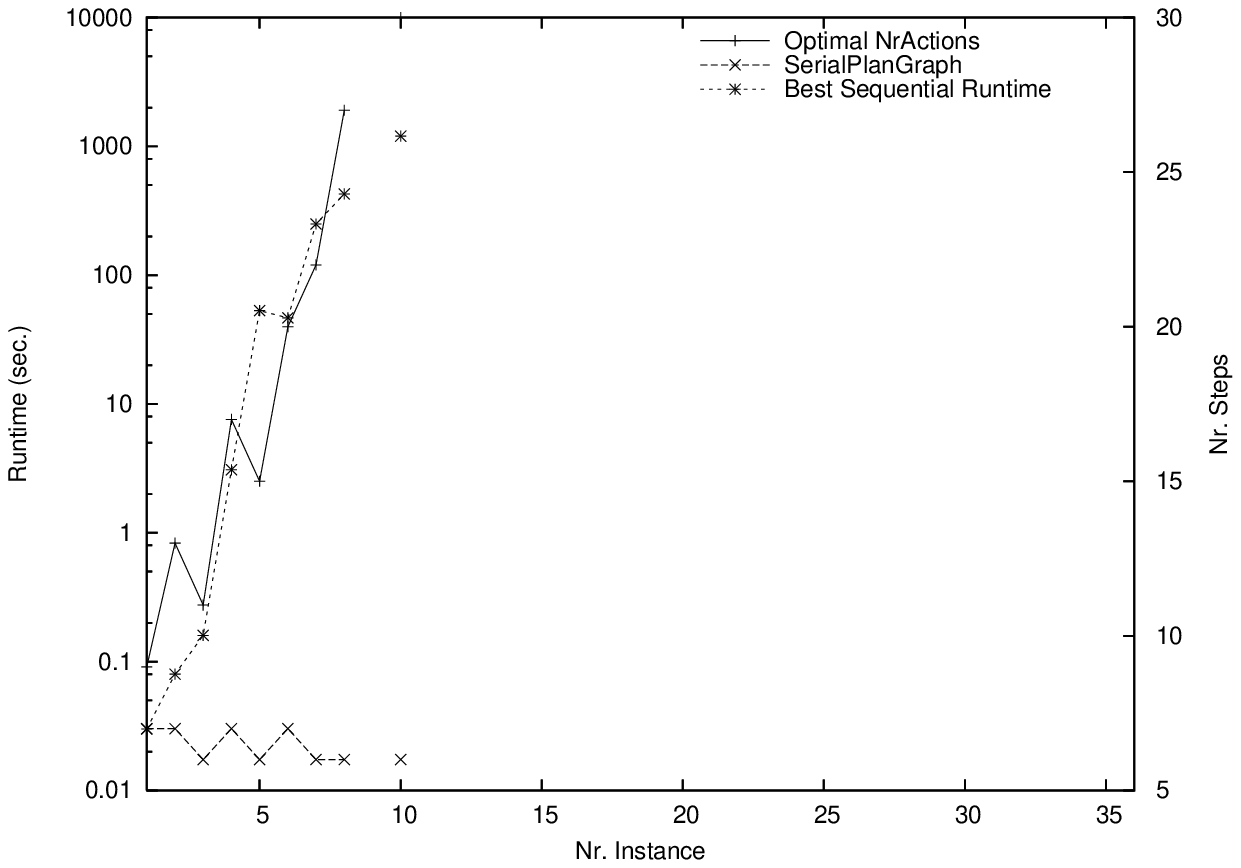}\\
(a) & (b)\\
\end{tabular}
\begin{center}
\scaleonlyfig{0.58}{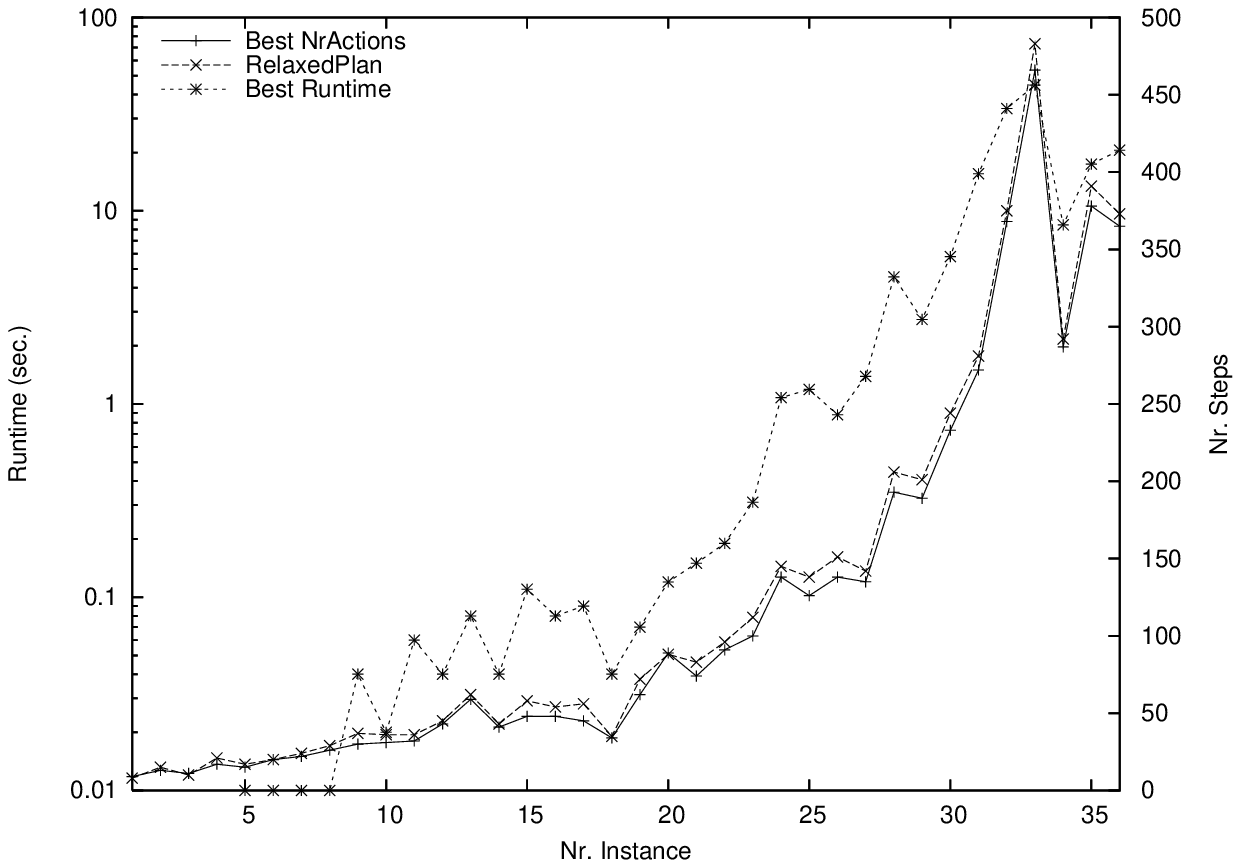}\\
(c)
\end{center}
\vspace{-0.5cm}
\caption{\label{new:heuristics:satellite}Satellite domain.  Plots of
  (parallel) plan length, its heuristic estimation, and runtime, for
  (a) optimal parallel planners, (b) optimal sequential planners, and
  (c) satisficing planners.}
\vspace{-0.4cm}
\end{figure}

In the Satellite domain, which has many temporal and some numeric
domain versions, we select, for our presentation here, the single pure
STRIPS version. In Figure~\ref{new:heuristics:satellite} (a) and (b),
we observe that, like Pipesworld and Promela, and unlike Airport and
PSR, Satellite is a domain where a serial plan graph provides much
worse heuristic values (for sequential planning) than a parallel
planning graph (for parallel planning). Over the few instances solved
by the optimal planners, parallel plan length and (serial or parallel)
plan graph length do not grow much, while sequential plan length does.
Consequently, the sequentially optimal planners scale much worse than
the parallel ones. In Figure~\ref{new:heuristics:satellite} (a), we
can also nicely see how, during instances 8, 9, 10, the parallel plan
length does a down-up movement (8, 6, 8) over the constant parallel
plan graph length (4), resulting in a movement of pretty much the same
shape -- on a logarithmic scale! -- of the best parallel runtime.

In Figure~\ref{new:heuristics:satellite} (c), we observe that, like in
Airport and unlike in any of the other domains, the relaxed plans for
the initial states have almost the same length as the real plans
(there is actually a slight over-estimation most of the time). As we
have seen earlier, c.f. Section~\ref{known:opth},
\citeA{hoffmann:jair-05} has shown that, for Satellite, the relaxed
plan length is, in fact, bound to be close to real plan length for all
states (in contrast to Airport, where unrecognized dead ends are
possible in principle). Indeed, Satellite is very easy to tackle for
almost all of the satisficing planners in IPC-4. While the runtime
shown in Figure~\ref{new:heuristics:satellite} (c) appears
non-trivial, remember that these instances are {\em huge}, see in
particular the number of ground actions in
Figure~\ref{new:encoding:data} (b). Up to instance 20, most
satisficing IPC-4 planners could solve each instance within a minute.

%: the largest
%ones take nearly 500 actions to solve -- as said, instances 21 to 36
%correspond to the 16 instances posed in IPC-3 to challenge the
%hand-tailored planners. With up to 250 actions (around instance 25),
%the instances are solved within a single second.

\vspace{0.0cm}
\begin{figure}[htb]
\begin{tabular}{cc}
\scaleonlyfig{0.58}{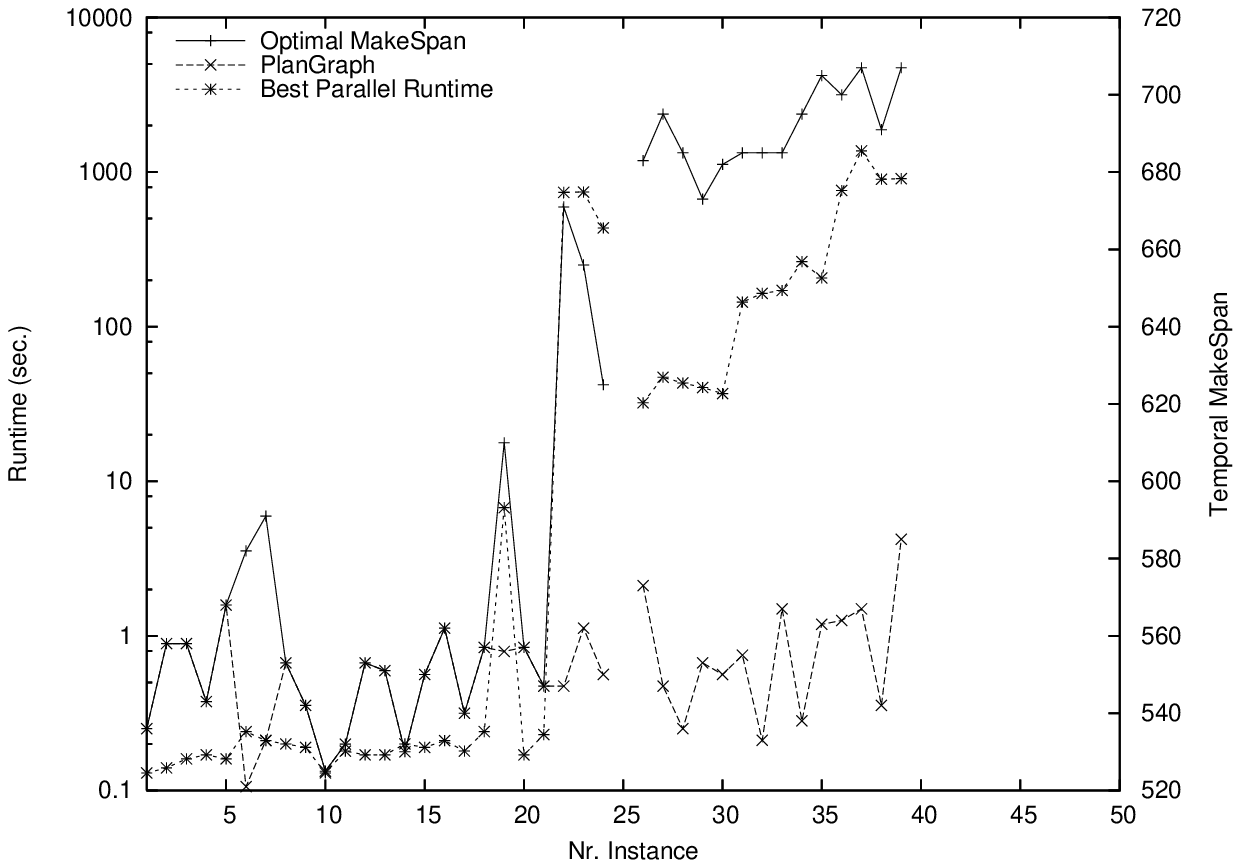} & \hspace{-0.3cm} \scaleonlyfig{0.58}{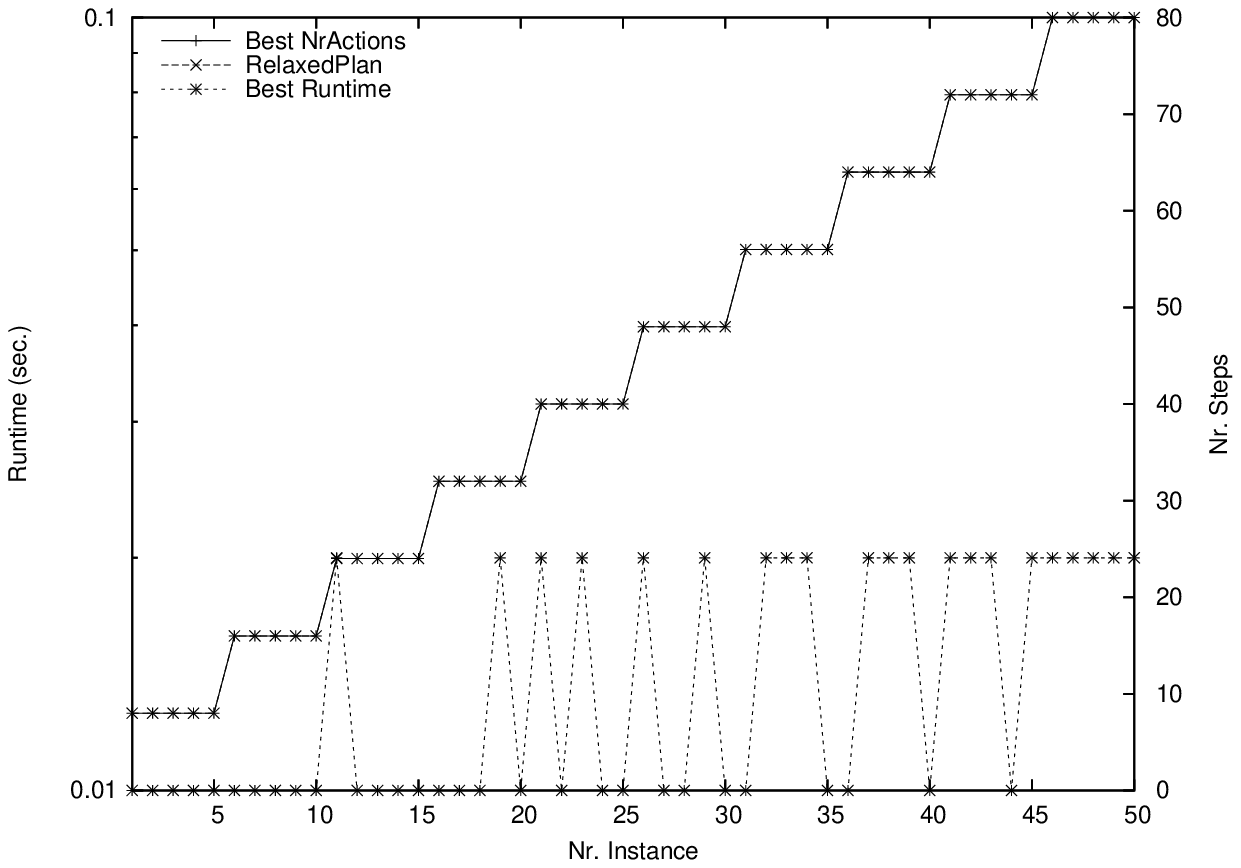}\\
(a) & (b)\\
\scaleonlyfig{0.58}{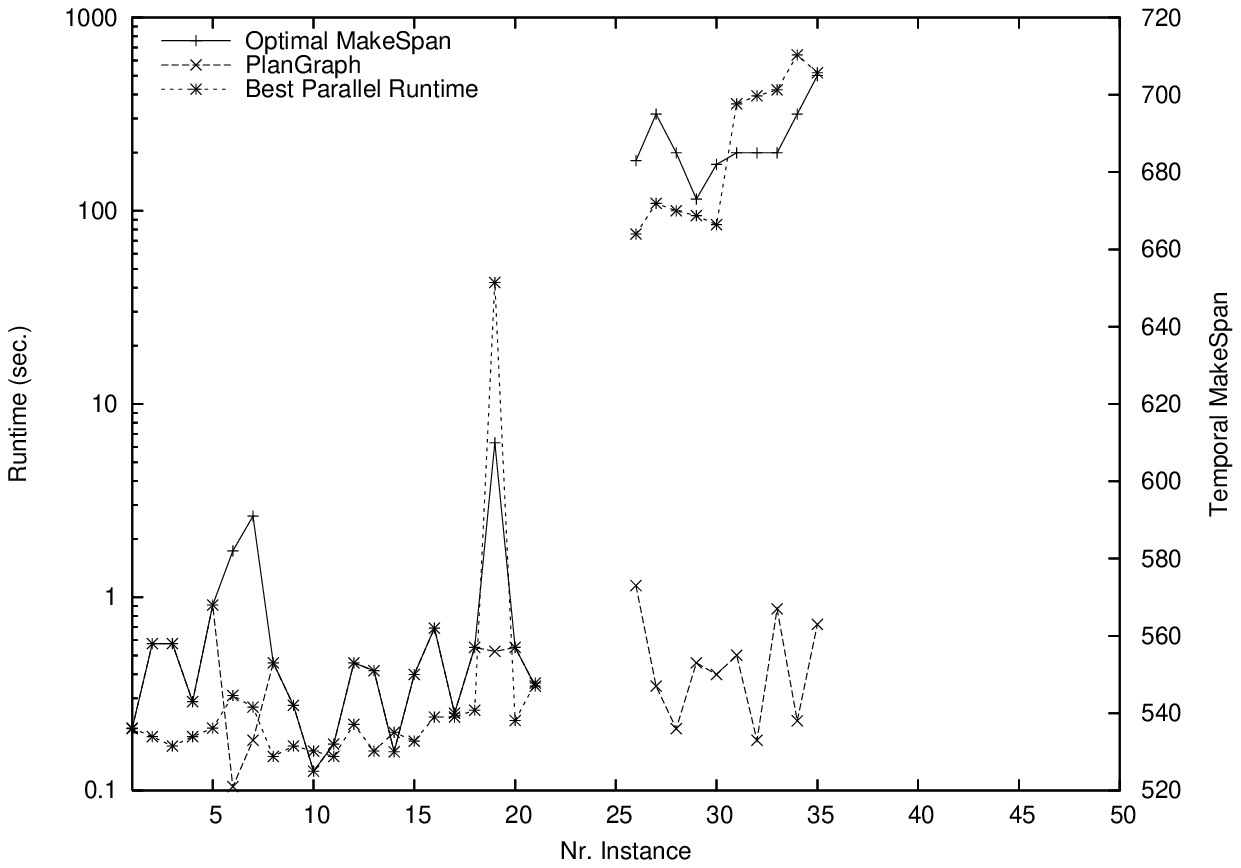} & \hspace{-0.3cm} \scaleonlyfig{0.58}{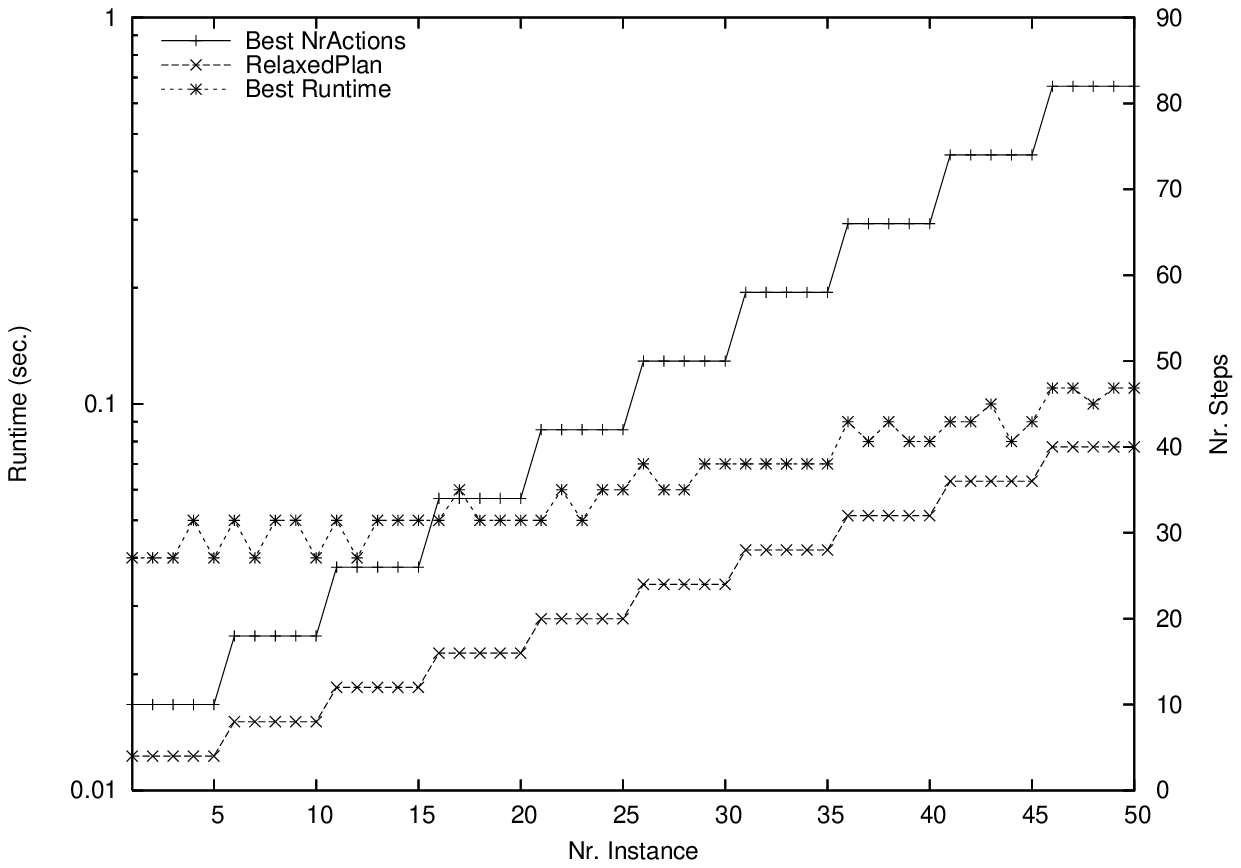}\\
(c) & (d)\\
\vspace{-0.8cm}
\end{tabular}
\caption{\label{new:heuristics:umts}UMTS domain.  Plots of (durational)
  plan length, its heuristic estimation, and runtime, for (a) optimal
  (b) satisficing planners in plain temporal version, (c) optimal (d)
  satisficing planners in temporal version with flaw action.}
\vspace{-0.4cm}
\end{figure}

We skip the Settlers domain since that relies almost exclusively on
numeric variables to encode the domain semantics, which makes it
rather incomparable with the other domains.
Figure~\ref{new:heuristics:umts} shows our data for the UMTS domain.
This has only temporal and numeric versions, half of which feature
also time windows. We consider the versions without time windows;
Figure~\ref{new:heuristics:umts} (a) and (b) concern the plain domain
version, Figure~\ref{new:heuristics:umts} (c) and (d) is with ``flaw''
action. Let us first consider the optimal planners, on the left hand
side of the overall figure. The only optimal planners that could
tackle this domain -- i.e., the domain's syntax -- were TP4 and
HSP$_a^*$ \cite{haslum:geffner:ecp-01}. These are makespan-minimizing
planners, and so there are no data for sequentially optimal planners
(which wouldn't make a lot of sense in the temporal setting anyway).
The ``PlanGraph'' curves in Figure~\ref{new:heuristics:umts} (a) and
(c) correspond to the makespan estimation delivered for the initial
state by TP4's temporal numeric extension of that heuristic. For the
effect of heuristic quality on runtime, we observe once again a very
strong correlation. In Figure~\ref{new:heuristics:umts} (a), up to
instance 21 the makespan estimate is very close to the real makespan
-- most of the time, the two actually coincide -- and the runtimes are
very good. Starting from instance 22, the real makespan makes a sudden
leap upwards that is not followed by the estimation, and the runtimes
shoot upwards. The phenomenon is also very clear in instances 18, 19,
20, where the makespan estimation exhibits a good, bad, good pattern,
and the runtime does just the same. In
Figure~\ref{new:heuristics:umts} (c), the very same sort of behavior
can be observed, meaning in particular that the flaw action does not
have an effect on makespan and its estimation by TP4. In fact, the
makespan and its estimation are exactly the same in all instances
solved in both domain versions.  As contained implicitly in the latter
sentence, the flaw action {\em does} affect runtime and with it the
set of solved instances. The runtime with the flaw action is
consistently more than a factor of 2 larger than without the flaw
action. In the most challenging instances the planners fail when the
flaw action is present. This decrease in performance is presumably due
to the larger state space incurred by the flaw action.

Consider the satisficing planners, Figure~\ref{new:heuristics:umts}
(b) and (d). We first observe that, once more, we are facing a very
individual and characteristic behavior, and that the domain is no
challenge at all to the satisficing planners. The latter shows that
the domain is not a useful benchmark for satisficing planners; it also
shows once again how heterogeneous our benchmark set is: while it is
common that satisficing planners are faster than optimal ones --
except in PSR -- there is no other domain where that picture is as
extreme as in UMTS. As stated earlier, the domain is a pure scheduling
problem, and obviously the satisficing planners provide
runtime-efficient greedy solutions to that problem.\footnote{In terms
  of quality of the solutions found, the satisficing planners also do
  reasonably well.  For example, LPG-td, which minimizes makespan in
  this domain, finds, with its version optimized for speed, plans that
  take maximally 10\% more time than the optimal ones found by TP4.
  For the version of LPG-td optimized for plan quality, this goes down
  to 1\%.} Looking at the plots in a little more detail, we find in
Figure~\ref{new:heuristics:umts} (b) that the sequential plan length
(the plans found are optimal) is a simple stepwise linear function in
these instances, and that relaxed plan length for the initial state
coincides once again with the real plan length -- which isn't a
surprise given the excellent runtimes of the satisficing planners, and
the fact that this is a scheduling domain. (In a sequentialized
schedule no harmful delete effects occur.) This picture changes a lot
in Figure~\ref{new:heuristics:umts} (d). The real plan length stays
basically the same (is increased by a constant of $2$), but the
relaxed plan length becomes a lot shorter due to the flaw action. The
satisficing planners are unaffected, largely keeping their excellent
runtime behavior. Apparently, these planners incorporate some
technique for recognizing the uselessness of the flaw action (this can
be done with simple domain analysis techniques), and getting rid of
its influence. This suspicion is confirmed by the fact that there is
one satisficing planner that {\em does} get affected by the flaw
action in the way one should expect. CRIKEY, a heuristic search
forward state space planner using a relaxed plan heuristic, solves
each task within 70 seconds without the flaw action, but sometimes
takes over 1000 seconds {\em with} the flaw action.

Let us briefly summarize the overall observations:
\begin{itemize}
\item In the presented data, most of the time the performance of the
  planners correlates well with the quality
%ST english
%of the respective most relevant one of the heuristic functions.
  of the relevant heuristic function.  The most notable exceptions to
  this rule -- as far as can be observed in our data here -- are Fast
  Downward in PSR ``large'', where relaxed plans are pretty much
  devoid of information, and SGPlan and YAHSP (to some extent also
  Fast Downward) in Pipesworld, where relaxed plans provide poor
  estimates and all other planners experience (much more) serious
  difficulties.
\item Usually, here and in the known benchmarks in general,
  satisficing planners are several orders of magnitude faster than
  optimal ones. Exceptions here are PSR -- where both groups perform
  almost equally -- and UMTS -- where the satisficing planners hardly
  need any time at all.
\item Usually, here and in the known benchmarks in general, parallel
  plan graph length is a much better estimator of parallel plan length
  than serial plan graph length is of sequential plan length. The
  exceptions here are Airport -- where there is often a huge
  difference between the lengths of the two kinds of plan graphs --
  and, to some extent, PSR ``small'' -- where the difference between
  parallel and sequential plan length is not very big. Note that none
  of our domains is purely sequential, i.e. some parallelism is
  possible in all of them.
\item Usually, here and in the known benchmarks in general, there is a
  considerable difference between the length of a relaxed plan for the
  initial state, and the length of a real plan for the initial state.
  Exceptions here are Airport, Satellite, and UMTS, where both lengths
  are identical or nearly so.
\item Usually, here and in the known benchmarks in general, the
  largest instances that can be solved within the given particular
  time and memory (30 minutes and 1GB) have plans with around a
  hundred steps or more. PSR is exceptional in that Fast Downward is
  the only planner able to find a plan with more than 35 (namely, with
  57) steps.
\end{itemize}
It once again indicates the diversity of the IPC-4 domains that almost
every one of them appears at least once in the ``exceptions'' listed
here.  The only domains that don't appear there are the Promela
domains and Pipesworld. This is a sort of exception in itself, meaning
that these domains contribute the more typical benchmark behaviors to
the overall set.

We take the existence of some of the mentioned distinguishing features
as evidence that the IPC-4 domains indeed have several novel aspects,
besides being oriented at applications and being structurally diverse.
In particular, the behavior of the PSR domain stands out from what one
typically observes. Note here that, while it is typically easy to
construct artificial domains that provoke some unusual behavior, the
domains we have here are {\em oriented at applications}, and so the
exhibited behavior, particularly that of the PSR domain, is not only
unusual, but also relevant in a very concrete sense.

\subsection{Fact Connectivity}
\label{new:connectivity}

We conclude our empirical analysis with some data aimed at assessing a
sort of ``connectivity'' of the facts. For each fact $p$, we measure
the number of {\em adders}: actions that have $p$ in their add list
(in the ADL case, that have an effect with $p$ in its adds list). This
gives an indication of the branching factor -- action choices -- that
comes with the fact. We further measure the number of {\em requirers}:
actions that have $p$ in their precondition (in the ADL case, that
have an effect with $p$ in its condition). This gives an indication of
how central a fact is to the task. For a given planning task, we
measure the parameters of the distribution of adders(p) and
requirers(p), over the set of facts $p$: the minimum ($min$), mean
($mean$), maximum ($max$), and standard deviation ($dev$). Within
domain versions, we plot these data over instance size (number).

\begin{figure}[htb]
\begin{tabular}{cc}
\scaleonlyfig{0.58}{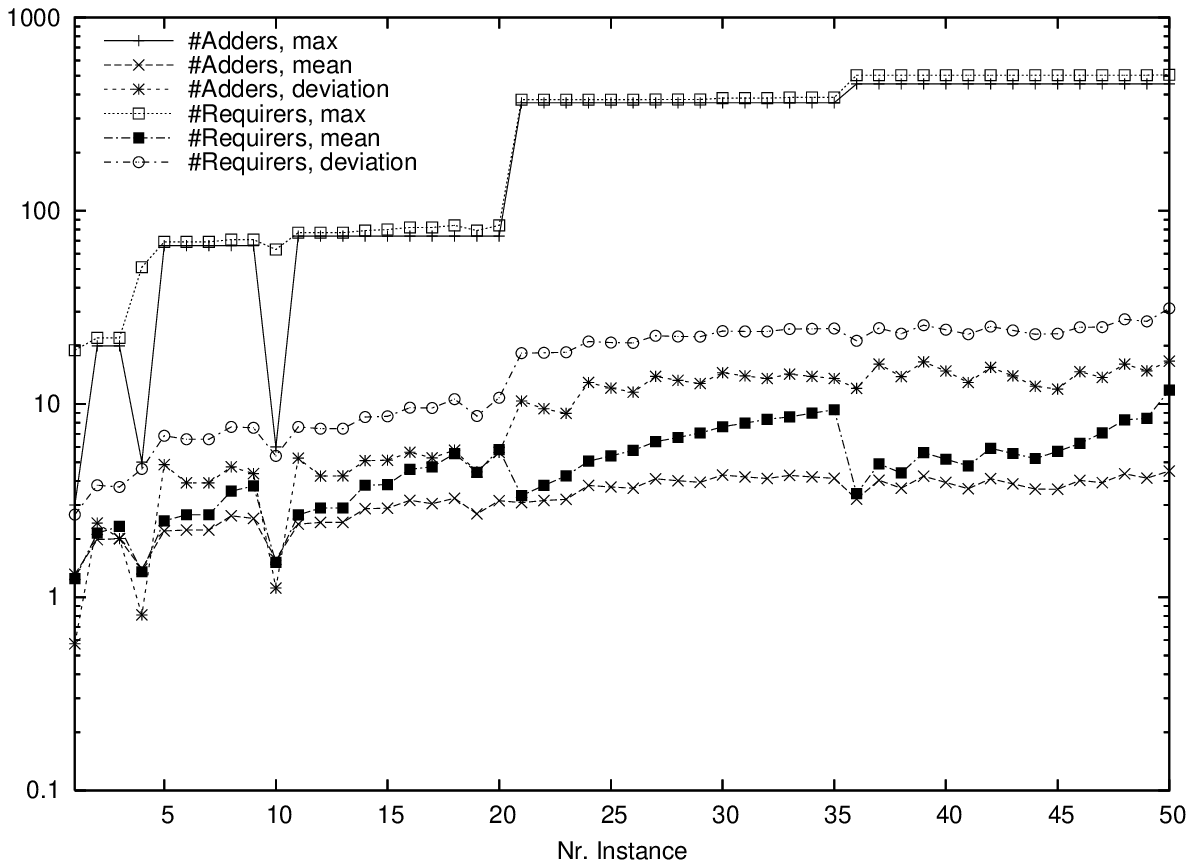} & \hspace{-0.3cm} \scaleonlyfig{0.58}{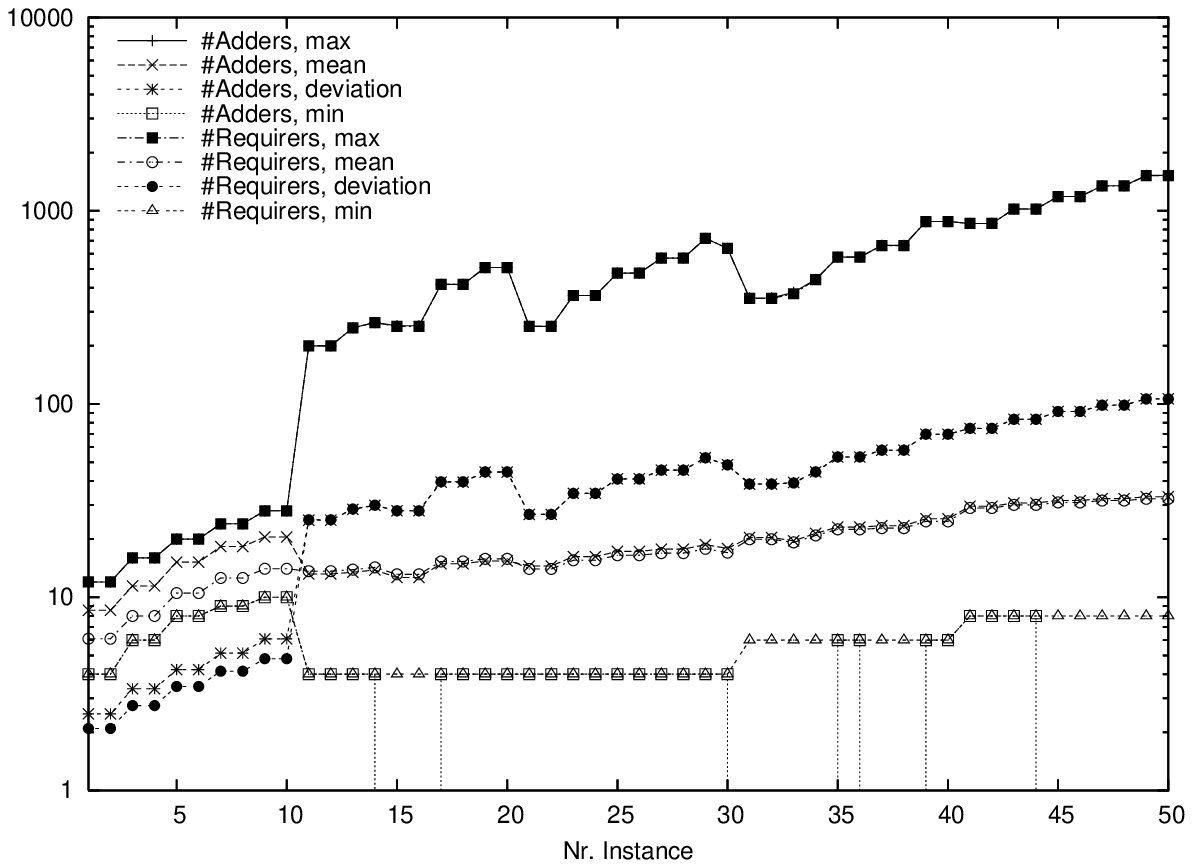}\\
(a) & (b)\\
\scaleonlyfig{0.58}{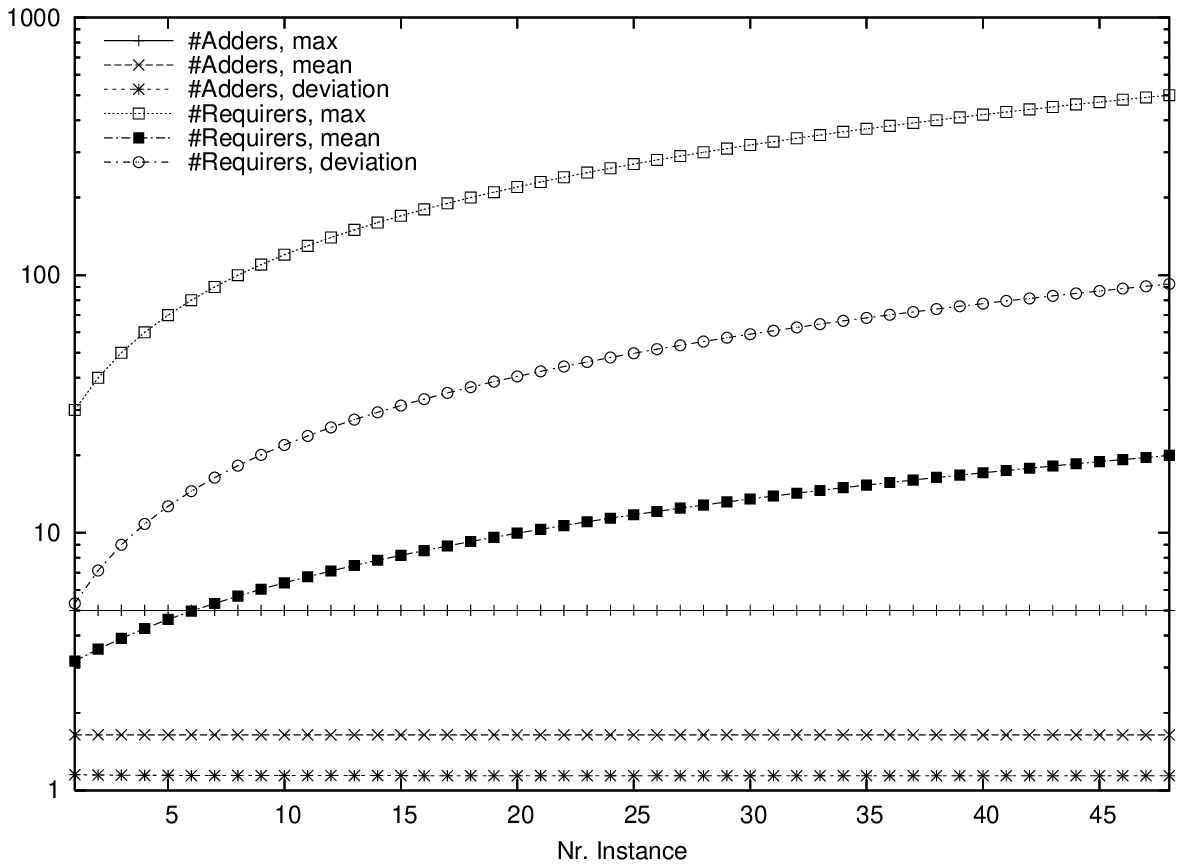} & \hspace{-0.3cm} \scaleonlyfig{0.58}{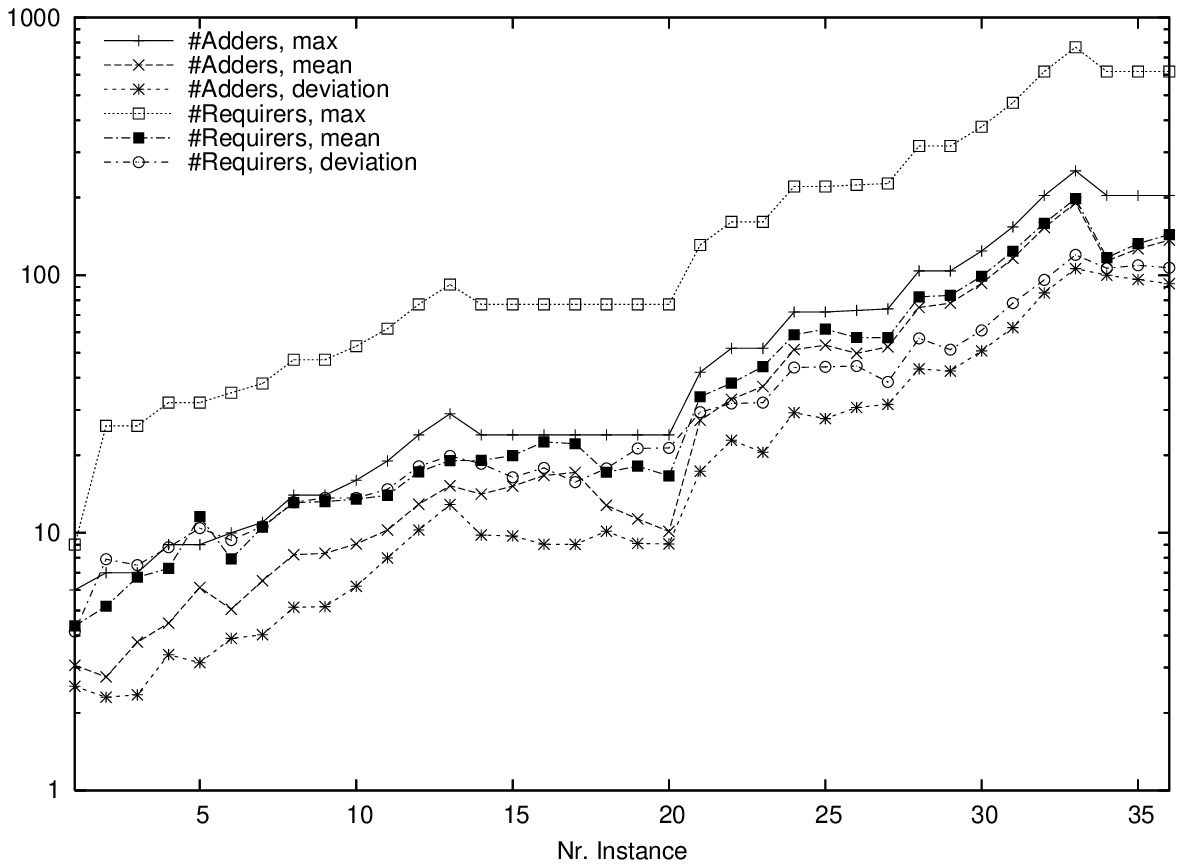}\\
(c) & (d)\\
\vspace{-0.8cm}
\end{tabular}
\caption{\label{new:connectivity:many}Distributions of the numbers of
  actions adding a fact, and of actions requiring a fact, in selected
  versions of some IPC-4 domains: (a) Airport, (b) Pipesworld, (c)
  Dining Philosophers, (d) Satellite.}
\vspace{-0.4cm}
\end{figure}

\begin{figure}[htb]
\begin{tabular}{cc}
\scaleonlyfig{0.58}{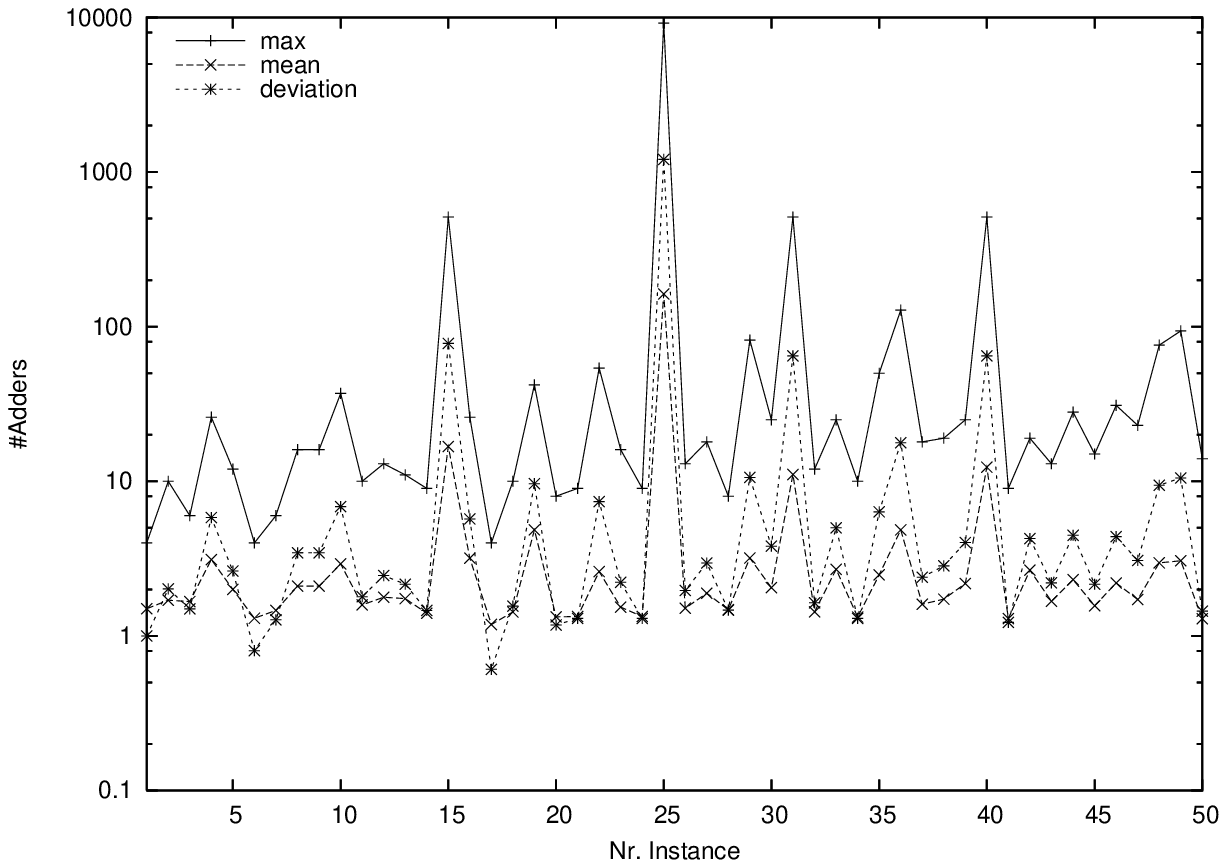} & \hspace{-0.3cm} \scaleonlyfig{0.58}{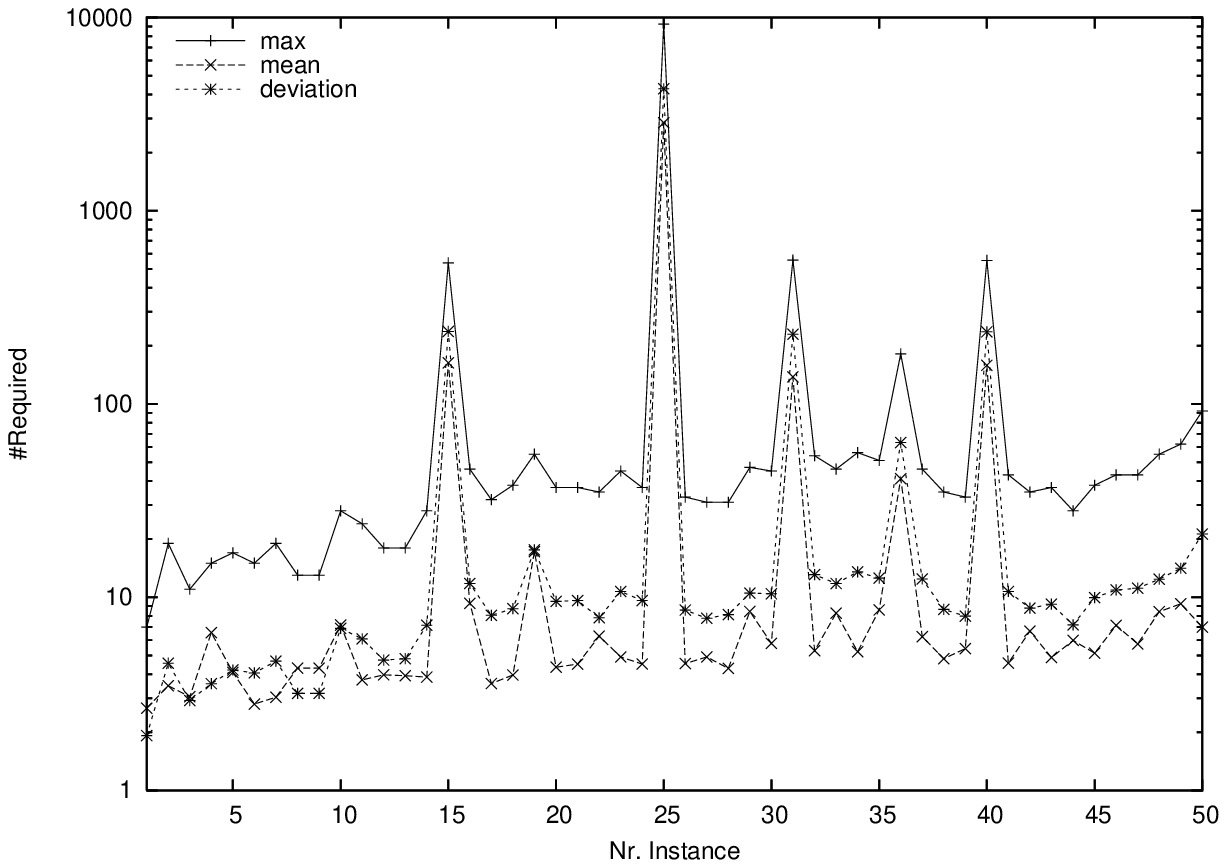}\\
(a) & (b)\\
\scaleonlyfig{0.58}{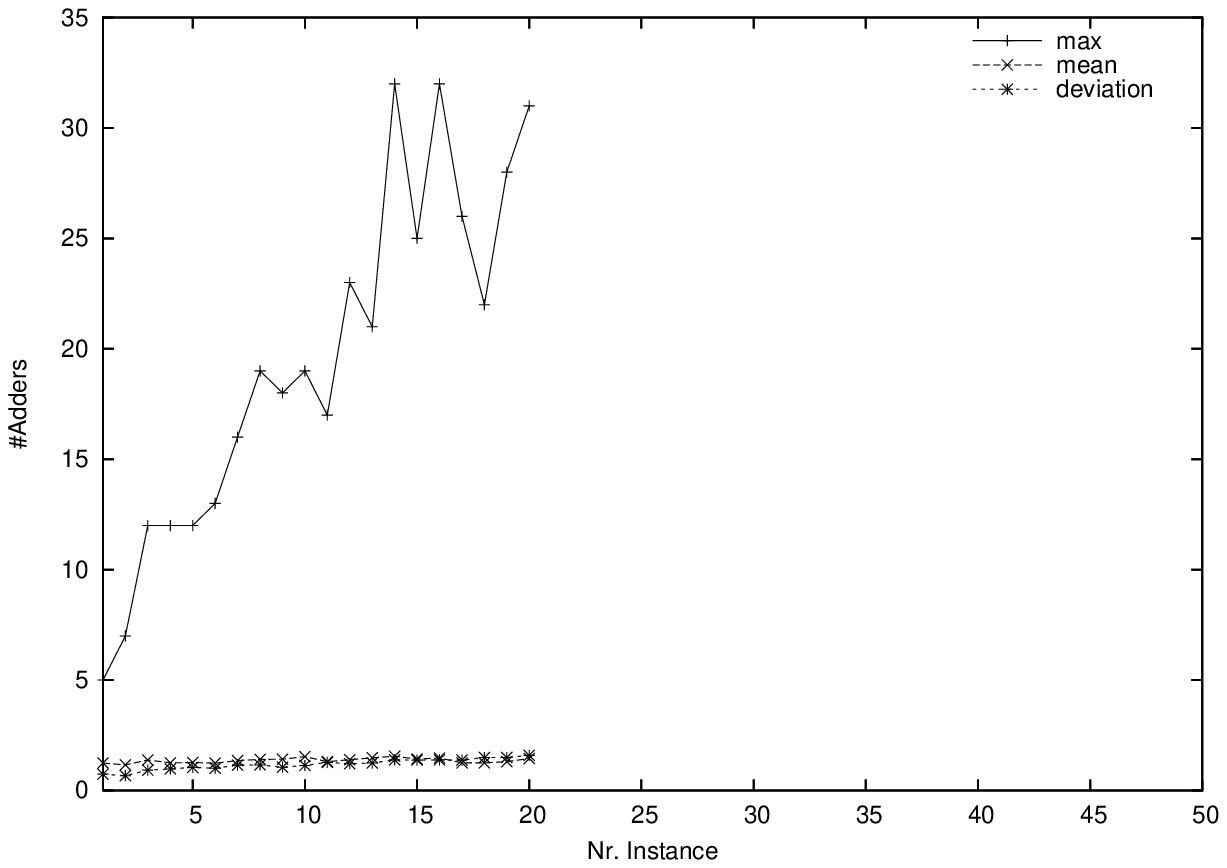} & \hspace{-0.3cm} \scaleonlyfig{0.58}{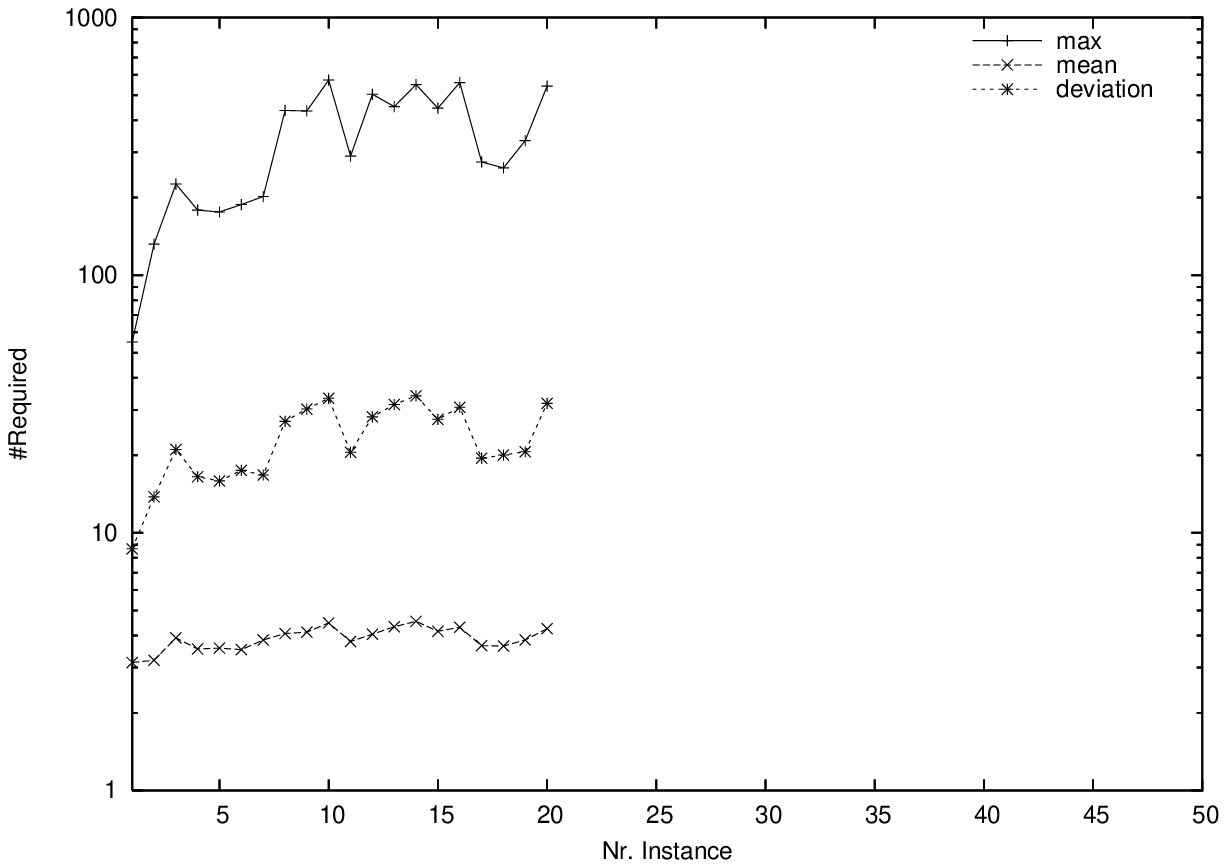}\\
(c) & (d)\\
\vspace{-0.8cm}
\end{tabular}
\caption{\label{new:connectivity:psr}Distributions of the numbers of
  actions adding a fact, and of actions requiring a fact, in PSR
  ``small'' and ``large'': (a) adders ``small'', (b) requirers
  ``small'', (c) adders ``large'', (d) requirers ``large''.}
\vspace{-0.4cm}
\end{figure}

The data are too abstract to allow deep conclusions about reasons for
planner performance, but we {\em are} able to highlight some more
characteristic features of the domains. In particular, we will see
that these abstract measurements behave more characteristically
different in the IPC-4 domains than in the IPC-3 domains.
Figure~\ref{new:connectivity:many} shows our plots for the IPC-4
domains Airport, Pipesworld, Dining Philosophers, and Satellite. The
picture for PSR is relatively complicated and shown separately in
Figure~\ref{new:connectivity:psr}. Settlers is left out because it is
exceptional. The picture for UMTS is extremely simple, and explained
in the text below.

\vspace{0.0cm}

Consider Figure~\ref{new:connectivity:many} (a), the (non-temporal)
Airport domain.  The $min$ curves are not shown since they are
constantly $0$: ``is-pushing-back(airplane)'' is never added since
pushback requests (of outbound traffic) are not modelled;
``occupied(segment)'' is only required in its negation. 
%ST in its negative form?
The $max$ curves are step functions since they follow the size of the
underlying airports: ``is-moving(airplane)'' has as many adders as
there are segments, since ``start-up-engine'' can be done at any
segment; ``is-pushing-back(airplane)'' is required by every such
action, leading to the overall similar form of the $max$ requirers
curve. The $mean$ adders curve is flattened because all facts other
than ``is-moving(airplane)'' are added only at certain places on the
airport. The $mean$ requirers curve, interestingly, shows a similar
downwards step behavior as the numbers of facts and actions shown in
Figure~\ref{new:encoding:data}. The reason lies in the
``not-occupied'' facts, that exist for every segment, and that are
needed in every action moving an (any) {\em airplane} across the
segment.  The number of these facts increases with the number of
airplanes. Since there are many of these facts, they have a strong
influence on the mean. There is not much of a correspondence to
runtime in the data, other than the trivial one that both tend to grow
with instance size.

Data for Pipesworld, no tankage non-temporal, are shown in
Figure~\ref{new:connectivity:many} (b). Several observations can be
made: 1. the $max$ and $mean$ curves clearly follow the scaling
pattern, with growing traffic on the $5$ growing underlying networks.
2.  the $min$ curves are non-zero. 3. there is a characteristic
difference between the curves up to instance 10, and afterwards. 4.
the curves for adders and requirers almost (but not exactly) coincide.
Apart from 1, which is also present in the Airport data, these
observations clearly distinguish Pipesworld from all the other
domains. As for observation 2, sometimes in the larger instances the
$min$ number of adders {\em does} drop to $0$. This is due to
interactions in more complex networks, where certain configurations
inside pipes are true initially but can not be re-achieved later on --
some of these interactions are recognized by the ``reachability''
pre-process made by FF for actions, c.f. the explanation in
Section~\ref{new:encoding}.  Observation 3 is due to a large contrast
between the smallest network and all larger ones: the smallest network
has only unitary pipelines (containing just a single batch), the
others have pipelines of at least length $2$.  Observation 4 is
particularly at odds with all the other domains, where there are large
differences between adders and requirers. In fact, measuring the
distribution of the {\em difference} between adders and requirers, we
found that these numbers (not only their distribution parameters) are
extremely close together: in instance 50, where the $max$ adders is
$1524$ and $max$ requirers is $1520$, the $max$ of the difference is
$29$, with a $mean$ of $1.63$ and $dev$ of $5.31$. In Pipesworld with
tankage restrictions, the phenomenon is somewhat less extreme but
still there.  Another characteristic is the enormously large $max$
number of adders and requirers, about an order of magnitude larger
than in the other domains. The $max$ adders and requirers come from
``do-normal'' facts, which control the status of individual pipelines,
and are affected by each action moving some combination of batches
through the respective pipeline; all other facts depend on only single
batches (not combinations of them), which flattens the $mean$ curves
by two orders of magnitude. Regarding runtime, as mentioned earlier,
in Pipesworld the scaling pattern does not have a clear correlation
with runtime; neither does the fact connectivity we measure here.

Consider the Promela domain in Figure~\ref{new:connectivity:many} (c),
data shown for Dining Philosophers with derived predicates. Once
again, the extreme characteristics of the domain are recognizable at
first glance. The data for Dining Philosophers without derived
predicates are identical, the data for Optical Telegraph differ only
in that the numbers are higher. The $min$ curves are both $0$, the
adders are constant, the requirers are linear. There exist facts
without adders due to an oddity in the encoding, where certain
start-up transitions put the forks on the table in the first place;
the facts without requirers are ``blocked-philosopher'', which are
only needed for the goal. The number of adders does not depend on the
instance size due to the very static sort of domain structure, where
size increases the number of parallel processes (philosophers), but
the form of the processes stays fixed, and every process interacts
with exactly two other processes. The number of requirers is linear
(non-constant, in particular) due to a technicality of the encoding,
where ``activating'' (requesting) and ``performing'' (executing) a
transition requires {\em all} communication channels to be in neutral
state; so the respective flags are required by all transitions, and
that number of course grows over size. All other facts are required
only locally, resulting in the much lower (easily two orders of
magnitude) $mean$. As one would expect in a domain with such a simple
scaling pattern, planner performance is pretty much a function of
size.

Data for Satellite (STRIPS version) are shown in
Figure~\ref{new:connectivity:many} (d).  The most characteristic
feature, in comparison to the other domains, is the extremely smooth
and parallel close-together growth of the curves. The only curve that
stands out a little is $max$ requirers; $max$ adders is due to
``pointing(satellite, direction)'' facts that can be added when
turning there from any other direction; $max$ requirers is due to
``power-on(instrument)'' facts, which are needed for every
``take-image'' with the instrument, which can be done in every {\em
  combination} of direction and image mode supported by the
instrument. Note that, in contrast to the other domains where the
$max$ curves are about two orders of magnitude higher than the $mean$,
here $max$ requirers is only one order of magnitude above all the
other curves, and these other curves are all roughly of the same
order. The $min$ curves are not shown since they are constantly $1$
for adders -- ``power-on(instrument)'' is only added by
``switch-on(instrument)'' -- and constantly $0$ for requirers --
``have-image(direction)'' is only needed for the goal. The runtime
performance of the IPC-4 planners scales relatively smoothly with size
in Satellite, like our parameters here do.

In UMTS, {\em all} the parameters are constants. This is another
consequence of the aforementioned scaling pattern, where the number of
specified applications is the same in all instances, and what changes
is (only) the goal, specifying which of the applications shall
actually be scheduled. Precisely, in the plain domain version, the
number of adders is $1$ for {\em all} facts, nicely showing the
scheduling-domain characteristic where there is no choice of {\em how}
to accomplish tasks, but only about {\em when} to accomplish them.
This is another illustration of why the satisficing planners find this
domain trivial, whereas an optimal planner like TP4
\cite{haslum:geffner:ecp-01} can spend a long time searching for the
optimal schedule. The number of requirers is minimum $0$, maximum $2$,
mean $0.89$, standard deviation 0.57. In the domain version with flaw
action, the most notable difference is that now $max$ adders is $2$ --
due to the alternative provided by the ``flaw'' action ($min$ is now
$0$, mean $1.2$, deviation $0.5$). It is interesting to note in this
context that, as mentioned above, in this domain version there {\em
  is} a satisficing planner, CRIKEY, that experiences serious trouble.

\vspace{0.0cm}

Data for PSR are shown in Figure~\ref{new:connectivity:psr}. Here, we
show plots for adders and requirers separately because that makes them
much more readable. Since the data contain some particularly
interesting phenomena, we show it for two domain versions, ``small''
and ``large''. The most obvious feature in ``small'',
Figure~\ref{new:connectivity:psr} (a) and (b), is, once again, the
huge amount of variance in the data.  The clearly discernible peaks in
the curves (instance nrs. 15, 25, 31, and 40) coincide with the peaks
in size as measured by numbers of facts and actions in
Figure~\ref{new:encoding:data}. We also note that there is a very
large range of values, spanning four orders of magnitude, even though
the instances are (except number 25) all very small in comparison to
the other domains shown in Figure~\ref{new:encoding:data}. The minimum
numbers of adders and requirers are constantly $1$:
``updated(breaker)'' is added only by a ``wait(breaker)'' action,
``not-closed(breaker)'' is only needed if one wants to close
it.\footnote{Sometimes there are $0$ minimum requirers due to an
  artificial ``goal-reached'' fact, introduced to get rid of complex
  goal formulas, c.f. Section~\ref{compilations}.}  Regarding the
maximum adders and requirers, in instance 25, which has by far the
highest (9400) total number of actions, $max$ adders (9216) is due to
the ``goal-reached'' fact, i.e., to the 9216 disjuncts in the DNF of
the goal formula; $max$ requirers (9251) is due to ``do-normal'',
which is a flag needed for every goal-reached action, plus the actions
opening or closing breakers. We remark that the same facts are
responsible for all of the peaks in the curves, i.e., the same happens
also in instances 15, 31, and 40.

It is highly characteristic for PSR ``small'' that the $max$ numbers
of adders and requirers approach and sometimes exceed two thirds of
the total number of actions. This is not the case for any other
domain, not even for any other domain {\em version} of PSR (see
below). The intuitive reason lies in one of the pre-compilation steps
that we employed in order to be able to formulate reasonably large PSR
instances in pure STRIPS: the compilation step
\cite{bertoli:etal:ecai-02} ``removes'' network reasoning (and with
it, the need for derived predicates) by basically enumerating the
breaker configurations and their effects on the flow of current in the
network. The result is a very dense structure where each end of the
network directly affects every other end, explaining the very high
degree of fact connectivity, in particular explaining the extremely
complex goal formulas in the four ``peak'' cases mentioned above.

The pre-compilation step is also the key to understanding the huge
difference between the behavior in ``small'', and in ``large''. The
latter is shown in Figure~\ref{new:connectivity:psr} (c) and (d).
There, the $max$ adders curve is a small linear function -- note the
non-logarithmic scale of the $y$ axis -- in spite of the (mostly) much
larger numbers of actions. For example, the instance with the highest
number (7498) of actions and derivation rules is number 20, where the
$max$ number of adders is 31, {\em less than half a percent of the
  total number of actions}. In the natural high-level domain encoding
that we have here, the flow of current through the network is modelled
as the transitive closure over derivation rules that each propagate
current based on the {\em local} status of the network. So in
particular the breaker configurations and their effects on the flow of
current are implicit in the structure of the network.

Once again, in PSR ``large'', the $min$ curves are constantly $0$ for
both adders and requirers; ``not-affected(breaker)'' is the negation
of a derived predicate (needed as precondition of open and close
actions), which isn't added by an inverse rule, but given its meaning
through the negation as failure semantics of derived predicates;
``fed(line)'' is only required for the goal. The $mean$ and $dev$ of
the adders are completely flattened by the numerous (5029 out of 5237,
in instance 20) ``upstream(x,y)'' facts, true if there is currently a
path open from a side of node x to a side of node y, that are added
only by a local derivation rule that relies on the same predicate for
the neighbors of y. Similarly to Satellite, the $max$ number of
requirers is generally a lot larger than the $max$ number of adders.
For example, 542 vs. 31 in instance 20, where $max$ requirers is due
to a fact ``closed(device)'' that is required in derivation rules
talking about {\em pairs} of devices; in instance 20, 7360 of the 7498
actions are such rules; there are 46 devices.

\vspace{0.0cm}
\begin{figure}[htb]
\begin{tabular}{cc}
\scaleonlyfig{0.58}{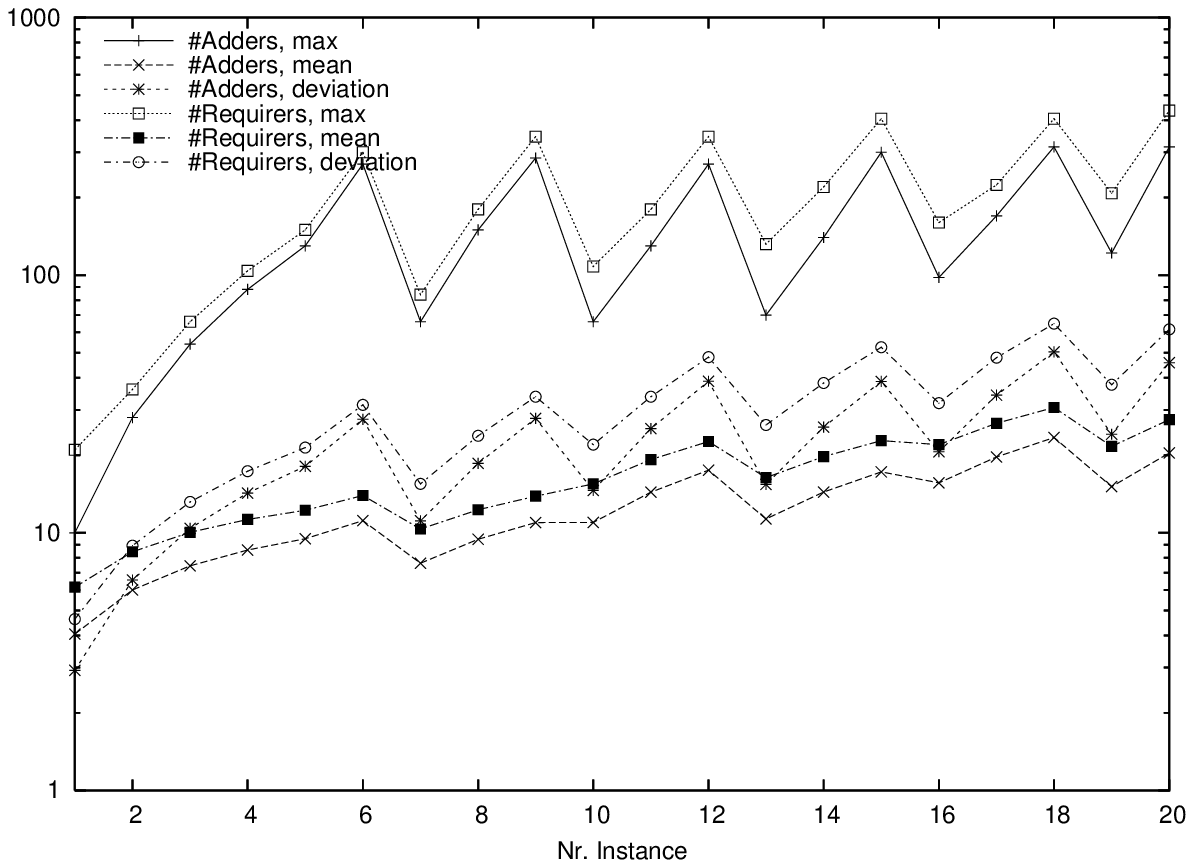} & \hspace{-0.3cm} \scaleonlyfig{0.58}{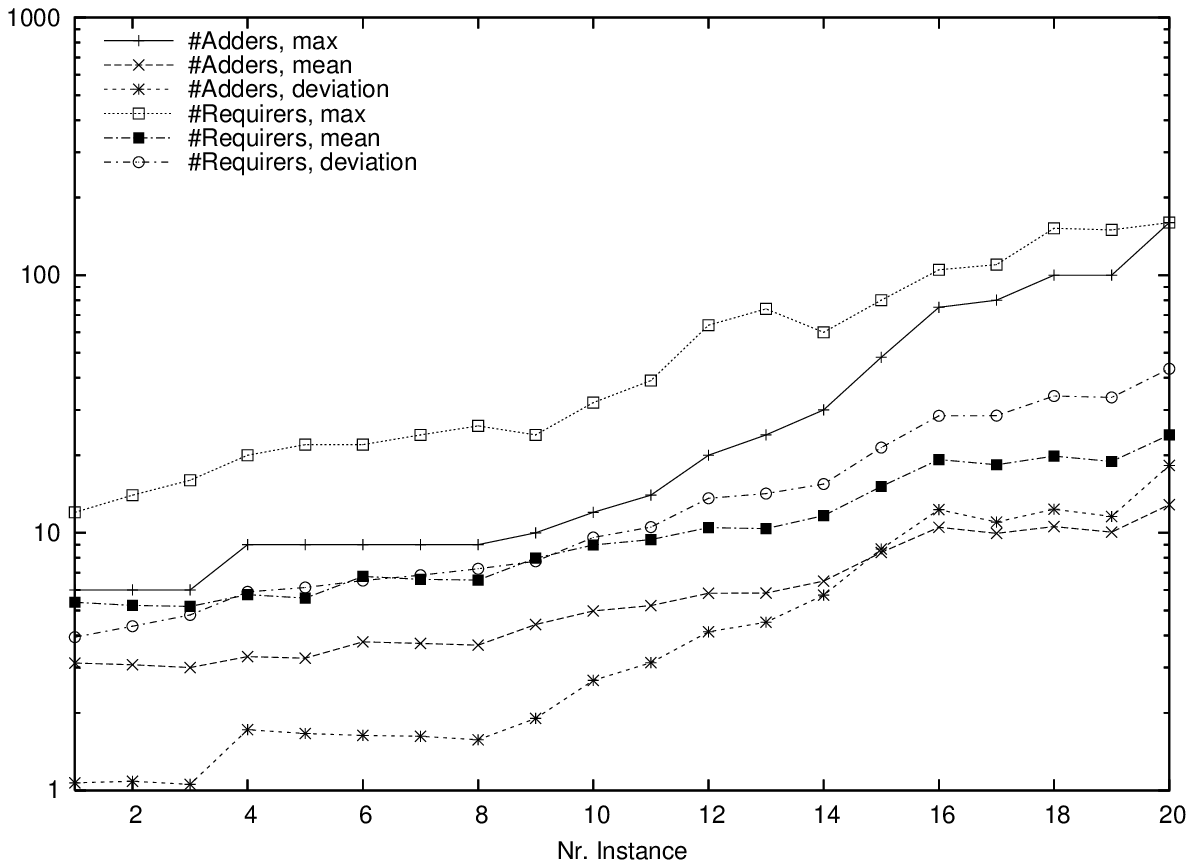}\\
(a) & (b)\\
\scaleonlyfig{0.58}{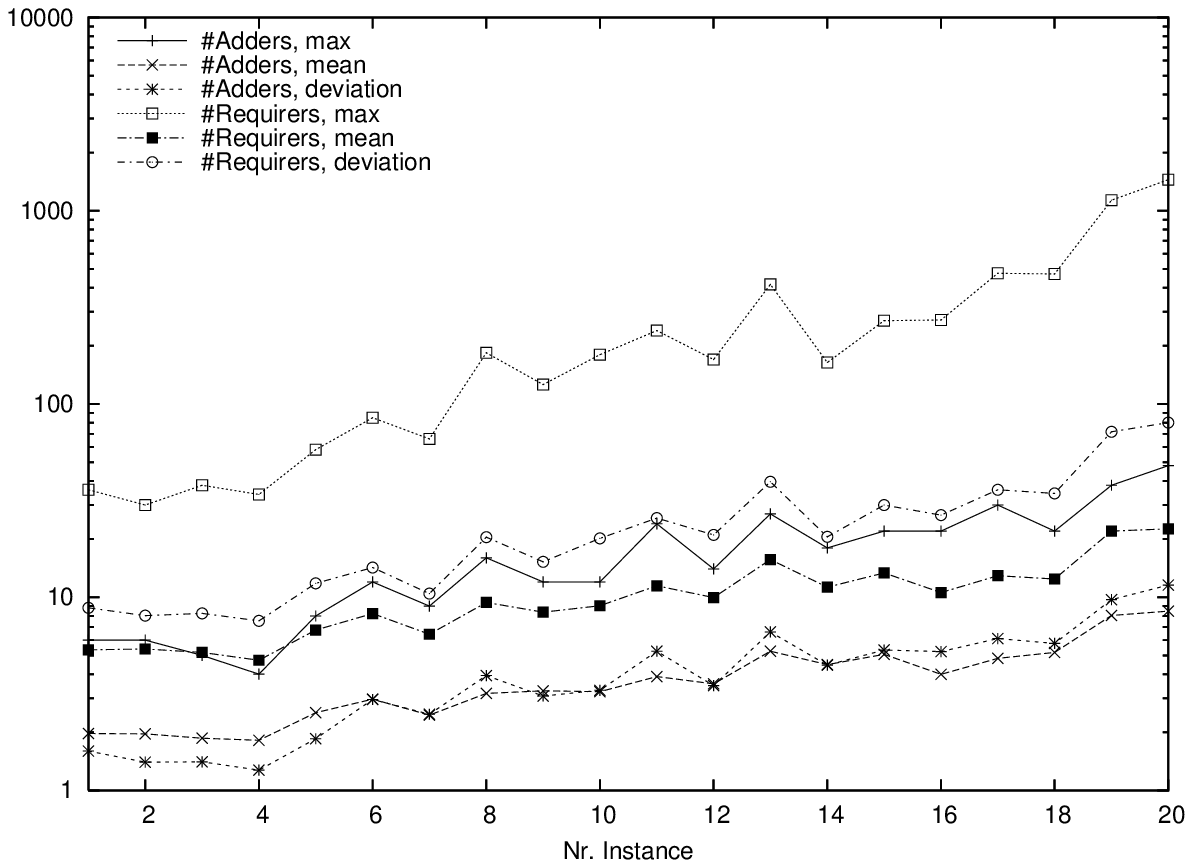} & \hspace{-0.3cm} \scaleonlyfig{0.58}{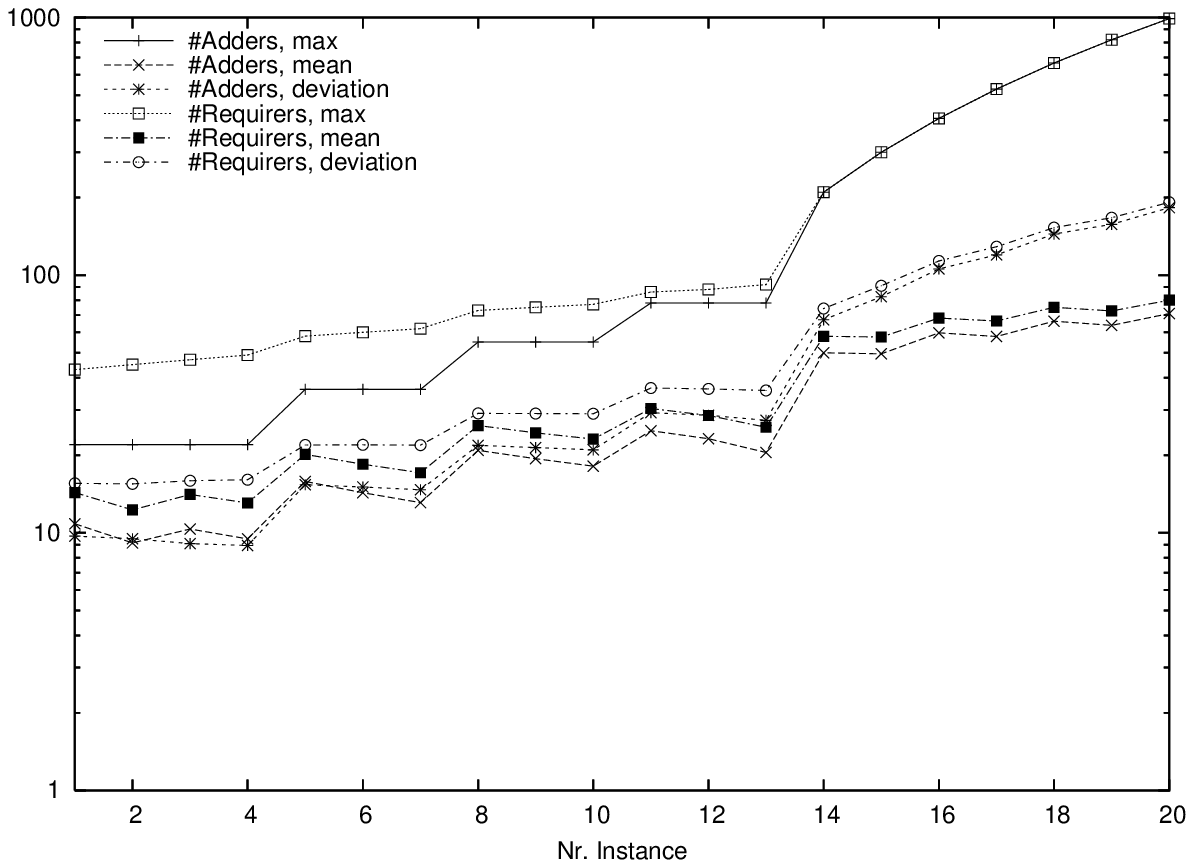}\\
(c) & (d)\\
\vspace{-0.8cm}
\end{tabular}
\caption{\label{new:connectivity:ipc3}Distributions of the numbers of
  actions adding a fact, and of actions requiring a fact, in the
  STRIPS versions of the IPC-3 domains except Freecell and Satellite:
  (a) Depots, (b) Driverlog, (c) Rovers, (d) Zenotravel.}
\vspace{-0.4cm}
\end{figure}

To sum up the sub-section, the data are, generally, too abstract to be
really tightly interconnected with the performance exhibited by
planners. On the other hand, certain characteristics are visible. Most
particularly: In Pipesworld, the numbers of adders and requirers are
almost identical. In Promela, the adders are constant and the
requirers are linear. In Satellite, all curves are very close
together. In PSR ``small'' there is a lot of variance, and the $max$
numbers of adders and requirers approach and sometimes exceed two
thirds of the total number of actions. In contrast, in PSR ``large''
the $max$ adders decline to less than half a percent of the total
number of actions. In UMTS, all the parameters are constant. Except
for PSR and UMTS, these phenomena are somewhat hard to interpret. If
nothing else, they certainly show us that the domains have some rather
different characteristics. Interestingly, the differences are not as
significant for the IPC-3 benchmarks shown in Figure 13. Clearly, the
behavior is not as characteristically diverse as what we have just
seen for the IPC-4 domains. For all the four domains in
Figure~\ref{new:connectivity:ipc3}, we basically observe mostly
parallel lines that are pretty close together except for the $max$
lines, which are about an order of magnitude higher than the others.
The only striking feature is the zig-zag nature of the curves in
Depots. This is due to the scaling pattern: In the smallest instances,
the number of crates (blocks) grows continually up to 15 crates in
instance 6. Thereafter, there come blocks of 3 instances each, of
which the first has 6 crates, the second 10 crates, and the third 15
crates (across the blocks, other instance size parameters grow). This
means that the zig-zag shape of the curves corresponds exactly to the
zig-zag shape of the crate numbers.

Note that the behavior of the plots in
Figure~\ref{new:connectivity:ipc3} is similar to the behavior of the
plot for Satellite in Figure~\ref{new:connectivity:many} (d), in
particular for the first 20 instances. These were the instances posed
to the fully automated planners in IPC-3, as also shown in
Figure~\ref{new:connectivity:ipc3}. The only IPC-3 domain that truly
stands out in terms of the behavior of these curves is
Freecell.\footnote{It somehow makes sense that it's precisely this
  domain that stands out, as it is also intuitively different from the
  other domains. Most notably, deciding plan existence in Freecell is
  {\bf NP}-hard while it is easy in the other domains, c.f.\
  Section~\ref{known:complexity}.} There, we observe a phenomenon
similar to that of the Pipesworld in
Figure~\ref{new:connectivity:many} (b), where the curves for adders
and requirers almost coincide. The phenomenon is a little weaker than
in Pipesworld: in the largest Freecell instance, number 20, the $max$
of (both) adders and requirers is $1638$, while the $max$ of the
difference is $102$, with a $mean$ of $14.30$ and $dev$ of $24.86$.
For comparison, in the largest Pipesworld instance, $max$ adders is
$1524$, $max$ requirers is $1520$, and the $max$ of the difference is
$29$, with a $mean$ of $1.63$ and $dev$ of $5.31$.

To sum up the overall empirical analysis, the data certainly don't
solve the mystery of what is behind the performance of every planner
in every domain (and instance). They do, however, provide some
interesting insights about how instances are scaled in the domains,
about certain subtleties and peculiarities of their encodings, and
about how standard heuristic methods, and groups of planners, react to
them. We can observe large characteristic differences between the
domains. In that sense the results nicely complement the technical
descriptions in Appendix~\ref{domains}, as well as the known
theoretical results from Section~\ref{known}.

%%% Local Variables: 
%%% mode: latex
%%% TeX-master: t
%%% End: 

\section{Conclusion}
\label{conclusion}

In a field of research about general reasoning mechanisms, such as AI
planning, it is essential to have useful benchmarks: benchmarks that
reflect possible applications of the developed technology, and that
help drive research into new and fruitful directions. In the
development of the benchmark domains and instances for IPC-4, the
authors have invested significant effort into creating such a set of
useful benchmarks for AI planning.

As explained in the introduction, the three main goals we tried to
achieve were 1. realism, 2. structural diversity, and 3. accessibility
of the benchmarks. It is debatable to what extent these goals were
achieved. To some extent, this is inherent in the conflicting nature
of the goals. Accessibility of a benchmark -- formulation in as simple
as possible PDDL dialects -- is obviously in conflict with realism.
Structural diversity is also in conflict with realism since, in the
time window available to create a competition benchmark set, there may
not be (and has not been, in our case) a large set of suitable
applications to choose from. One must make do with what's available.
We stressed on realism since the lack of realism was traditionally
considered as one of the main weaknesses of AI Planning -- achieving
``just'' structural diversity and accessibility would, in fact, have
been comparatively easy (see also below). That said, to adapt the
applications for the IPC we had to make many significant
simplifications. Still, having derived the domains from applications,
one can expect that they capture some important features even after
simplification; on top of that, there is a clear path towards more
realism.

We believe that the domains constitute the best possible compromise
for IPC-4. To name the most distinguishing features of the domain set:
\begin{enumerate}
\item Airport, Pipesworld, PSR, and UMTS are derived directly from
  applications (Promela is a special case since the model checking
  instances we could encode are very simplistic). This was previously
  the case only for the Elevator domain (IPC-2) and the Rovers and
  Satellite domains (IPC-3).
\item The complexity of satisficing and optimal planning in the STRIPS
  domain versions covers the entire range {\bf P}, {\bf NP}, and {\bf
    PSPACE} -- deciding (bounded) plan existence is in {\bf P} for PSR
  and {\bf PSPACE}-complete for Airport and general Promela. We are
  not aware of a previous {\bf PSPACE}-complete STRIPS benchmark; the
  polynomial algorithm for finding plans in PSR is, in contrast to
  those for all the other STRIPS benchmarks with such algorithms,
  quite non-trivial.
\item In Hoffmann's \citeyear{hoffmann:jair-05} taxonomy of domain
  classes with different $h^+$ topology, the IPC-4 domains lie in
  classes with sparse coverage by previous benchmarks. In particular,
  none of our new domains has nearly as simple a topology as proved by
  Hoffmann for most of the traditional benchmarks. When taking into
  account that Pipesworld actions {\em can} be inverted in (not one
  but) two steps, each of the domains lies in a different class of
  Hoffmann's taxonomy, covering more classes (6) than any previous IPC
  benchmark set (3, 5, and 4 for IPC-1, IPC-2, and IPC-3,
  respectively). Dining Philosophers is exceptional in that it lies in
  a ``simple'' class but doesn't have a simple topology; Airport is
  exceptional in that it lies in a ``very hard'' class but is
  typically (in real-world instances) easy.
\item The behavior of the different kinds of planners in IPC-4 shows a
  lot of very characteristic patterns in the individual domains. In
  Airport, sheer size is the main obstacle. In Pipesworld,
  particularly with tankage restrictions, the known heuristic
  functions do very badly. In the Promela domains, the main obstacle
  is, in a lot of cases, the impossibility of compiling the PDDL
  description into a fully grounded simpler representation. In PSR,
  there is an extremely large amount of variance, and optimal planners
  perform just as well (or poorly) as satisficing planners. In UMTS,
  satisficing planners need no time at all.
\item At a very abstract level that just looks at the numbers of
  actions adding/needing each fact, the behavior of the domains is
  more characteristically diverse than that of the IPC-3 domains.
\item Last but not least, the STRIPS versions of our domains preserve
  much more of the original domain structure than what was previously
  the case. The IPC-2 STRIPS version of Elevator is hardly an elevator
  problem anymore, and the IPC-3 STRIPS versions of Satellite and
  Rovers are devoid of all of the more interesting problem
  constraints. In contrast, the STRIPS versions of Airport and Promela
  are semantically identical to the ADL versions, and the PSR STRIPS
  version, while pre-compiled a lot, still preserves much of the
  original difficulty of the domain (judging, e.g., by the behavior of
  the IPC-4 planners in it).
\end{enumerate}
Feature 1 is, obviously, a point for realism. Features 2 to 5 are
points for diverse structure; particularly Feature 4 shows how the
domains pose very different challenges to (current) planning
technology. Feature 6 is a point for realism combined with
accessibility. We would like to stress that accessibility in this
respect is really quite important. Of the 19 planners entered into
IPC-4, only 8 could handle (some) ADL features. Our compilation
approach enabled us to confront the other 11 planners with reasonably
realistic problems. That said, it certainly is debatable what role
STRIPS plays or should play for the community. Some people may say
that many of the core algorithms, e.g., planning graphs
\cite{blum:furst:ai-97} and relaxed plan heuristics
\cite{mcdermott:ai-99,bonet:geffner:ai-01,hoffmann:nebel:jair-01},
have been invented in STRIPS. Others may say that the focus on
STRIPS-like languages and algorithms distracts us from considering
temporal and numerical problems of a truly different nature. This
notwithstanding, STRIPS is still the most widely used language among
the research community. This cannot be ignored by competition
organizers.

Having pointed out the advantages of our benchmark set, we should also
point out a few of the disadvantages. As explained in detail in the
individual sections in Appendix~\ref{domains}, we had to make many
simplifications in order to make the applications fit for use in
IPC-4. To some extent, whether or not the simplifications preserve the
original domain structure is a debatable matter. We feel that our
Airport encoding is very close to the real ``physical'' thing. Not
being able to represent the real optimization criterion is bad, but
ameliorated by the fact that, out of 19 planners, only a single one
(LPG-td) could actually deal with user-defined optimization
criteria.\footnote{This is a good example of a case where PDDL has
  been moving faster than the actual planning technology.} In
Pipesworld, the simplifications are more severe. The IPC-4 domain
still resembles some of the core difficulties, but is more reminiscent
of a (complicated) toy example than of software that could be used to
control real pipelines. The Promela examples go to show that toy
examples in the model checking area are not any better than the
traditional toy examples in planning. In PSR, removing the uncertainty
and the numerical optimization renders the IPC-4 domain unsuitable for
practical use.

Of course, the domain set is not exhaustive, meaning that there
presumably are numerous applications whose essential structure is not
similar to any of the IPC-4 domains. Some examples that spring to mind
are action choice in autonomous robots, detecting security holes in
computer networks \cite{boddy:etal:icaps-05}, and online manufacturing
\cite{ruml:etal:icaps-05}. As for structural diversity, it would be
easy to construct a set of artificial domains that explore more of the
possible extreme cases. Such domains would probably be completely
infeasible for current planners, thus posing very strong challenges.
Just think of, for example, Rubik's Cube, Sokoban, or Rintanen's
\citeyear{rintanen:icaps-04} purely randomly generated instance
distributions. Then again, such a domain set would be devoid of
realism. At some point during the preparation of IPC-4, we considered
introducing a separate class of domains, called ``Diverse Structure'',
which would have contained domains of this sort. We decided to not do
so since the competition event was already very large without it.
Also, we felt that our applications were already quite diverse on the
structural side. As pointed out above, several theoretical and
empirical phenomena suggest that the latter is indeed the case.

During our work, we experienced various successes and failures in
accurately formulating our application domains in PDDL. People have
asked us if, through this, we obtained a picture of how suitable PDDL
is, in its current form, to formulate applications, and in what sorts
of domains it works well. The answer is, we don't feel like we
obtained many insights into these matters that are particularly deep
or haven't been known before. A few lessons learned are these. First
and foremost, formulating an application in STRIPS takes a huge amount
of engineering expertise unless one just drops all problem
constraints; some simplifications are unavoidable. Second, the
discrete nature of action instantiations in all previous IPC PDDL
dialects seriously impedes formulation of domains with continuous
aspects. A discretization must be chosen, which is sometimes easy
(Airport) and sometimes very hard (Pipesworld) to do. A good way out
seems to be to adopt the ``duration inequalities'' suggested by
\citeA{fox:long:jair-03}. Third, the community should pay more
attention to lifted encodings, and how to deal with them in modern
planning algorithms: one lesson from our compilation activities is
that grounding out all parameters is often simply not possible
(Promela, PSR). Since compiling away ADL constructs is often not
feasible without grounding (c.f. Section~\ref{compilations}), this is
also very relevant in the ADL/STRIPS context. As a final ``lesson'',
we (the AI Planning community) are still, mostly, far away from as-is
applicability of planners in the real world. But we are on the right
track.

To conclude, we spent significant time and effort creating a useful
set of planning benchmarks for IPC-4. We hope that they will become
standard benchmarks in the coming years.

%%% Local Variables: 
%%% mode: latex
%%% TeX-master: t
%%% End: 

\bigskip

\noindent {\large {\bf Acknowledgements.}} We would like to thank the
competitors for their detailed comments about bugs found in our
domains, and we would like to thank Malte Helmert for various useful
tools that helped remove some of these bugs.

We further thank Malte Helmert for providing us with his -- yet
unpublished, at the time of writing -- results on computational
complexity \cite{helmert:personnal-05,helmert:icaps-06}. We thank
Patrik Haslum for providing us with the TP4 temporal numerical plan
graph estimates of makespan in the UMTS domain. We are indebted to the
anonymous reviewers, and very much to David Smith and Maria Fox, whose
detailed and extensive comments contributed greatly to the development
of this paper. We finally thank David Smith for his extensive advice
on language, including some corrections even for these very
acknowledgements.

J\"org Hoffmann thanks Wolfgang Hatzack for his support in the
development of the Airport domain and benchmark instances.

Frederico dos Santos Liporace is supported by Conselho Nacional de
Desenvolvimento Cient\'ifico e Tecnol\'ogico, Brazil. He would like to
acknowledge the support of his PhD supervisor, Ruy Milidiu, in the
development of the Pipesworld application.

Sylvie Thi\'ebaux thanks Piergiorgio Bertoli, Blai Bonet, and John
Slaney for their contributions to the development of the PSR domain
and instances. She also would like to acknowledge the support of
National ICT Australia. NICTA is funded through the Australian
Government's {\it backing Australia's Ability} initiative, in part
through the Australian Research Council.

\begin{appendix}

\section{Detailed Domain Descriptions}
\label{domains}

We now provide detailed descriptions of all the domains, in
alphabetical order. Each section (except those for the Satellite and
Settlers domains, which were adapted from the IPC-3) is organized in
sub-sections as follows. We first give an outline of the application
domain. We then explain the main adaptations made to model the
application as a PDDL domain in IPC-4, we explain the IPC-4 domain
structure, i.e., the domain versions and their formulations as used in
IPC-4, and we explain how we generated the example instances for the
IPC-4 test suites. Finally, we discuss possible future extensions.

\subsection{Airport}
\label{airport}

We had a contact person for this application domain, Wolfgang Hatzack,
who has been working in this application area for several years. The
domain was adapted for IPC-4 by J\"org Hoffmann and Sebastian Tr\"ug.

\subsubsection{Application Domain}
\label{airport:application}

The task is to control the ground traffic on an airport.  Timed travel
routes must be assigned to the airplanes so that they reach their
targets. There is inbound and outbound traffic; the former are
airplanes that must take off (reach a certain runway), the latter are
airplanes that have just landed and have to get parked (reach a
certain parking position). The main problem constraint is, of course,
to ensure the safety of the airplanes. This means to avoid collisions,
and also to prevent airplanes from entering the unsafe zones behind
large airplanes that have their engines running. The optimization
criterion is to minimize the summed up travel time (on the surface of
the airport) of all airplanes.\footnote{This criterion is what {\em
    the airport} wants to minimize, in order to maximize its
  throughput. From the point of view of the airlines, it would be
  better to minimize delay, e.g., by minimizing the summed up squared
  delay of all airplanes. The two criteria may be in conflict. Neither
  of the two can be easily modelled in PDDL2.2, see below.} There
usually are {\em standard routes}, i.e., routes that any airplane
outbound from a certain park position area, or inbound from a certain
runway, {\em must} take. The reason for introducing such routes is,
simply, the sheer complexity of managing the situation otherwise,
without significant computer support (which is as yet not available on
real airports). We will see below that whether or not standard routes
are present makes a big difference also computationally.

The airplanes move on the airport infrastructure, which consists of
runways, taxiways, and parking positions. The runways and taxiways are
sub-divided into smaller {\em segments}. The position of an airplane
is given by the segment it is currently located in, plus its direction
and the more precise position {\em within} the segment -- several
airplanes can be in the segment at the same time.

Airplanes are generally divided into three categories, {\em light,
  medium, and heavy}, which classify them according to their engine
exhaust ({\em jet blast}). An airplane that has to be moved is either
in-bound or out-bound.  In-bound airplanes have recently landed and
are on their way from the runway to a parking position, usually a
gate.  Out-bound airplanes are ready for departure, meaning they are
on their way to the departure runway.  Since airplanes cannot move
backwards, they need to be pushed back from the gate onto the taxiway,
where they start up their engines. Some airports also provide
different park positions that allow an airplane to start its engines
directly.

To ensure safety, an airplane must not get too close to the back of
another airplane whose engines are running. How far the safety
distance has to be depends on the category (jet blast) of the second
airplane.

The ground controller -- the planner -- has to communicate to the
airplanes which ways they shall take and when to stop. While such
guidance can be given purely reactively, it pays off to base decisions
on anticipating the future. Otherwise it may happen that airplanes
block each other and need more time than necessary to reach their
destinations on the airport. The objective is, as said, to minimize
the overall summed up traveling times of all airplanes.

As instances of the domain, one considers the traffic situation at
some given point in time, with a time horizon of, say, one hour. If
new airplanes are known to land during given time slots inside the
time horizon, then during these time slots the respective runways are
considered blocked, and the planner has to make sure these runways are
free at these times. Of course, because the situation changes
continually (new planes have to be moved and plans cannot be executed
as intended), continuous re-planning, i.e., consideration of the
domain instance describing the new traffic situation, is necessary.
Solving instances optimally (the corresponding decision problem) is
{\bf PSPACE}-complete without standard routes \cite{helmert:icaps-06}
and {\bf NP}-complete if {\em all} routes are standardized
\cite{hatzack:nebel:ecp-01}. In the latter case, we have a pure
scheduling problem. In the former case, complicated (highly
unrealistic, of course) airport topologies can lead to exponentially
long solutions, c.f.\ Section~\ref{known:complexity}.

% An instance
% of the domain is given by the airport road map topology, the current
% positions of all airplanes, as well as the inbound traffic (planes
% that need to go to a parking position) and outbound traffic (planes
% that need to go to a runway). The task is to find a plan of movement
% actions that brings all the airplanes safely to their destinations.
% The safety constraints that must be adhered to are the following. The
% airport road map is divided into segments. No two planes can share a
% segment at any point in time. Moreover, a plane must not get too close
% to the back of another airplane whose engines are running. How far the
% safety distance has to be depends on the size category of the second
% airplane.

\subsubsection{IPC-4 PDDL Adaptation}
\label{airport:adaptation}

The PDDL encoding (as well as our example instance generation process,
see below) is based on software by Wolfgang Hatzack, namely on a
system called {\em Astras}: Airport Surface ground TRAffic Simulator.
This is a software package that was originally designed to be a
training platform for airport controllers.  Astras provides a
two-dimensional view of the airport, allowing the user to control the
airplanes by means of point and click. Astras can also simulate the
traffic flow on an airport over the course of a specified time window.

We made three simplifications, one of them benign, to the airport
model.  As for the benign simplification: we did not model park
positions where the airplane can start up its engines directly,
without being pushed back to the taxiway first. While it is not
difficult to model such park positions in PDDL, they seldom occur in
reality and so are not very relevant to the application. Our first
more important simplification was to assume a somewhat cruder notion
of airplane locatedness, by requiring that only a single airplane can
be located in a segment at any time.  That is, we use the term
``segment'' with the meaning of a smallest indivisible unit of space.
To minimize the loss of precision, (some of) the original ``segments''
were sub-divided into several new smaller segments. The safety
distance behind the back of an airplane whose engines are running is
then also measured in terms of a number of segments. While this
discretization makes us lose precision, we believe that it does not
distort the nature of the problem too much: due to the amount of
expected conflicting traffic at different points on the airport (high
only near parking positions), it is relatively easy to choose a
discretization -- with segments of {\em different} length -- that is
precise and small enough at the same time.\footnote{The need for
  smallest indivisible units (of space, in this case) is a fundamental
  consequence of the discrete nature of PDDL2.2; some more on this is
  said in Section~\ref{airport:future}.} The last simplification is
more severe. We had to give up on the real optimization criterion. We
say more on this rather strong simplification below. We did not use
full standard routes, thus allowing the airplanes a choice of where to
move. We {\em did} use standards for some routes, particularly the
regions near runways in large airports. For one thing, this served to
keep large airports manageable for the PDDL encoding and planners; for
another thing, it seems a good compromise at exploiting the
capabilities of computers while at the same time keeping close to
traditions at airports. We get back to this matter in
Section~\ref{airport:future}.

The full PDDL description of our domain encoding can be downloaded
from the IPC-4 web page at http://ipc.icaps-conference.org/. Briefly,
the encoding works as follows. The available actions are to
``pushback'' (move a plane away backwards from a parking position), to
``startup'' the engines, to ``move'' between segments, to ``park''
(turning off the engines), and to ``takeoff'' (which amounts to
removing the plane from the airport).  The semantics of these actions
are encoded based on predicates defining the current state of the
airplane. At any point in time, an airplane is either moving, pushed,
parked, or airborne. An airplane always occupies one segment and, if
its engines are running, may block several other segments depending on
the size of the occupied segment and the category of the airplane. The
action preconditions ensure that blocked segments are never occupied
by another airplane. In the initial state, each plane is either
parked, or moving. A parked plane can be pushed back, and after
starting up its engines, it is moving. A moving airplane can either
move from its current segment to a neighboring segment, park -- at a
parking position -- or take off -- on a runway.

As an example, have a look at the PDDL encoding of the
(non-durational) ``move'' action (one of the preconditions was used as
an example in Section~\ref{compilations} already):

{\small
\begin{tabbing}
(\=:action move\\
\> :parameters\\
\> (?a - airplane ?t - airplanetype ?d1 - direction ?s1 ?s2  - segment ?d2 - direction)\\
\> :precondition\\
\>  (and \= (has-type ?a ?t) (is-moving ?a) (not (= ?s1 ?s2)) (facing ?a ?d1) (can-move ?s1 ?s2 ?d1)\\
\> \>       (move-dir ?s1 ?s2 ?d2) (at-segment ?a ?s1)\\
\> \>       (not (exists (?a1 - airplane) (and (not (= ?a1 ?a)) (blocked ?s2 ?a1))))\\
\> \>       (forall (?s - segment)  (imply  (and  (is-blocked ?s ?t ?s2 ?d2) (not (= ?s ?s1))) (not (occupied ?s)))))\\
\> :effect\\
\>  (and \> (occupied ?s2) (blocked ?s2 ?a) (not (occupied ?s1)) (not (at-segment ?a ?s1)) (at-segment ?a ?s2)\\
\> \>       (when   (not (is-blocked ?s1 ?t ?s2 ?d2)) (not (blocked ?s1 ?a)))\\
\> \>       (when   (not (= ?d1 ?d2)) (and (not (facing ?a ?d1)) (facing ?a ?d2)))\\
\> \>       (forall (?s - segment)  (when   (is-blocked ?s ?t ?s2 ?d2) (blocked ?s ?a)))\\
\> \>       (forall \= (?s - segment) (when\\  
\> \> \>    (and  (is-blocked ?s ?t ?s1 ?d1) (not (= ?s ?s2)) (not (is-blocked ?s ?t ?s2 ?d2)))\\
\> \> \>    (not (blocked ?s ?a))))))
\end{tabbing}}

The six parameters -- which is a lot compared to most of the usual
benchmarks -- do not cause a prohibitive explosion in instantiations
since there is a lot of restriction through static predicates.
Airplane ``?a'' moves; its type (category) is ``?t''; it is at segment
``?s1'' facing in direction ``?d1'', and will be at ``?s2'' facing in
direction ``?d2'' after the move. ``Direction'' here is a very simple
concept that just says which end of the segment the airplane is
facing. Of course, moves from ``?s1 ?d1'' to ``?s2 ?d2'' are only
possible as specified by the -- static -- topology of the airport
(``can-move'', ``move-dir''). The first of the two more complex
preconditions says that ``?s2'' must not currently be blocked by any
airplane other than ``?a'' itself. The second complex precondition
makes sure that, after the move, ``?a'' will not block a segment that
is currently occupied (by another airplane, necessarily):
``(is-blocked ?s ?t ?s2 ?d2)'' is a static predicate that is true iff
``?s'' is endangered -- blocked -- if a plane of type ``?t'' is at
``?s2''w facing direction ``?d2''. The effects should be
self-explanatory; they simply update the ``at'', ``occupied'', and
``blocked'' information. The only effect that looks a little
complicated -- the last one -- says that those segments that were
blocked before the move, but are no longer blocked after the move,
become un-blocked. Note that the conditions of all conditional effects
are static, so the conditions disappear once the parameter
instantiation is chosen.

In durational PDDL, the actions take time according to some simple
computations. The time taken to move across a segment depends,
naturally, on the segment length and the speed. We assumed that
airplanes move at the same speed regardless of their category. The
time taken to start up the engines is proportional to the number of
engines. The other actions have some fixed duration.

If some planes are known to land in the near future, blocking runways,
then we model the blocking during these time windows using timed
initial literals, respectively their compilation into artificial
(temporal) PDDL constructs. The timed literals are simply instances of
the usual ``blocked'' predicate, becoming true when the respective
time window starts, and becoming false again when it ends.

We were not able to model the real optimization criterion of airport
ground traffic control. The standard criterion in PDDL is to minimize
the execution time, i.e., makespan, of the plan. In our encoding of
the domain this comes down to minimizing the arrival time (meaning,
arrival at the destination on the airport) of the last airplane. But
the real objective is, as said above, to minimize the overall summed
up travel time of all airplanes. There appears to be no good way of
modeling this criterion in current PDDL. The difficulty lies in
accessing the waiting times of the planes, i.e. the times at which
they stay on a segment waiting for some other plane to
pass.\footnote{Modelling summed up (squared) delay of all airplanes,
  the optimization criterion for airlines, would pose essentially the
  same difficulty: it also involves computing the arrival time (in
  order to compute the delay).}

The only way (we could think of) to get access to the waiting times,
in current PDDL, is to introduce an explicit waiting action.  But then
one must be able to tell the planner, i.e., to encode in the action,
how long the plane is supposed to wait. One option is to use the
``duration inequalities'' proposed by \citeA{fox:long:jair-03}. There
the action imposes only some constraints on its duration, and the
planner can/has to choose the actual duration of the action, at each
point where it is used in the plan, as an additional (rational-valued)
parameter. The potential disadvantage of this approach is that the
choice of the waiting time introduces, in principle, an infinite
branching factor into the state space, and may thus make the problem
much harder for automated planners. Moreover, duration inequalities
were not put to use in IPC-3, and were not a part of PDDL2.1.
% (At the time we prepared the
% Airport domain for IPC-4, there were indeed no automated planning
% systems able to handle duration inequalities.) 
When not using duration inequalities, the only way to encode the
requested waiting time into the action is to use a discretization of
time. One can then introduce new objects representing every considered
time interval, and give the waiting action a parameter ranging over
these objects. Apart from the loss of precision involved in the
discretization, this approach is also likely to cause huge performance
problems for automated planners.  As an alternative way out, we
considered introducing a special ``current-time'' variable into
PDDL2.2, returning the time of its evaluation in the plan execution.
Using such a ``look at the clock'', one could make each plane record
its arrival time, and thus formulate the true optimization criterion
without any major changes to the domain structure. The IPC-4
organizing committee decided against the introduction of a
``current-time'' variable as it seemed to be problematic from an
algorithmic point of view (it implies a commitment to precise time
points at planning time), and didn't seem to be very relevant anywhere
except in Airport.

All in all, the IPC-4 PDDL encoding of the Airport domain is realistic
except for the optimization criterion, which demands to minimize
maximal arrival time -- makespan -- instead of summed up travel time.
It remains to remark that all but one (LPG-td) of the IPC-4 planners
ignored the optimization criterion anyway. Also, minimizing the latest
arrival time does appear a useful (if not ideal) objective.

\subsubsection{IPC-4 Domain Structure}
\label{airport:structure}

\begin{sloppypar}
  The Airport domain versions used in IPC-4 are {\em non-temporal},
  {\em temporal}, {\em temporal-timewindows}, and {\em
    temporal-timewindows-compiled}.  The first of these versions is,
  as the name suggests, non-durational PDDL. In the second version,
  actions take time as explained above. The third and fourth versions
  also consider runways blocked in the future by planes known to land
  during given time windows. The third version encodes these time
  windows using timed initial literals, the fourth version uses those
  literals' compilation into standard temporal PDDL constructs, c.f.\
  Section~\ref{compilations}.
\end{sloppypar}

In all the domain versions, the problem constraints are modeled using
ADL, i.e., complex preconditions and conditional effects. We compiled
the ADL encodings to STRIPS with domain-specific software implemented
for this purpose. We grounded out {\em most} -- not all -- of the
operator parameters, precisely, all the parameters except, for each
action, the one giving the name of the affected individual airplane.
Once all the other parameters are fixed, the formulas and conditional
effects can be simplified to the usual STRIPS constructs. Each Airport
domain version contains the original ADL formulation, as well as its
compilation to STRIPS. The result of the grounding process depends on
the specific airport considered in the instance, and on the set of
airplanes that are travelling. So, in the STRIPS formulations, to each
instance there is an individual domain file (the same applies to all
STRIPS compilations in the other domains described later).

The domain versions, as well as the blow-up incurred by the
compilation, are overviewed in
Table~\ref{airport:versionstable}.\footnote{The instantiation process
  is, of course, planner-dependent. Similarly as before in
  Section~\ref{new}, our data are based on FF's pre-processor. We
  extended that pre-processor (precisely, the one of Metric-FF
  \cite{hoffmann:jair-03}) to deal with temporal constructs.} The
numbers shown in the table indicate numbers of PDDL operators, and
numbers of grounded actions.  For each domain version/formulation, the
maximum such number of any instance is shown. Note that, in the ADL
formulations except temporal-timewindows-compiled, there is just a
single domain file so the number of operators is identical for all
instances. In the STRIPS formulations, the number of operators is high
because, as explained, most of the operator parameters are grounded.
The difference in the number of ground actions between the STRIPS and
the ADL formulations is because, with our automated software, we were
not able to generate the ground actions in the larger ADL instances;
the data shown are for the largest instances that we {\em could}
handle. The numbers shown in parentheses refer to the situation {\em
  before} FF's ``reachability'' pre-process; as said before, this
builds a relaxed planning graph for the initial state, and removes all
actions that do not appear in that graph. The difference between the
numbers inside and outside of the parentheses indicates how much this
simple pre-process helps. We see that it helps quite a lot here,
pruning almost half of the actions (which would never become
applicable, in a forward search at least, but which blow up the
representation regardless of what algorithm is used).

\begin{table}
\centering
\begin{small}
\centering
\begin{tabular}{|l|l||c|c|}
\hline  
version          & formulation &   $max\mbox{-}\#op$ &    $max\mbox{-}\#act$\\\hline\hline 
non-temporal     & ADL         &             5       &      (1048) 989   \\\hline 
non-temporal     & STRIPS      &           1408     &         (21120) 13100  \\\hline 
temporal         & ADL         &             5       &  (1408) 989        \\\hline 
temporal         & STRIPS      &           1408     &   (21120) 13100  \\\hline 
temporal-tw      & ADL         &             5       &          (995) 854         \\\hline 
temporal-tw      & STRIPS      &           1408     &        (22038) 13100  \\\hline 
temporal-twc     & ADL         &             14      &         (911)  861        \\\hline 
temporal-twc     & STRIPS      &           1429     &         (21141) 13121  \\\hline 
\end{tabular}
\end{small}
\caption{\label{airport:versionstable}Overview over the different 
  domain versions and formulations of Airport. Abbreviations used: 
  ``temporal-tw'' for ``temporal-timewindows'', ``temporal-twc'' for 
  temporal-timewindows-compiled; $max\mbox{-}\#op$ is the maximum number of 
  (parameterized) PDDL operators for any instance, $max\mbox{-}\#act$ is the 
  maximum number of ground actions for any instance. For the ADL formulations, 
  the set of ground actions could not be generated for the largest instances; 
  data are shown for the largest instances that could be handled. Data in 
  parentheses are collected before FF's ``reachability'' pre-process (see text).}
\vspace{-0.5cm}
\end{table}

\subsubsection{IPC-4 Example Instances}
\label{airport:examples}

\begin{figure}[htb]
\begin{center}
\begin{tabular}{c}
\scaleonlyfig{0.2}{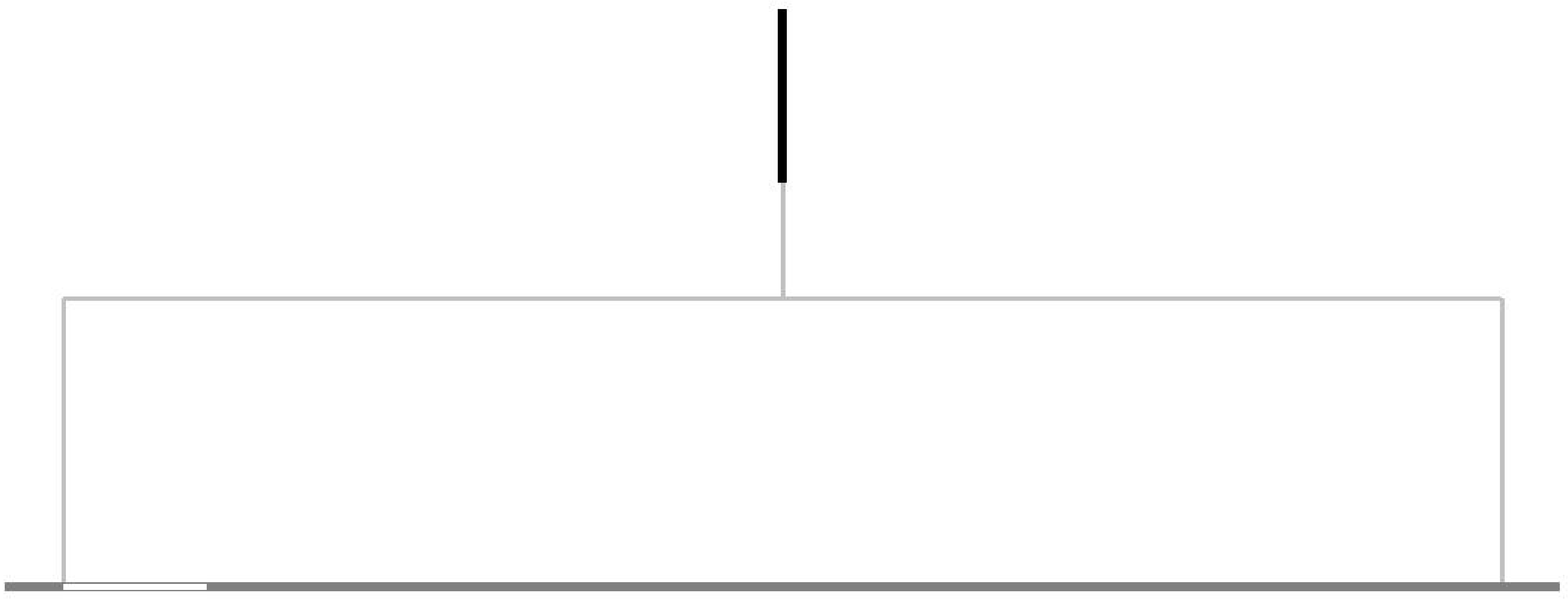}\\
(a)\\
\scaleonlyfig{0.3}{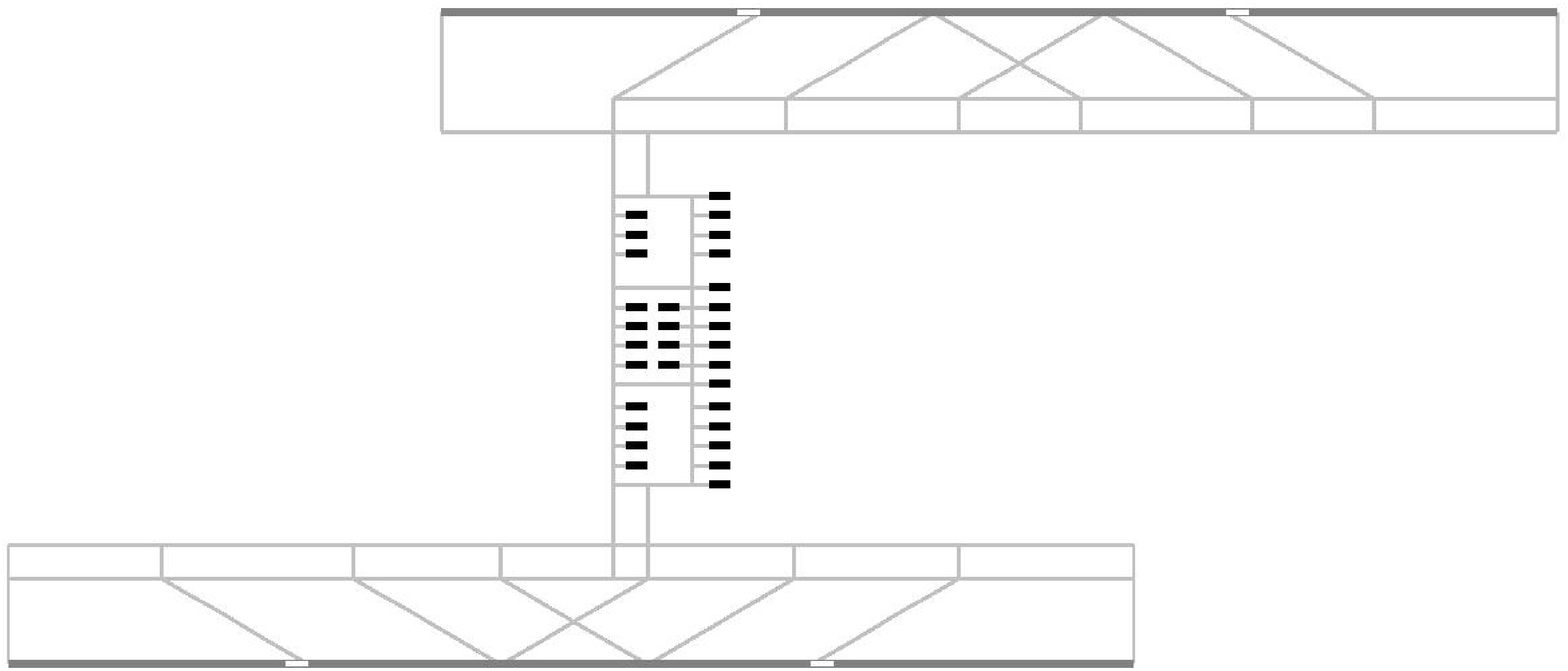}\\
(b)\\
\end{tabular}
\end{center}
\caption{\label{fig:airports} The smallest (a), and the largest (b) of
  the IPC-4 Airport topologies.  Park position segments are marked in
  black (e.g., at the top of part (a)), while the segments airplanes
  can takeoff from are marked in white (e.g., at the left bottom side
  of part (a)). The lines show the road network on the airport.
  Topology (b) corresponds to MUC airport.} \vspace{-0.5cm}
\end{figure}

The Airport example instances were generated by Sebastian Tr\"ug, with
an implementation based on the aforementioned airport simulation tool
Astras. Five scaling airport topologies were designed, and used as the
basis for the instance generation. The airports are named ``Minimal'',
``Mintoy'', ``Toy'', ``Half-MUC'', and ``MUC''. The smallest of these
airports is the smallest possible airport Astras can handle. The two
largest airports correspond to one half of Munich Airport (MUC), and
to the full MUC airport.  Figure~\ref{fig:airports} shows sketches of
the ``Minimal'' airport, and of the ``MUC'' airport.

Sebastian Tr\"ug implemented PDDL instance generation software inside
Astras. During a simulation of the traffic flow on an airport, if
desired by the user the software exports the current traffic situation
in the various PDDL encodings explained above. The simulator was run
with the different airports, and 50 scaling traffic situations were
exported (3 on ``Minimal'', 6 on ``Mintoy'', 11 on ``Toy'', 15 on
``Half-MUC'', and 15 on ``MUC''). For each airport, the instances
scale in terms of the number of travelling airplanes. The largest
instance features 15 planes to be moved to their destinations on
Munich airport, with 10 planes landing in the future to be considered
(in the respective domain versions). This can be considered a
realistically sized traffic situation, at this airport.

\subsubsection{Future Work}
\label{airport:future}

It remains to explore how to relax some of the simplifications we had
to make. Most importantly, how to overcome the discrete model of space
(locatedness), and how to model the real optimization criterion. Our
difficulties with both are, as partly described above already, mostly
due to the discrete nature of PDDL2.2, which does not allow a {\em
  continuous} choice in the instantiation of an action. Such a
continuous choice would be the most natural way of saying {\em how
  far} a plane will be moving and {\em how long} it will be waiting.
So the best way to go about this direction is, probably, to assume the
``duration inequalities'' proposed by \citeA{fox:long:jair-03},
together with the numeric variables already contained in PDDL2.2. This
should be easy on the modelling side. The main problem is probably on
the technology side, i.e., to develop planners that can deal
efficiently with such continuous choice points. At the time of IPC-4,
as said, continuous choice appeared too much to demand from the
planners.

One interesting topic for future work arises if one restricts the
airplanes completely to standard routes, i.e., leaves them no choice
at all of what route to take to their destination. As said, first,
this is usually done at real airports, for the sheer complexity of
managing the situation otherwise, without significant computer support
(which is as yet not available at real airports).  Second, in IPC-4 we
made only limited use of this feature, to retain some of the
flexibility that could be offered by automatized methods.  Third, the
restriction turns the {\bf PSPACE}-complete ground traffic control
problem into a pure, {\bf NP}-complete \cite{hatzack:nebel:ecp-01},
scheduling problem, where the only question is {\em when} the planes
move across what segment. One could exploit this to create a much more
concise PDDL encoding. The restricted problem comes down to resolving
all {\em conflicts} that arise when two planes need to cross the same
airport segment. One could thus try to not encode in PDDL the physical
airport, but only the conflicts and their possible solutions, ideally
in connection with the real optimization criterion.  It can be
expected that planners will be much more efficient in such a simpler
and more concisely encoded problem.

%%% Local Variables: 
%%% mode: latex
%%% TeX-master: t
%%% End: 

\subsection{Pipesworld}
\label{pipesworld}

% REVISION: JOERG: revise formulations and insert ``snippets''

Frederico Liporace has been working in this application area for
several years; he submitted a paper on an early domain version to the
workshop on the competition at ICAPS'03.  The domain was adapted for
IPC-4 by Frederico Liporace and J\"org Hoffmann.

\subsubsection{Application Domain}
\label{pipesworld:application}

Pipelines play an important role in the transportation of Petroleum
and its derivatives, since it is the most effective way to transport
large volumes over large distances. The application domain we consider
here deals with complex problems that arise when transporting oil
derivative products through a multi-commodity pipeline system. Note
that, while there are many planning benchmarks dealing with variants
of transportation problems, transporting oil derivatives through a
pipeline system has a very different and characteristic kind of
structure, since it uses stationary carriers whose cargo moves rather
than the more usual moving carriers of stationary cargo. In
particular, {\em changing the position of one object directly results
  in changing the position of several other objects.} This is less
reminiscent of transportation domains than of complicated
single-player games such as Rubic's Cube. It can lead to several
subtle phenomena. For example, it may happen that a solution must
reverse the flow of liquid through a pipeline segment several times.
It may also happen that liquid must be pumped through a ring of
pipeline segments in a cyclic fashion, to achieve the goal (we will
see an example of this later).

In more detail, the application domain is the following. A pipeline
network is a graph of operational areas connected by pipeline
segments. Operational areas may be harbors, distribution centers or
refineries. These may be connected by one or more pipeline segments.
The oil derivatives are moved between the areas through the pipelines.

There can be different types of petroleum derivative products. Each
area has a set of tanks that define the storage capacity for each
product type. Each pipeline segment has a fixed volume and speed. The
volume depends on the segment's length and cross section diameter, and
the speed depends on the power of the pumps that move the contents. A
segment may be uni-directional, i.e. only usable for transportation
in one direction.

Pipeline segments are always pressurized, that is, they must be always
completely filled with petroleum derivative products. Because of that,
the only way to move a pipeline segment's contents is by pumping some
amount of product from an adjacent area into the segment. This
operation results, assuming incompressible fluids, in the same amount
of a possibly different product being received in the area at the
other end of the segment.

The pumping operations can only be executed if they do not violate any
interface or tanking constraints. As for the former, distinct products
have direct contact inside the pipeline segment, so it is unavoidable
that there is some loss due to the mixture in the interface between
them. These interface losses are a major concern in pipeline
operation, because the mixed products can not be simply discarded.
They must pass through a special treatment that may involve sending
them back to a refinery, and that may require the use of special
tanks. The severity of interface losses depends on the products that
interface inside the pipeline segment. If two product types are known
to generate high interface losses, the pipeline plan must not place
them adjacently into the segment. Such a pair of product types is said
to have an interface restriction.

Tanking constraints are limits on the product amounts that can be
stored in an area, arising from the respective tank capacities. Such
constraints may effectively block a pipeline segment, if there is no
room in the receiving area to store the product that would leave the
segment in the process of a pumping operation.

The task in the application is to bring certain amounts of products to
the areas in which they are required, i.e. one has to find a plan of
pumping operations that shifts the positions of the product amounts in
a way so that the goal specifications are met. Sometimes there is a
deadline specifying when, at the latest, a product amount has to
arrive at its destination area. It may also be the case that an area
(typically, a refinery) is known to produce some given amount of a
product at a given point in time, and that the plan must make sure
that there is enough tank space available at the respective area to
store the new product amount. Similarly, an area (typically, a harbor
or a distribution center) may be known to consume some given amount of
a product at a given point in time, thereby freeing the respective
amount of tank space.

\subsubsection{IPC-4 PDDL Adaptation}
\label{pipesworld:adaption}

The main adaptations made in the PDDL encoding are unitary batches,
split pumping operations, and ``personalized'' goals (see below for
the latter). The term ``batch'' is used in the oil pipeline industry
to refer to an amount of a product that must be transported through
the pipeline.  Batches are thus associated with a single product and
have predefined volume.  Batches are also indivisible. When a batch
$B_i$ is pumped from an area $A_j$ into a segment $S_{j,k}$, it is not
possible for another batch to be pumped from $A_j$ into $S_{j,k}$
until all of $B_i$'s volume is pumped. Of course, in reality the
product amount in a batch is a rational number.  Using such a numeric
encoding in IPC-4 seemed completely infeasible due to complications in
the modeling, and the expected capabilities of the participating
planners (see Section~\ref{pipesworld:future}).  Instead, we based the
encoding on the concept of what we called unitary batches.  These are
the smallest considered -- indivisible -- portions of product. The
pumping operations refer to unitary batches.  The pipeline segments'
volumes and the volumes of tanks are also defined in terms of unitary
batches.  When encoding a real-world instance of the domain, the
actual volume associated with a unitary batch is a choice variable.
Smaller unitary batches decrease the rounding error in the PDDL
encoding, at the cost of a larger encoding size. Note that, like the
smallest units of space in the Airport domain, this is a
discretization the need for which is due to the non-continuous nature
of actions in PDDL2.2; we get back to this in
Section~\ref{pipesworld:future}.

We modeled pipe segments in a directional fashion, i.e. there is a
default direction assigning one area the ``from'' role, and the other
area the ``to'' role. The pumping operations accordingly distinguish
between ``push'' actions, which move liquid in the respective
segment's default direction, and ``pop'' actions, which move liquid in
the opposite direction. This is simply a technical device to enable
the encoding of the pipe segment contents through predicates defining
the ``first'' and ``last'' batches in the segments (as well as a
``successor'' relation). The ``push'' and the ``pop'' actions receive
(amongst other things) as arguments the pipeline segment whose
contents are being moved, and the batch that is being inserted into
the segment. The batch that leaves the segment depends on the segment
content before the action is executed.
Figure~\ref{XFIG/push-pop-example.xfig.eps} shows an example.

\scalefig{0.7}{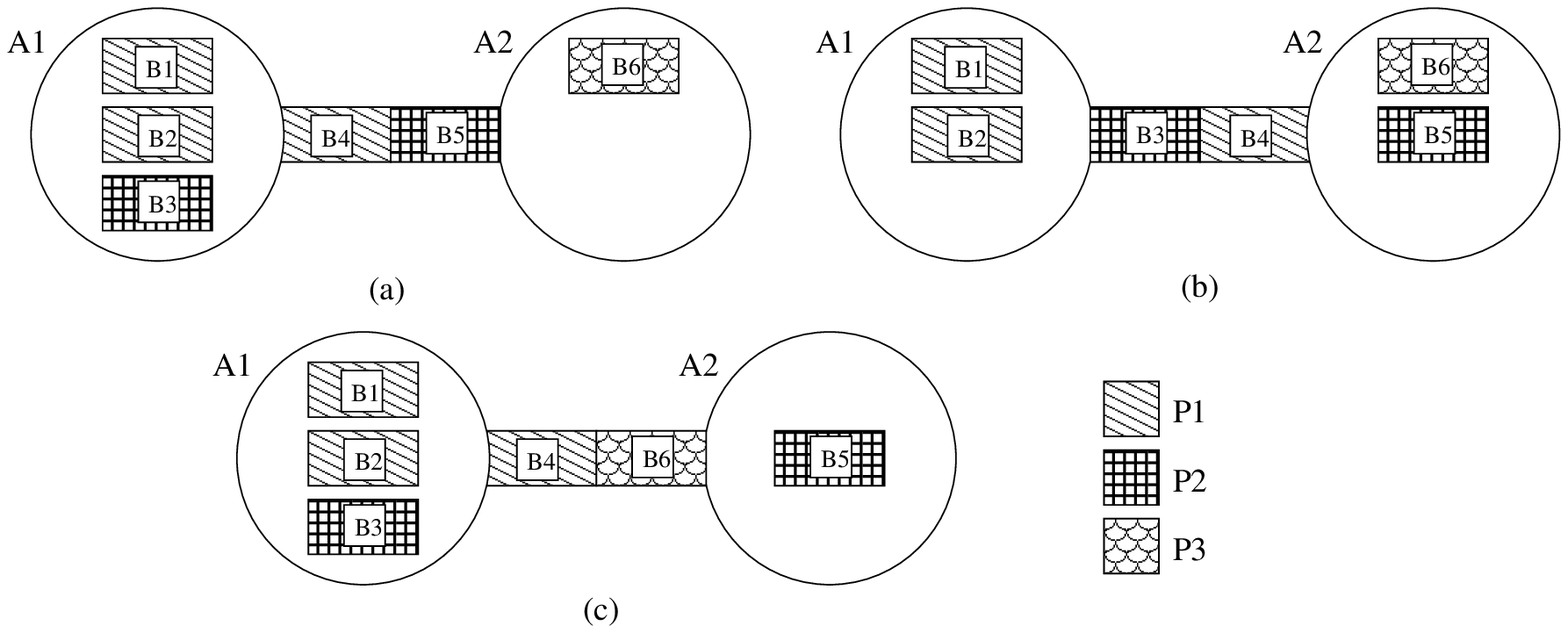}{A small example. A1
  plays the ``from'' role. The fill pattern for each batch represents
  its product. (a) shows the initial state, (b) shows state (a) after
  a ``push'' operation with B3 being inserted into the segment, (c)
  shows state (b) after a ``pop'' operation with B6 being inserted
  into the segment.}

Apart from the pipe segment and the batch being inserted, the ``push''
and ``pop'' actions have to take several more parameters regarding,
e.g., product types and tank slots. In particular, in order to be able
to update the segment contents correctly, the actions also need
parameters giving the respective first, last, and second last batch in
the current contents of the segment. Thus such an action has four
parameters ranging over batches, yielding at least $n^4$ ground
instances of the action when there are $n$ (unitary) batches in the
considered task. We found that this made the domain completely
infeasible for any planning system that grounded out the actions.
Since many unitary batches are needed to encode even relatively small
Pipesworld examples, such planners typically died in the
pre-processing phase already.\footnote{Matters may be easier for
  planning systems that do not ground out actions in a pre-process.
  This didn't affect our design decision here since the large majority
  of systems around at the time of IPC-4 {\em did} employ such a
  pre-process.} We avoided this phenomenon by splitting the actions
into two parts, a ``start'' action taking as batch parameters only the
inserted batch and the first batch in the pipe, and an ``end'' action
taking as batch parameters only the last and second last batches in
the pipe. To make this more concrete, here is the split ``push''
action:

{\small
\begin{tabbing}
(\=:action PUSH-START\\
 \> :parameters\\
\> (?pipe - pipe ?batch-atom-in - batch-atom ?from-area - area ?to-area - area\\
\>    ?first-batch-atom - batch-atom ?product-batch-atom-in - product\\
\> ?product-first-batch - product)\\
\> :precondition\\
\> (and \= (normal ?pipe) (first ?first-batch-atom ?pipe) (connect ?from-area ?to-area ?pipe)\\
\> \> (on ?batch-atom-in ?from-area) (not-unitary ?pipe)\\
\> \> (is-product ?batch-atom-in ?product-batch-atom-in)\\
\> \> (is-product ?first-batch-atom ?product-first-batch)\\
\> \> (may-interface ?product-batch-atom-in ?product-first-batch))\\
\> :effect\\
\>  (and \> (push-updating ?pipe) (not (normal ?pipe)) (first ?batch-atom-in ?pipe)\\ 
\> \> (not (first ?first-batch-atom ?pipe)) (follow ?first-batch-atom ?batch-atom-in)\\
\> \> (not (on ?batch-atom-in ?from-area))))\\
(\>:action PUSH-END\\
\> :parameters\\ 
\> (?pipe - pipe ?from-area - area ?to-area - area ?last-batch-atom - batch-atom\\ 
\> ?next-last-batch-atom - batch-atom)\\
\>  :precondition\\
\> (and \= (push-updating ?pipe) (last ?last-batch-atom ?pipe) (connect ?from-area ?to-area ?pipe)\\
\> \> (not-unitary ?pipe) (follow ?last-batch-atom ?next-last-batch-atom))\\
\> :effect\\
\> (and \> (not (push-updating ?pipe)) (normal ?pipe)\\
\> \> (not (follow ?last-batch-atom ?next-last-batch-atom))\\
\>  \>   (last ?next-last-batch-atom ?pipe) (not (last ?last-batch-atom ?pipe))\\
\> \> (on ?last-batch-atom ?to-area)))
\end{tabbing}}

The constructs should be largely self-explanatory. The static
predicates used are: ``connect'', encoding the topology of the
network; ``is-product'', encoding the types of liquid;
``may-interface'', encoding the interface restrictions;\footnote{Note
  here that we do {\em not} model the interface loss for those products
  that {\em may} interface.} ``not-unitary'', saying whether or not a
pipe segment contains just one batch -- in which case the ``push'' and
``pop'' actions are much simpler and need not be split (the ``first''
and ``last'' elements in the pipe are identical).  The predicates
``normal'' and ``push-updating'' ensure, in the obvious way, that the
two parts of the split action can only be used as intended. Finally,
``on'', ``first'', ``follow'', and ``last'' encode where the relevant
batches are. The role of ``on'' should be clear, it just encodes
locatedness in areas. As for the pipe contents, they are modelled in a
queue-like fashion, with a head ``first'', a tail ``last'', and a
successor function ``follow''. The two parts of the ``push'' action
update this representation accordingly.

We did not encode uni-directional pipe segments, i.e. for all segments
both ``push'' and ``pop'' actions are available in the IPC-4
encodings. We modeled tankage restrictions with simple constructs
involving tank slots located in areas, each slot having the capacity
to store one unitary batch of some given product type -- that is, the
``push'' and ``pop'' actions now also specify what tank slot the
inserted/outgoing batch comes from/is inserted into. For simple
examples regarding interface and tankage restrictions, re-consider
Figure~\ref{XFIG/push-pop-example.xfig.eps}. If the storage capacity
for $P_2$ in $A_2$ is equal to zero, then the transition from state
(a) to state (b) becomes invalid. If we forbid the interface between
$P_1$ and $P_3$, then the transition from state (b) to state (c)
becomes invalid.

Pipe segment speed can be easily taken account of (in durational
PDDL). If the speed of a segment is $s$, then simply assign the
``push''/``pop'' actions regarding that segment a duration
proportional to $\frac{1}{s}$. (In the IPC-4 encoding, each
``start''/``end'' action takes exactly that time, while the non-split
actions regarding length-1 segments take time $\frac{2}{s}$.)

In reality, as outlined above the goals refer to amounts of product
requested to be at certain destination areas. With our encoding based
on batches, formulating such a goal would mean to introduce a
potentially large disjunction of conjunctive goals. If one wants to
say, e.g., that three unitary batches of product $P$ are requested in
area $A$, then the needed goal condition is the disjunction
$\bigvee_{\{ b_1, b_2, b_3 \} \subseteq B} (at b_1 A) \wedge (at b_2
A) \wedge (at b_3 A)$ of the respective conjunctive goal for all
three-subsets $\{ b_1, b_2, b_3 \}$ of the batches $B$ of type $P$. To
avoid exponential blow-ups of this kind, in our encoding we used
``personalized'' goals instead, referring to specific batches instead
of product amounts. Basically, this comes down to pre-selecting one of
the $\{ b_1, b_2, b_3 \}$ subsets in the above
disjunction.\footnote{Note that a bad choice of $\{b_1, b_2, b_3\}$
  can make the task harder to solve. We are, however, currently
  investigating the computational complexity of different variants of
  the Pipesworld, and our preliminary results suggest that
  allowing/disallowing personalized goals does not affect the
  complexity.} One could also avoid the blow-up by replacing the
disjunction with an existential quantification; but that step would be
undone in the compilation to STRIPS anyway.

Deadlines on the arrival of batches are, in durational PDDL, easily
modeled by their compilation to timed initial literals. For each goal
deadline there is a literal saying that the respective batch can still
be ejected from the end of a pipe segment. The literal is initially
true, and becomes false at the time of the deadline. As described
above, in the application there can also be pre-specified time points
at which an area produces or consumes a given amount of a product. We
did not model this in the IPC-4 domain (see also
Section~\ref{pipesworld:future}).

\scalefig{0.7}{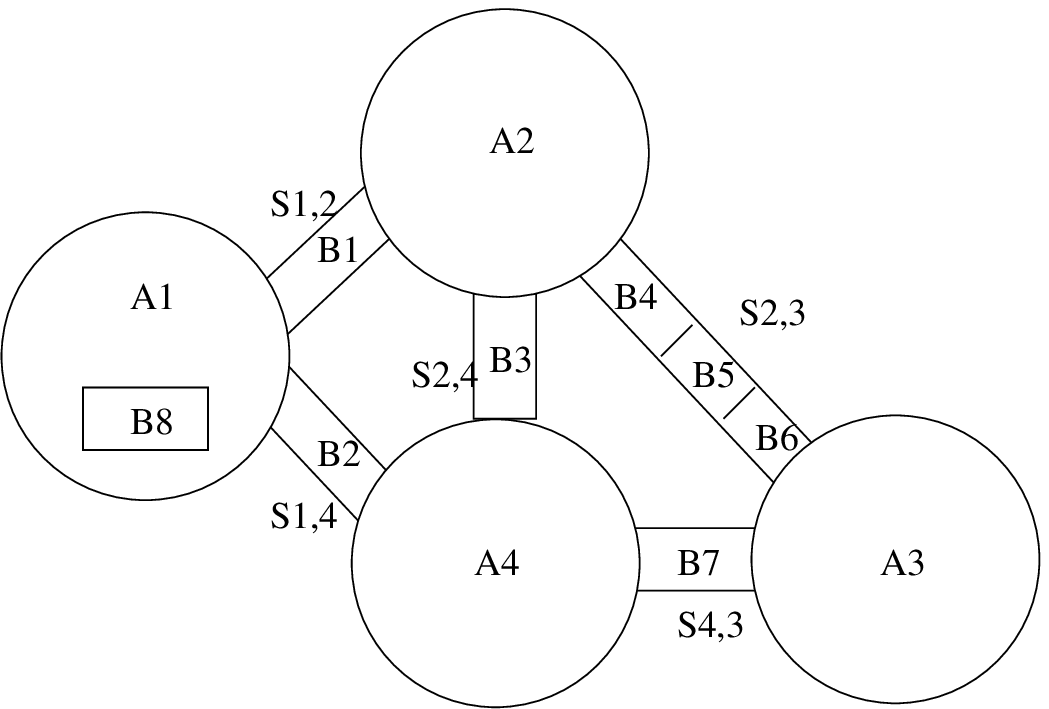}{An example where cycling is
  required to achieve the goal (place B8 in A3). Pipe segment $Si,j$
  is directed from $Ai$ to $Aj$.}

As mentioned above, the structure of the Pipesworld domain can lead to
several subtle phenomena in the possible plans. An example where plans
have to perform a cyclic sequence of pumping operations is depicted in
Figure~\ref{XFIG/cycling.xfig.eps}. The goal is to place B8 in A3. The
shortest plan is the following (for readability, in the action
parameters only the batches going into and out of the pipes are
shown): {\em 0: PUSH S1,4 B8 B2, 1: POP S2,4 B2 B3, 2: POP S1,2 B3 B1,
  3: PUSH S1,4 B1 B8, 4: PUSH S4,3 B8 B7, 5: POP S2,3 B7 B4, 6: PUSH
  S2,4 B4 B2, 7: PUSH S4,3 B2 B8}.  Observe that this plan contains
two cyclic patterns. Action $0$ inserts $B8$ into $S14$. Actions
$1,2,3$ then form a cycle $\{S_{2,4},S_{1,2},S_{1,4}\}$ that brings
$B8$ into A4.  Thereafter, action $4$ inserts $B8$ into $S43$, and
actions $5,6,7$ form another cycle $\{S_{2,3},S_{2,4},S_{4,3}\}$
bringing $B8$ to its goal position $A3$.\footnote{Note that the need
  for such cyclic patterns is {\em not} an oddity introduced by our
  encoding. It is something that may (but is probably not very likely
  to) happen in reality: like in the example, it becomes necessary if
  there isn't enough liquid in an origin area (here, A1 and A4) to
  push the needed amount of liquid (here, B8) through to its
  destination.}

\subsubsection{IPC-4 Domain Structure}
\label{pipesworld:structure}

\begin{sloppypar}
The Pipesworld domain versions used in IPC-4 are {\em
  notankage-nontemporal}, {\em tankage-nontemporal}, {\em
  notankage-temporal}, {\em tankage-temporal}, {\em
  notankage-temporal-deadlines}, and {\em
  notankage-temporal-deadlines-compiled}. All versions include
interface restrictions. The versions with ``tankage'' in their name
include tankage restrictions. In the versions with ``temporal'' in
their name, actions take different amounts of time depending on the
pipeline segment that is being moved, as explained above. The versions
with ``deadlines'' in their name include deadlines on the arrival of
the goal batches. One of these versions models the deadlines using
timed initial literals, in the other version (naturally, with
``compiled'' in its name) these literals are compiled into artificial
(temporal) PDDL constructs. None of the encodings uses any ADL
constructs, and of each version there is just one (STRIPS)
formulation.
\end{sloppypar}

\begin{table}[t]
\centering
\begin{tabular}{|l|l||c|c|}
\hline 
version               & formulation  & $max\mbox{-}\#op$ & $max\mbox{-}\#act$\\\hline\hline 
notankage-nontemporal & STRIPS               & 6 &  (14800) 13696   \\\hline 
notankage-temporal    & STRIPS      & 6 &  (14800) 13696  \\\hline 
notankage-temporal-d  & STRIPS   & 6 &    (8172) 7740   \\\hline 
notankage-temporal-dc & STRIPS      & 9 &    (8175) 7742  \\\hline 
tankage-nontemporal   & STRIPS         & 6 &    (107120) 101192  \\\hline 
tankage-temporal   & STRIPS  & 6 &       (107120) 101192  \\\hline 
\end{tabular}
\caption{\label{pipesworld:versionstable} Overview over the different 
  domain versions of Pipesworld. Abbreviations used: 
  ``temporal-d'' for ``temporal-deadlines'', ``temporal-dc'' for 
  deadlines-compiled; $max\mbox{-}\#op$ is the maximum number of 
  (parameterized) PDDL operators for any instance, $max\mbox{-}\#act$ is the 
  maximum number of ground actions for any instance. Data in 
  parentheses are collected before FF's ``reachability'' pre-process 
  (see text).}
\end{table}

The domain versions and numbers of ground actions are overviewed in
Table~\ref{pipesworld:versionstable}. As before, the data were
measured using (a temporal extension of) FF's pre-processor. The
numbers shown in parentheses refer to the situation {\em before} that
pre-processor's ``reachability'' pre-process, which builds a relaxed
planning graph for the initial state and removes all actions that do
not appear in that graph. We can observe that the numbers of ground
actions are very low in the domain versions with deadlines, and
extremely high in the versions with tankage restrictions. The former
is simply because, due to the complicated generation process
(explained in the next sub-section), examples with deadlines were
generated only up to a smaller size. The latter -- high numbers of
actions in the presence of tankage restriction -- is due to the
additional blow-up incurred by the choice of tank slots from which to
draw/in which to put the batches. We note that the effect of the
``reachability'' pruning is relatively moderate, in particular much
lower than, e.g., in Airport, c.f.\ Section~\ref{airport:structure}.

\subsubsection{IPC-4 Example Instances}
\label{pipesworld:examples}

\begin{figure}[htb]
\begin{tabular}{cc}
\scaleonlyfig{0.74}{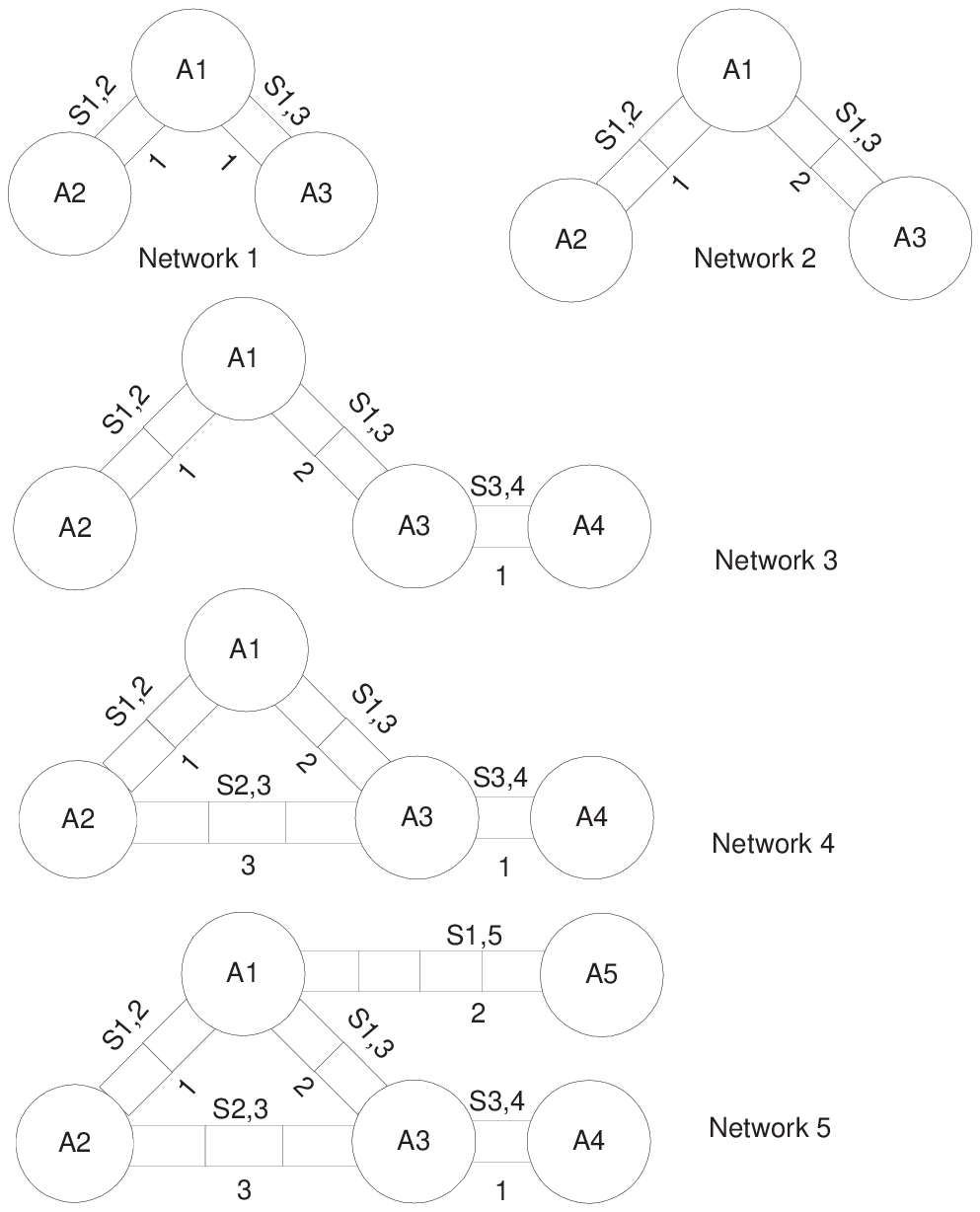} & \hspace{-0.9cm} \scaleonlyfig{0.43}{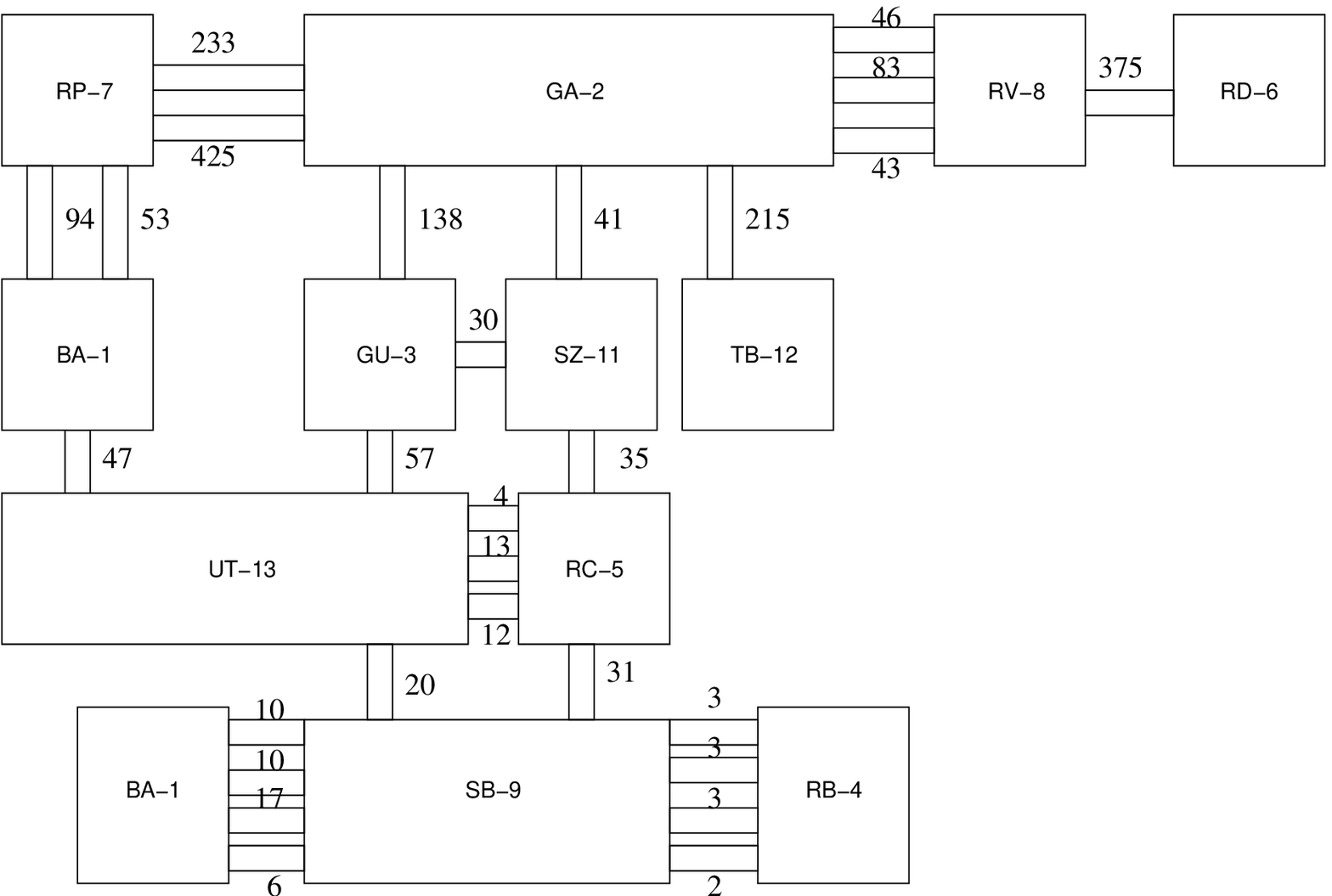}\\
(a) & (b)\\
\vspace{-0.0cm}
\end{tabular}
\caption{\label{fig:topologies} The IPC-4 Pipesworld network
  topologies (a), and a real network topology (b). The segment volumes
  in the latter are annotated in $100 m^{3}$ units.}
\end{figure}

The Pipesworld example instances were generated by Frederico Liporace,
in a process going from random generators to XML files to PDDL
files.\footnote{The same XML file is mapped into different PDDL files
  depending on the kind of encoding used; there was a lot of trial and
  error before we came up with the final IPC-4 encoding.} Five scaling
network topologies were designed and used as the basis for the
instance generation. Figure~\ref{fig:topologies} shows the network
topologies, as well as a real-world network topology for comparison.
As one can see, the largest network topology used in IPC-4 is not
quite yet in the same ballpark as the real network; but neither is it
trivially small in comparison. The volumes for pipeline
segments that connect the same areas in the real-world example
are not necessarily the same because
the segments may have different cross section diameters.

\vspace{0.0cm}

% \scalefig{0.7}{XFIG/topologies.eps}{IPC-4 Pipesworld network
%   topologies.}

For the domain versions without tankage restrictions and deadlines,
for each of the network topologies 10 scaling random instances were
generated. Within a network, the instances scaled in terms of the
total number of batches and the number of batches with a goal
location. For the instances featuring tankage restrictions or
deadlines, the generation process was more complicated because we
wanted to make sure to obtain only solvable instances. For the tankage
restriction examples, we ran Mips \cite{Own:Taming} on the respective
``notankage'' instances, with incrementally growing
tankage.\footnote{Mips was a convenient choice since it is one of our
  own planners, and can also deal with temporal constructs.} We chose
each instance at a random point between the first instance solved by
Mips, and the maximum needed tankage (enough tankage in each area to
accommodate all instance batches). Some instances could not be solved
by Mips even when given several days of runtime, and for these we
inserted the maximum tankage. For the deadline examples, we ran Mips
on the corresponding instances without deadlines, then arranged the
deadline for each goal batch at a random point in the interval between
the arrival time of the batch in Mips's plan, and the end time of
Mips's plan. The instances not solved by Mips were left out.

% The largest network topology used in IPC-4 is not quite yet in the
% same ballpark as the network topologies one has to deal with in the
% real world; but neither is it very far away from these. See
% Figure~\ref{XFIG/campono_network.eps} for an example of a real
% network.
% \scalefig{0.4}{XFIG/campono_network.eps}{Example of a real pipeline
%   network. 

\subsubsection{Current and Future Work}
\label{pipesworld:future}

There is ongoing work on developing a Pipesworld specific solver,
named Plumber
\cite{milidiu:liporace-hms-04,pipesworld_ipipelinec2004}. Plumber
incorporates a pipeline simulator, domain specific heuristics, and
procedures for reducing the branching factor by symmetry elimination.
It also lets the user choose between different search strategies, such
as enforced hill climbing \cite{hoffmann:nebel:jair-01} and learning
real time A*\cite{lrta}. Currently it is being extended to support
temporal planning as well.

The availability of this solver will enable the extension of the
Pipesworld benchmark, since it will be easier to overcome the
aforementioned difficulties in generating large feasible instances. We
hope to be able to generate feasible instances for real-world pipeline
topologies, like the one shown in Figure \ref{fig:topologies}.

In addition to generating larger instances, the Pipesworld benchmark
may be extended in many ways to make it closer to the real application
scenario. The relevant possible extensions include:

\begin{itemize}
\item Defining some pipeline segments with a single flow direction,
  that is, segments where only ``push'' or ``pop'' actions are
  allowed. Note that this introduces dead ends/critical choices into
  the problem.
\item Un-personalized goals. This could be accomplished, e.g., by
  imposing the desired tank volume for the goal products in the
  respective areas. The planner then also has to decide which batches
  will be used to bring the tank volume up to the desired level.
\item Modeling production and consumption of products at pre-specified
  points in time, as described above.
\item Using rational numbers to model tank capacities and current
  volumes, instead of the encoding based on unitary tank slots. Apart
  from being a more precise model of the real world (when combined
  with rational-valued batch sizes, see below), such an encoding would
  avoid unnecessary symmetries that currently arise from the
  availability of several non-distinguishable tank slots (in the same
  area, for the same product).
\end{itemize}

The most important shortcoming of our encoding is the use of unitary
batches. It would be much more appropriate to base the encoding on
product amounts given by real numbers.  One problematic aspect of such
an encoding is that it would, most naturally, demand a continuous
choice of {\em how much} liquid to pump into a pipeline.  Like in
Airport (c.f.\ Section~\ref{airport:future}), such a choice could
naturally be modelled using Fox and Long's \citeyear{fox:long:jair-03}
``duration inequalities'', but it is unclear how to develop planners
that can deal with these reasonably well. Unlike in Airport,
implementing such a choice is not the end of the difficulties on the
modelling side. How to model the continuous contents of a pipeline?
The number of distinct regions of liquid in the pipeline can grow
arbitrarily high, in principle. One solution might be to fix some
upper bound, and simply disallow a pumping operation if it would
result in too many distinct regions. This may be a bearable loss of
precision, given the upper bound is high enough. But even then, it is
bound to be awkward to correctly update the contents of the pipeline
when some amount $x$ of product is pushed in: the number of different
products leaving the pipe depends on $x$. An option here may be to use
a complicated construct of conditional effects.

All in all, our impression is that pipeline scheduling won't be
realistically modelled in PDDL, and successfully solved with planners,
unless one introduces into the language a data structure suitable for
modelling the contents of pipes. Basically, this would be queues whose
elements are annotated with real numbers, and whose basic operations
are the usual ``push'' and ``pop''. The semantics of the pipes could
then be explicitly computed inside the planner, rather than awkwardly
modelled using language constructs that are likely to disturb a
general search mechanism.

%%% Local Variables: 
%%% mode: latex
%%% TeX-master: t
%%% TeX-master: t
%%% TeX-master: t
%%% End: 

\newcommand{\makemath}[1]{\relax\ifmmode #1\relax\else $#1$\fi}
\newcommand{\IR}{\makemath{I\kern -0.25em R\/}}

\subsection{Promela}
\label{promela}

% REVISION: STEFAN: short description up front, shorten, revise
% formulations, and insert ``snippets'' - done

This domain was created for IPC-4 by Stefan Edelkamp.

\subsubsection{Application Domain}
\label{promela:application}

 Before dropping into the Promela domain, we briefly recall its
 origin.

 % Model Checking 

 %The model of the system itself is checked
 %against the property requirement, that is usually specified in form 
 %of some temporal logic.

 % SPIN
 
 The model checker \emph{SPIN}~\cite{Holzmann:NewBook} targets
 efficient software verification.  It has been used to trace logical
 design errors in distributed systems design, such as operating
 systems, data communications protocols, switching systems, concurrent
 algorithms, railway signaling protocols, etc. The tool checks the
 logical consistency of a specification. SPIN reports on deadlocks,
 unspecified receptions and identifies race conditions, and
 unwarranted assumptions about the relative speeds of processes.  SPIN
 (starting with Version 4) provides support for the use of embedded C
 code as part of model specifications. This makes it possible to
 directly verify implementation level software specifications, using
 SPIN as a driver and as a logic engine to verify high level temporal
 properties.  SPIN works on-the-fly, which means that it avoids the
 need to construct a global state graph as a prerequisite for the
 verification of system properties.  SPIN supports property checking
 in linear temporal logic (LTL). LTL expresses state trajectory
 constraints, using temporal modalities like \emph{eventually},
 \emph{always}, and \emph{until}\footnote{Note that some fragments of
   LTL are likely to be included into the PDDL language for the next
   international planning competition~\cite{GereviniLong}}.  SPIN uses
 specific mechanisms for specifying deadlock-freeness and other safety
 properties, in addition to general LTL specifications.  To explore
 the state space an ordinary or a nested search algorithm is applied,
 depending on whether or not a state-based (a.k.a. safety) property is
 to be verified.
  
 \emph{Promela} is SPIN's input specification language. Its
 computational model is that of asynchronous communicating finite
 state machines. Promela allows to define classes of finite processes.
 A special process called \emph{init} is started first and usually
 governs the instantiation of the other processes of the system. As it
 is possible for a process to invoke another one, Promela allows
 modeling systems with dynamic creation of state components.
 Communication in Promela is achieved via shared variables and message
 channels. Two kind of message channels are distinguished for
 synchronous and asynchronous communication. An asynchronous channel is
 basically a FIFO queue, while synchronous channels imply rendezvous
 communication in which a transition of the system involves two
 processes, one reading a message from the channel and another sending
 a message to it. Here, we consider only asynchronous communication.
 The body of each process class is basically a sequence of statements.
 Each statement is interpreted as a transition of the process. 
 %The
 %executability of a statement determines the guard of the
 %corresponding transition. 
 Typical statements include assignments, numerical and boolean
 expressions and channel operations.  Promela also allows to define
 atomic regions, whose are a sequence of transitions that should be
 treated as an atomic action. They can be interpreted as weighted
 transitions whose costs are the number of steps within the
 regions.\footnote{Further documentation for the Promela specification
   language can be found on the web site for SPIN at
   http://netlib.bell-labs.com/netlib/spin/whatispin.html}

%\scalefig{0.001}{PICS/philosophers.eps}{The Dining philosopher example.}

%    \begin{figure}[t]
%    \centerline{
%    \qquad \qquad\includegraphics[angle=90,width=1.5cm]{PICS/philosophers.ps}
%    }
%    \caption{The Dining Philosophers example.}\label{Pic:Phil}
%    \end{figure}

 For IPC-4, we used two example communication protocols formulated in
 Promela: Dijkstra's \emph{Dining Philosophers} problem, and the
 so-called \emph{Optical Telegraph} protocol. We briefly describe the
 latter protocol in Section~\ref{promela:examples}. To illustrate the
 Promela language, let us consider the Dining Philosophers problem,
 where $n$ philosophers sit around a table to have lunch.  There are
 $n$ plates, one for each philosopher, and $n$ forks located to the
 left and to the right of each plate.  Since two forks are required to
 eat the spaghetti on the plates, not all philosopher can eat at a
 time.  Moreover, no communication except taking and releasing the
 forks is allowed. The task is to devise a local strategy for each
 philosopher that lets all philosophers eventually eat. The simplest
 solution to access the left fork followed by the right one, has an
 obvious problem. If all philosophers wait for the second fork to be
 released there is no possible progress; a deadlock has occurred.

\begin{figure}[htb]
{
\begin{tabbing}
\#define MAX\_PHILOSOPHERS N\\
mtype={fork}\\
\#define left forks[\_pid]\\
\#define right forks[(\_pid+1) \% MAX\_PHILOSOPHERS]\\
\\
chan forks[MAX\_PHILOSOPHERS] = [1] of {bit};\\
\
active \= [MAX\_PHILOSOPHERS] proctype philosopher()\\
\{\\
\>        left!fork;\\
\>       do\\
\>        ::left?fork \= -$>$ /* try to get left fork */\\
\> \>                right?fork; /* try to get right fork */\\
\> \>                /* eat... */\\
\> \>                left!fork; right!fork /* release forks */\\
\> \>                /* meditation... */\\
\>        od\\
\}
\end{tabbing}
}
\vspace{-0.3cm}
\caption{\label{figure:philosophers_promela}Promela specification for a model of the Dining
  Philosophers problem.}
\end{figure}

 It is not difficult and probably insightful to derive a \emph{bottom-up} 
 PDDL encoding for the Dining Philosophers domain, using actions like
 \emph{eat}, \emph{wait} and \emph{think}. Our motivation, however,
 was to come up with a top-down encoding, starting from a Promela 
 specification, automatically translating it into PDDL.

 The deadlock model of the Dining Philosophers is specified in
 Promela as shown in Figure~\ref{figure:philosophers_promela}. The first lines
 define some macros and declare the array of $N$ boolean variables
 that represent the availability of the forks. The following lines
 define the behavior of a process of type \texttt{philosopher}.  The
 process iterates indefinitely in an endless loop (\texttt{do}) with
 one unique entry marked by symbol \texttt{::}.  Statements are
 separated by a semicolon. The first transition \texttt{left!fork}
 consists of the send operation of tag \texttt{fork} to channel
 \texttt{left}, which itself is a macro to address \texttt{forks} with
 the current process id \texttt{\_pid}. It represents the availability
 of the left fork of the philosopher.  The access transition
 \texttt{left?fork} can be executed only if reading tag \texttt{fork} from 
 channel \texttt{left} is successful. The next transition 
 \texttt{right?fork} is similar to the first, while the last two ones
 sends tag \texttt{fork} back to the channels \emph{left} and \emph{right}.

\subsubsection{IPC-4 PDDL Adaptation}
\label{promela:adaptation}

Model Checking and Action Planning are closely related, c.f.
Section~\ref{overview}.  While a model checker searches for a
counterexample in the form of a sequence of transitions to falsify a
given specification, a planner searches for a sequence of actions that
satisfies a given goal. In both cases, the basic models (STRIPS
Planning, Kripke structures), refer to implicit graphs, where the
nodes are annotated with atomic propositions.

\begin{sloppypar}
  For automatically generating a PDDL model from the Promela syntax we
  wrote a compiler~\cite{Own:Promela}.  It is restricted to safety
  properties, especially deadlocks, but assertions and global
  invariances are not difficult to obtain. We also concentrated on
  models with a fixed number of processes, since most of the models of
  communication protocols adhere to this restriction.\footnote{The
    dynamic creation of processes with PDDL would require a language
    extension for \emph{dynamic object creation}. This extension was
    dismissed since it would involve heavy changes to existing planner
    technology, and its relevance (beyond Promela) is unclear.}

  The compiler does not parse the Promela code itself, but
  takes as the input the intermediate representation of the problem that
  is generated by the SPIN validation tool\footnote{More precisely,
    the Promela input file was taken, the corresponding c-file was generated,
    the verifier was compiled and the executable was run with option
    \texttt{-d}.}.  Figure~\ref{figure:philosophers_automata} shows the
  textual automata representation for the philosopher process.  In this case,
  the value \texttt{N} has been initialized with 10 philosophers. While
  this file contains almost all necessary information for the translation, the
  number of processes and queues (i.e., message channels) as well as
  the queue capacities had to be read from the original Promela input
  file\footnote{To avoid
    conflicts with pre-compiler directives, we first invoked the c-compiler with 
    command line option \texttt{-E}, which only executes the pre-compiler.}.
\end{sloppypar}

\begin{figure}[htb]
\begin{tabbing}
proctype \= philosopher\\
\>  state 1 -(trans 3)-$>$ state 6  line 11 =$>$ forks[\_pid]!fork\\    
\>  state 6 -(trans 4)-$>$ state 3  line 12 =$>$ forks[\_pid]?fork \\   
\>  state 3 -(trans 5)-$>$ state 4  line 14 =$>$ forks[((\_pid+1)\%10)]?fork \\ 
\>  state 4 -(trans 3)-$>$ state 5  line 16 =$>$ forks[\_pid]!fork   \\
\>  state 5 -(tras 6)-$>$ state 6  line 16 =$>$ forks[((\_pid+1)\%10)]!fork \\
\end{tabbing}
\vspace{-0.8cm}
\caption{\label{figure:philosophers_automata}Automata representation 
  for the model of the 10 Dining Philosophers problem.}
\end{figure}

\begin{sloppypar}
  To derive a suitable PDDL encoding of the domain, each process
  is represented by a finite state automata. Hence, the propositional
  encoding simulates the automaton.
 % $G_P=(S_P,\Sigma,\mbox{\em init}_P,\mbox{\em
 %   curr}_P,\delta_P)$, with $S_P$ being $P$'s set of (local)
 % states, $\Sigma$ being the set of transitions, $\mbox{\em init}_P
 % \in S_P$ (and $\mbox{\em curr}_P \in S_P$) being the initial (and
 % current) state of~$P$, and $\delta_P: S_P \times \Sigma
 % \rightarrow S_P$ being the transition relation on $P$. 
  Some propositional atoms true in the initial state of 
  one process in the running example problem is shown in 
  Figure~\ref{fig:connectcommunication} (a)\footnote{Here we use 
  transition IDs, in the competition  
  a less accessible textual representation of the label was chosen.}. 

\begin{figure}[htb]
\begin{tabular}{cc} \\
\begin{minipage}{8cm}
\begin{tabbing}  
\=   (is-a-process philosopher-0 philosopher)\\
\>   (at-process philosopher-0 state-1)\\
\>   (trans philosopher trans-3 state-1 state-6)\\
\>   (trans philosopher trans-4 state-6 state-3)\\
\>   (trans philosopher trans-5 state-3 state-4)\\
\>   (trans philosopher trans-3 state-4 state-5)\\
\>   (trans philosopher trans-6 state-5 state-6)\\
\end{tabbing}
\end{minipage}
&
\begin{minipage}{6cm}
\begin{tabbing}
\=   (is-a-queue forks-0 queue-1) \\ 
\>   (queue-head forks-0 qs-0)\\
\>   (queue-tail forks-0 qs-0) \\
\>   (queue-next queue-1 qs-0 qs-0)\\
\>   (queue-head-msg forks-0 empty)\\
\>   (queue-size forks-0 zero) \\
\>   (settled forks-0)\\
\end{tabbing} 
\end{minipage} \\
(a) & (b) \\
\\
\multicolumn{2}{c}{
\begin{minipage}{6cm}
\begin{tabbing}
\=   (writes philosopher-0 forks-0 trans-3) (trans-msg trans-3 fork)\\
\>   (reads philosopher-0 forks-0 trans-4)  (trans-msg trans-4 fork)\\
\>   (reads philosopher-0 forks-1 trans-5)  (trans-msg trans-5 fork)\\
\>   (writes philosopher-0 forks-1 trans-6) (trans-msg trans-6 fork)\\
\end{tabbing}
\end{minipage}
}
\\
\multicolumn{2}{c}{
(c)}
\end{tabular}
\caption{
Propositional encoding of one philosopher's process (a),
Propositional encoding of a (single-cell) communication channel (b), 
Connecting communication to local state transitions (c).
}
\label{fig:connectcommunication}
\end{figure}
  
  The encoding of the communication structure represents channels as  
  graphs. The PDDL encoding additionally exploits a cyclic embedding 
  of a queue into an array.  
  More formally, each (FIFO) channel $Q$ is represented by a structure 
  $G_Q=(S_Q,\mbox{\em head}_Q,\mbox{\em
    tail}_Q,\delta_Q,\mbox{\em mess}_Q$,$\mbox{\em cont}_Q)$, with
  $S_Q$ being the set of queue cells, $\mbox{\em head}_Q$,
  $\mbox{\em tail}_Q \in S_Q$ being the head and tail cells of $Q$,
  $\mbox{\em mess}_Q \in \mbox{\em M}^{|S_Q|}$ being the vector of
  messages in $Q$ ($\mbox{\em M}$ is the set of all messages),
  $\mbox{\em cont}_Q \in \IR^{|S_Q|}$ being the vector of variable
  values in $Q$ and $\delta_Q: S_Q \rightarrow S_Q$ being the
  successor relation for $Q$; if $S_Q = s[1],\ldots,s[k]$ then
  $\delta(s[i]) = s[(i+1)\ \mbox{mod}\ k]$.  Explicitly modeling head
  and tail positions in the queue trades space for time, since queue
  updates reduce to constant time.
\end{sloppypar}
A queue is either
empty (or full) if both pointers refer to the same queue state. As a
special case, very simple queues (as in our example) may consist of only 
one queue state, so the successor bucket of queue state~0 is the 
queue state~0 itself. In this case the
grounded propositional encoding includes operators where the add and
the delete lists share an atom. We here make the standard assumption
that deletion is done first.
The propositional atoms for one queue and the adaption of two queues to one process 
 are exemplified in 
 Figure~\ref{fig:connectcommunication} (b) and (c).

Queue content, shared and local variables are modeled by PDDL fluents.  The only
difference of local variables compared to shared ones is the
restricted visibility scope, so that local variables are prefixed with
the process they appear in. The two benchmark protocols we selected for IPC-4
rely on pure message passing, so that no numerical state variables there are
involved. This allowed us to supply a propositional model
for all problems.

\begin{figure}[htb]
\begin{tabbing}
(\=:action activate-trans\\ 
\> :parameters (?p - process ?pt - proctype ?t - transition ?s1 ?s2 - state)\\ 
\> :precondition (and \= (forall (?q - queue) (settled ?q)) (trans ?pt ?t ?s1 ?s2)\\  
\> \> (is-a-process ?p ?pt) (at-process ?p ?s1) (pending ?p))\\
\> :effect (and (activate ?p ?t) (not (pending ?p)))))
\end{tabbing}
\caption{Testing if a transition is enabled and activating it.}
\label{fig:activatetransition}
\end{figure}

\begin{sloppypar}
  Our PDDL domain encoding uses seven operators, named
  \emph{activate-trans}, \emph{queue-read}, \emph{queue-write},
  \emph{advance-queue-head}, \emph{advance-empty-queue-tail},
  \emph{advance-non-empty-queue-tail}, and \emph{process-trans}.  The
  activation of a process is shown in
  Figure~\ref{fig:activatetransition}.  Here we see that a pending
  process is activated, if all queues are settled and there is a
  transition that matches the current process state.
  
  Briefly, the operators encode the protocol semantics as follows.
  Operator \emph{activate-trans} activates a transition in a process
  of a given type from local state $s_1$ to $s_2$.  
  The operator sets the predicate \emph{activate}. This 
  boolean flag is a
  precondition of the \emph{queue-read} and \emph{queue-write}
  actions, which set propositions that initialize the
  reading/writing of a message. For queue $Q$ in an activated
  transition querying message $m$, this corresponds to the Promela
  expression $Q?m$, respectively $Q!m$. After the read/write operation
  has been initialized, the queue update operators must be applied,
  i.e.  \emph{advance-queue-head}, \emph{advance-empty-queue-tail}, or
  \emph{advance-non-empty-queue-tail} as appropriate. As the names
  indicate, these operators respectively update the head and the tail
  positions, as needed to implement the requested read/write
  operation.  The operators also set a \emph{settled} flag, which is a
  precondition of every queue access action. Action
  \emph{process-trans} can then be applied. It executes the transition
  from local state $s_1$ to $s_2$, i.e. sets the new local process
  state and re-sets the flags.
\end{sloppypar}

\begin{figure}[htb]
{\small
\begin{tabbing}
(\=:derived (blocked-trans ?p - process ?t - transition)\\
\> (exists \= (?q - queue)\\ 
\>  \> (exists \= (?m - message)\\ 
\> \> \>  (exists \= (?n - number)\\ 
\> \> \> \> (and \= (activate ?p ?t) (reads ?p ?q ?t) (settled ?q)\\
\> \> \> \> \>        (trans-msg ?t ?m) (queue-size ?q ?n) (is-zero ?n))))))\\
\\
(\>:derived (blocked ?p - process)\\ 
 \> (exists \= (?s - state)\\
 \> \>  (exists \= (?pt - proctype)\\ 
 \> \> \>  (and \= (at-process ?p ?s) (is-a-process ?p ?pt)\\
\> \> \> \> (forall \= (?t - transition)\\
\> \> \> \> \> (or (blocked-trans ?p ?t) (forall (?s2 - state) (not (trans ?pt ?t ?s ?s2)))))))))\\
\end{tabbing}}
\caption{Derivation of a deadlock.}
\label{fig:derivedeadlock}
\end{figure}

If the stored message does not match the query, or the queue
capacity is either too small or too large, then the active local state
transition will block. If all active transitions in a process block,
the process itself will block. If all processes are blocked, we have a
deadlock in the system. Detection of such deadlocks is implemented, in
different domain versions, either as a collection of specifically
engineered actions or, more elegantly, as a set of derived predicates.
In both cases one can infer, along the lines of argumentation outlined
above, that a process/the entire system is blocked. The goal condition
that makes the planners detect the deadlocks in the protocols is
simply a conjunction of atoms requiring that all processes are
blocked. As an example of the derivation rules for derived predicates,
the PDDL description for the derivation of a deadlock based on blocked
read accesses is shown in Figure~\ref{fig:derivedeadlock}.

\subsubsection{IPC-4 Domain Structure}
\label{promela:structure}

For each of the two benchmark protocols in IPC-4, we created three
different domain versions: \emph{derivedpredicates}, which contains
derived predicates to infer deadlocks; \emph{plain}, a purely
propositional specification with specific actions that have to be
applied to establish the deadlock (the later actions are basically the
\citeA{gazen:knoblock:ecp-97} compilation of derived predicates, c.f.
Section~\ref{compilations}); \emph{fluents} an alternative to the
latter with numerical state variables that encodes the size of the
queues and the messages used to access their contents. We also made a
version called \emph{fluents-derivedpredicates}, the obvious
combination, but none of the IPC-4 competitors participated in there,
so we omit it herein. Within each domain version, there is one
formulation that includes the ADL constructs \emph{quantification},
\emph{disjunctive preconditions}, and \emph{negated preconditions}. In
those domain versions without fluents, another formulation is in pure
STRIPS, obtained from the respective ADL encodings using the
\emph{adl2strips} compiler (which can not handle numeric variables).
Unfortunately, some of the larger problem instances lead to STRIPS
files that were too big to be stored on disk (remember that adl2strips
grounds out all operator parameters). These too-large instances were,
of course, left out of the respective test suites.

We kept \emph{fluent}-domains as separated domain \emph{versions},
rather than domain version formulations, in order be able to compare
propositional and numerical exploration efficiencies, and to emphasize
that fluent variables are essential in real-world model checking and
should be treated separately.
 
\begin{table}[t]
\centering
\begin{tabular}{|l|l||c|c|}
\hline
version & formulation & $max\mbox{-}\#op$ &
$max\mbox{-}\#act$\\\hline\hline
optical-telegraph & STRIPS & 3345 & (3345) 3345
\\\hline
optical-telegraph & ADL & 11 & (5070) 3345
\\\hline
optical-telegraph-dp & STRIPS DP & 4014 & (4014) 4014
\\\hline
optical-telegraph-dp & ADL DP & 11 & (6084) 4014
\\\hline
optical-telegraph-fluents & ADL & 11 & (1337) 1169
\\\hline \hline
philosophers & STRIPS & 840 & (840) 840
\\\hline
philosophers & ADL & 11 & (930) 840
\\\hline
philosophers-dp & STRIPS DP & 1372 & (1372) 1372
\\\hline
philosophers-dp & ADL DP & 11 & (1519) 1372
\\\hline
philosophers-fluents & ADL & 11 & (930) 930
\\\hline
\end{tabular}
\caption{\label{promela:versionstable} Overview over the different 
  domain versions of Promela. Abbreviations used: ``dp'' derived predicates; 
  $max\mbox{-}\#op$ is the maximum number of 
  (parameterized) PDDL operators for any instance, $max\mbox{-}\#act$ is the 
  maximum number of ground actions for any instance. Data in 
  parentheses are collected before FF's ``reachability'' pre-process 
  (see text). Derivation rules (ground derivation rules) are counted as operators 
  (ground actions).}
\end{table}

The domain versions and numbers of operators and ground actions are
overviewed in Table~\ref{promela:versionstable}. Consider the rows in
the table from top to bottom. As before, times in parentheses are
values before FF's ``reachability'' pre-process, which builds a
relaxed planning graph for the initial state and removes all actions
that do not appear in that graph. The STRIPS formulation is fully
grounded using the adl2strips program, derived from FF's pre-processor
(c.f.  Section~\ref{compilations}). This is both the reason why the
number of operators is the same as the number of ground actions, and
why FF's pre-process -- identical to the one run by adl2strips -- has
no effect. In the ADL formulation, we see that the reachability
pruning reduces the number of actions by a factor of almost 2, similar
to the Airport domain (c.f. Section~\ref{airport:structure}). The
picture for the next two domain versions, with derived predicates, is
very similar. In fact, since, consistently with the data in
Section~\ref{new}, we count derivation rules as actions, the data are
{\em identical}. The only reason why it is not identical in
Table~\ref{promela:versionstable} is that, using derived predicates
instead of operators, FF's pre-processor scales to larger instances
(presumably, due to some unimportant implementation detail). In the
next domain version, formulated with numeric variables, FF's
pre-processor scales even worse. However, even in instances with the
same number of telegraphs, there are less ground actions than before,
due to the more different encoding. The observations to be made in
Dining Philosophers are exactly the same, only with different numbers.
The only notable difference is that the effect of FF's reachability
pruning is weaker, yielding only a slight decrease in the number of
actions in the versions without fluents, and no decrease at all in the
version with fluents. Apparently, the more complex process structure
of Optical Telegraph leads to more useless action instances.

\subsubsection{IPC-4 Example Instances}
\label{promela:examples}

As said, we have selected two simple communication protocols as benchmarks
for IPC-4: the encoding of the \emph{Dining Philosopher} problem as
described above, and the so-called \emph{Optical Telegraph} protocol
\cite{Holzmann}.

The Optical Telegraph protocol involves $n$ pairs of communicating
processes, each pair featuring an ``up'' and a ``down'' process. Such
a pair can go through a fairly long, heavily interactive, sequence of
operations, implementing the possible data exchange between the two
stations. Before data are exchanged, various initializing steps must be
taken to ensure the processes are working synchronously. Most
importantly, each process writes a token into a ``control channel''
(queue) at the beginning of the sequence, and reads the token out
again at the end. This causes a deadlock situation because there are
only $n$ control channels, each of which is accessed by two processes.
When every pair of up/down processes has occupied just one control
channel, the overall system is blocked.

In both the \emph{Dining Philosopher} and the \emph{Optical Telegraph}
benchmark, the instances scale via a single parameter, the number of
philosophers and the number of control stations, respectively. We
scaled that parameter from $2$ to $49$ for the competition instances.
The Promela models of the benchmarks are distributed together with our
experimental model checking tool HSF-SPIN~\cite{Own:STTT}, that
extends SPIN with heuristic search strategies to improve error
detection.

\subsubsection{Future Work}
\label{promela:future}

In general terms, we see the Promela planning benchmark as another
important step towards exploiting synergies between the research areas
of Planning and Model Checking~\cite{Fausto:PlanningMC}. For example,
complement to recent progress in planning, \emph{explicit directed
  model checking} in the domain of protocol validation~\cite{Own:STTT}
and \emph{symbolic directed model checking} in the domain of hardware
validation~\cite{Own:Hardware} has led to drastic improvements to
state-of-the-art model checkers. This and other work, e.g.,
\cite{Yang:Guided,Bloem:Guided}, show that in model checking there is
a growing interest in guided exploration, mostly to find errors faster
than blind state space enumeration algorithms. With the compilation of
the Promela domain model, an alternative option of applying heuristic
search to model checking problems is available. More work is needed to
understand when planning heuristics work or fail in model checking
benchmarks.
   
We strongly believe that both communities will profit from a wide-spread
availability of techniques that represent Model Checking problems in
PDDL. This allows a direct comparison of exploration 
efficiencies. Based on the design of the Promela domain, suitable PDDL 
domain encodings of two further expressive model checking input languages,
Graph Transformation Systems~\cite{Own:GTS} and 
Petri Nets~\cite{Own:PetriNets}, have been proposed. The encodings exploit 
the expressive power of PDDL as well as the efficiency of current planners.
As a result, state-of-the-art planners are often faster compared to
model checkers in these benchmarks.

%%% Local Variables: 
%%% mode: latex
%%% TeX-master: t
%%% TeX-master: t
%%% TeX-master: t
%%% TeX-master: t
%%% End: 

\subsection{PSR}
\label{psr}

% REVISION: JOERG: revise formulations and insert ``snippets''

Sylvie Thi\'ebaux and others have worked on this application domain.
The domain was adapted for IPC-4 by Sylvie Thi\'ebaux and J\"org
Hoffmann.

\begin{figure}[htb]
\begin{center}
\scaleonlyfig{0.50}{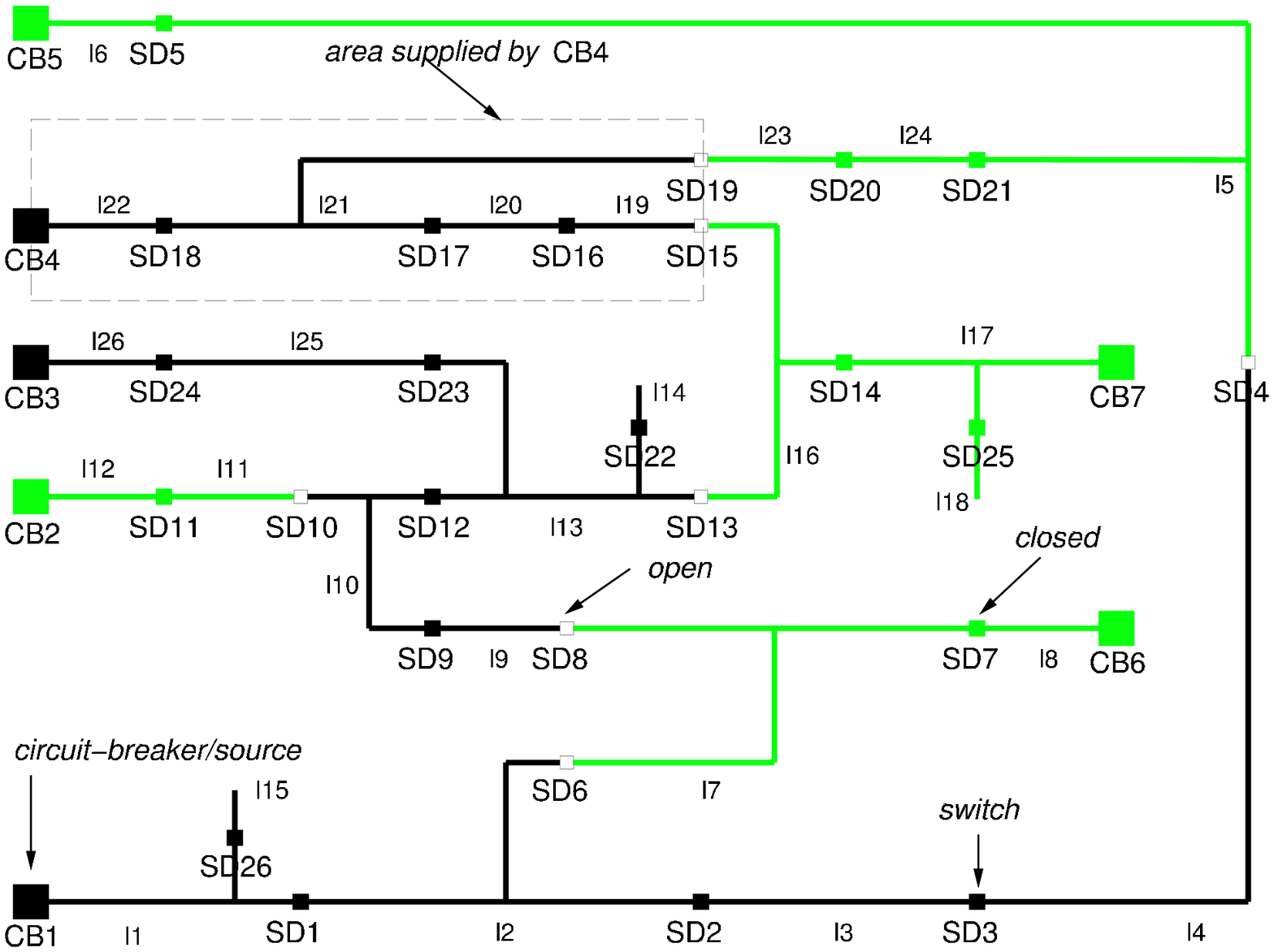}\\
\end{center}
\caption{\label{fig:psr} Sample power distribution system. 
Sources/circuit-breakers (e.g., CB4) are
represented by large squares, and switches (e.g., SD3) by small
squares. Open switches (e.g., SD8) are white. The area fed by CB4 is
boxed. Gray and dark are used to distinguish adjacent areas fed by
different sources}
\end{figure}

\subsubsection{Application Domain}
\label{psr:application}

The Power Supply Restoration (PSR) domain we consider here is derived
from an application investigated by Sylvie Thi\'ebaux and others
\cite{thiebaux:etal:uai-96,thiebaux:cordier:ecp-01}. PSR
deals with reconfiguring a faulty power distribution 
system to resupply customers affected by the faults. This is a topic
of ongoing interest in the field of power distribution.

In more detail, a power distribution system (see Figure~\ref{fig:psr}),
is viewed as a network of electric lines connected by switches and fed
via a number of power sources that are equipped with
circuit-breakers. Switches and circuit-breakers have two possible
positions, open or closed, and are connected to at most two lines.
There is no restriction on the connectivity of lines, some
extremities of which can also be connected to earth.
When the circuit-breaker of a power source
is closed, the power flows from the source to the lines downstream,
until the flow is stopped by an open switch.  The switches are used to
appropriately configure the network and their position is initially
set so that each line is fed by exactly one source.

Due to bad weather conditions, permanent faults can affect one or
more lines of the network. When a power source feeds a faulty line, the
circuit-breaker fitted to this source opens to protect the rest of the
network from overloads. This leaves {\em all} the lines fed by the
source without power. The problem consists in planning a sequence of
switching operations (opening or closing switches and
circuit-breakers) bringing the network into a configuration where
a maximum of non-faulty lines are resupplied. For instance, suppose that line
l20 becomes faulty. This leads the circuit-breaker CB4
to open and the boxed area to be without power. A possible
restoration plan would be the following: open switches SD16 and
SD17 to isolate the faulty line, then close SD15 to have source
CB7 resupply l19, and finally re-close CB4 to resupply the others.

In the original PSR problem \cite{thiebaux:cordier:ecp-01}, the
maximal capacity of sources and lines, as well as the load requested
by customers are taken into account. The plan must optimize various
numerical parameters such as breakdown costs, power margins, and
distance to the initial configuration, subject to the capacity
constraints.  Furthermore, due to the fault sensors and switches being
unreliable, the location of the faults and the current network
configuration are only partially observable. When optimizing, this
leads to a complex tradeoff between acting to resupply lines and
acting (intrusively) to reduce uncertainty.

\subsubsection{IPC-4 PDDL Adaptation}
\label{psr:adaptation}

In the PDDL adaptation, we benefited from contributions by Piergiorgio
Bertoli, Blai Bonet, Alessandro Cimatti, and John Slaney
\cite{bertoli:etal:ecai-02,bonet:thiebaux:icaps-03}.  Compared to the
original PSR domain described above, the IPC-4 version underwent 3
major adaptations. Firstly, the IPC deals with fully observable
domains. Hence, while partial observability in PSR is a crucial issue
\cite{thiebaux:etal:uai-96,bertoli:etal:ecai-02,bonet:thiebaux:icaps-03},
the IPC version assumes complete observability. Secondly, given the
difficulty of encoding even the basic problem, we chose to ignore the
numerical and optimization aspects of PSR (capacities, power margins,
\ldots). Thirdly, the IPC-4 version is set up as a pure
goal-achievement problem, where the goal specifies a set of lines that
must be (re)-supplied. We considered a more realistic goal asking the
planner to supply any line that can be. However, we were unable to
compile this goal into STRIPS in reasonable space, and opted for the
simpler goal to keep the STRIPS formulation as consistent as possible
with others.

Our highest level and most natural IPC-4 encoding of PSR involves ADL constructs and
derived predicates.  Briefly, the encoding works as follows. PSR
problem instances specify (1) the network topology, i.e., the objects
in the network and their connections (the lines, the switching devices, that is, the
switches and the sources/circuit-breakers, two ``side'' constants
side1 and side2 to denote the two connection points of a switching
device, and the connection relations between those objects), 
(2) the initial configuration, i.e.,
the initial positions (open/closed) of the switching devices, and (3)
the modes (faulty or not) of the various lines.  Among those, only the
devices' positions can change.  A number of other predicates are
derived from these basic ones. They model the propagation of the
current into the network with a view to determining which lines are
currently fed and which sources are {\em affected} by a fault,
i.e. feed a fault.  The closed-world assumption semantics of PDDL2.2
derived predicates is exactly what is needed to elegantly encode such
relations. These require a recursive traversal of the network paths
which is naturally represented as the transitive closure of the
connection relation of the network. The most complex of these derived
predicates, $upstream$, requires four parameters, two of which,
however can only take two possible values, and expresses that the
power flows from one of the two sides of some device (side ?sx of
device ?x) to one of the sides of another (side ?sy of device ?y)
This happens when the side of ?x which is opposite to ?sx is directly connected to ?sy
(via some line), or if there exists some closed device ?z one side of which 
is upstream of ?sx and the other side of which is connected to ?sy:
\begin{tabbing}
(:derived \= (upstream ?x - DEVICE ?sx - SIDE ?y - DEVICE ?sy - SIDE)\\
          \>(and \= (closed ?x)\\
          \>     \> (or \= (and (= ?sx side1) (con ?x side2 ?y ?sy))\\
          \>     \>     \> (and (= ?sx side2) (con ?x side1 ?y ?sy))\\
          \>     \>     \> (exists \= (?z - DEVICE)\\
	  \>     \>     \>         \> (and \= (closed ?z)\\
          \>     \>     \>         \>      \> (or \= (and (con ?z side1 ?y ?sy) (upstream ?x ?sx ?z side2))\\
          \>     \>     \>         \>      \>     \> (and (con ?z side2 ?y ?sy) (upstream ?x ?sx ?z side1))))))))
\end{tabbing}
From upstream, it is relatively easy to define predicates stating whether 
a given line is fed or a given source is affected.

The goal in a problem instance asks that given lines be fed and all
sources be unaffected.\footnote{Note that after the circuit-breaker of
  an affected source opens, this source is not affected any more, as
  it does not feed any line. 
Then, if the circuit-breaker is closed again, the source will stay
unaffected unless it re-starts feeding a faulty line.}  
The available actions are closing and
opening a switching device. Their effect is simply to set
the device position as requested. In addition, there is an action {\em
  wait}, which models the event of circuit-breakers opening when they
become affected. Wait is applicable when an affected source exists,
and is the only applicable action in that case (the open and close
actions require as a precondition that no source is affected). This, 
together with the goal, ensures that the wait action is applied as soon
as a source is affected. The effect of the wait action is to open all
the affected circuit-breakers. Concretely, the wait and close 
actions are as follows (note that open is similar to close and that earth is
treated as a device whose position cannot be changed by the actions):
\begin{tabbing}
(:ac\=tion close\\
    \>:parameters  (?x - DEVICE)\\
    \>:precondition (and \= (not (= ?x earth))\\
    \>                   \> (not (closed ?x))\\
    \>                   \> (forall (?b - DEVICE) (not (affected ?b))))\\
    \>:effect (closed ?x))\\
\\
(:action wait\\
    \>:parameters  ()\\
    \>:precondition (exists (?b - DEVICE) (affected ?b))\\
    \>:effect (forall (?b - DEVICE) (when (affected ?b) (not (closed ?b)))))\\
\end{tabbing}
It would have been possible to encode the opening of affected breakers
as a conditional effect of the close action.  However, this would have
required more complex derived predicates with an additional device as
parameter and a conditional flavor, specifying, e.g., whether or not a
circuit-breaker {\em would be}\ affected {\em if}\ we were to close
that device.

\subsubsection{IPC-4 Domain Structure}
\label{psr:structure}

We used four domain versions of PSR in IPC-4. Primarily, these
versions differ by the size of the problem instances encoded. The
instance size determined in what languages we were able to formulate
the domain version. We tried to generate instances of size appropriate
to evaluate current planners, i.e, we scaled the instances from
``push-over for everybody'' to ``impossibly hard for current automated
planners'', where we got our intuitions by running a version of FF
enhanced to deal with derived predicates. The largest instances are of
the kind of size one typically encounters in the real world. More on
the instance generation process is said in Section~\ref{psr:examples}.

The domain versions are named 1. {\em large}, 2. {\em middle}, 3. {\em
  middle-compiled}, and 4. {\em small}. Version 1 has the single
formulation {\em adl-derivedpredicates}. Version 2 has the
formulations {\em adl-derivedpredicates}, {\em
  simpleadl-derivedpredicates}, and {\em strips-derivedpredicates}.
Version 3 has the single formulation {\em adl}, and version 4 has the
single formulation {\em strips}. The formulation names simply give the
language used. Version 1 contains the largest instances, versions 2
and 3 contain (the same) medium instances, and version 4 contains the
smallest instances. The {\em adl-derivedpredicates} formulation is
inspired by \citeA{bonet:thiebaux:icaps-03}; it makes use of derived
predicates as explained above, and of ADL constructs in the derived
predicate, action, and goal definitions. In the {\em
  simpleadl-derivedpredicates} and {\em strips-derivedpredicates}
formulations, all ADL constructs (except conditional effects in the
{\em simpleadl}\ case) are compiled away.  The resulting fully
grounded encodings are significantly larger than the original, while
on the other hand the length of plans remains nearly
unaffected\footnote{The only variation is due to the fact that the
  existential precondition of the wait action causes the compilation
  to split this action into as many wait actions as circuit-breakers}.
The pure {\em adl} formulation is obtained from the {\em
  adl-derivedpredicates} formulation by compiling derived predicates
away, using the method described by
\citeA{thiebaux:etal:ijcai-03,thiebaux:etal:ai-05}.  While there is no
significant increase in the domain size, the compilation method can
lead to an increase in plan length that is exponential in the arity of
the derived predicates (no compilation method can avoid such a blow-up
in the worst case, see
\citeR{thiebaux:etal:ijcai-03,thiebaux:etal:ai-05}). Indeed, in our
particular PSR example instances, we observed a considerable blow up
in plan length. We felt that this blow up was too much to allow for a
useful direct comparison of data generated for {\em
  adl-derivedpredicates} as opposed to {\em adl}, and we separated the
{\em adl} formulation out into domain version 3 as listed above.

The {\em strips}\ domain formulation proved quite a challenge.  All
the 20 or so schemes we considered for compiling both derived
predicates and ADL constructs away led to either completely
unmanageable domain descriptions or completely unmanageable plans. The
problem is that feasible compilations of derived predicates create new
actions with highly conditional effects, and that compiling those away
is impractical. We therefore adopted a different fully-grounded
encoding inspired by \citeA{bertoli:etal:ecai-02}. The encoding is
generated from a description of the problem instance by a tool
performing the reasoning about power propagation. In the resulting
tasks, the effects of the close actions directly specify which
circuit-breakers open as a result of closing a switch in a given
network configuration. No derived predicates are needed, and
consequently the STRIPS encoding is much simpler and only refers to
the positions of the devices and not to the lines, faults, or
connections. Nevertheless, we were still only able to formulate
comparatively small instances in STRIPS, without a prohibitive blow-up
in the encoding size.

\begin{table}[t]
\centering
\begin{tabular}{|l|l||c|c|}
\hline
version & formulation & $max\mbox{-}\#op$ &
$max\mbox{-}\#act$\\\hline\hline
large & ADL DP & 7 & (14038) 7498
\\ \hline
middle & ADL DP & 7 & (7055) 3302
\\ \hline
middle & SIMPLE-ADL DP & 3485 & (3485) 3485
\\ \hline
middle & STRIPS DP & 3560 & (3560) 3560
\\ \hline
middle-compiled & ADL & 5 & (99) 71
\\\hline
small & STRIPS & 9400 & (9400) 9400
\\\hline
\end{tabular}
\caption{\label{psr:versionstable}Overview over the different domain 
  versions and formulations
  of PSR. Abbreviations used: ``dp'' derived predicates; $max\mbox{-}\#op$ 
  is the maximum number of 
  (parameterized) PDDL operators for any instance, $max\mbox{-}\#act$ is the 
  maximum number of ground actions for any instance.  Data in 
  parentheses are collected before FF's ``reachability'' pre-process 
  (see text). Derivation rules (ground derivation rules) are counted as operators 
  (ground actions).}
\end{table}

The domain versions, formulations, and their respective numbers of
operators and ground actions, are shown in
Figure~\ref{psr:versionstable}. Data in parentheses are collected
before FF's ``reachability'' pre-process, building a relaxed planning
graph for the initial state and removing all actions that do not
appear in that graph. In the encodings using ADL and derived
predicates, this reduces the number of ground actions by a factor of
around $2$; for only ADL, the factor is much smaller; for the other
encodings, no reduction at all is obtained, simply due to the fact
that these encodings are obtained with adl2strips, which uses the same
pruning process. Some interesting observations can be made in the
``middle'' versions and formulations. The data shown there correspond
to the largest instance that FF's pre-processor could handle in {\em
  all} versions/formulations, to enable direct comparison. We see
that, for formulation in SIMPLE-ADL and STRIPS, we need to introduce
some more ground actions. We also see that, curiously, in the
compilation of derived predicates (compilation to
``middle-compiled''), the number of ground actions decreases
dramatically. The reason for this lies in that these data count ground
derivation rules as ground actions, and in the subtleties of the
compilation of derived predicates. In the ``middle'' formulations,
almost all ground actions are in fact ground derivation rules. These
are compiled away for ``middle-compiled'' following
\citeA{thiebaux:etal:ijcai-03,thiebaux:etal:ai-05}, introducing a {\em
  single} action that has one distinct conditional effect for each
derivation rule, c.f. Section~\ref{compilations}. Which just means
that the complexity of thousands of derivation rules is replaced with
the complexity of an action with thousands of conditional effects.

\subsubsection{IPC-4 Example Instances}
\label{psr:examples}

Due to contractual agreements, we were unable to use real data in the
competition. Instead, PSR instances were randomly generated using
``randomnet'', a special purpose tool implemented by John Slaney.

Power distribution networks often have a mesh-able structure
exploited radially: the path taken by the power of each source forms a
tree whose nodes are switches and whose arcs are electric lines;
terminal switches connect the various trees together.  Randomnet takes
as input the number of sources, a percentage of faulty lines, and a
range of parameters for controlling tree depth, branching, and tree
adjacency, whose default values are representative of real
networks. Randomnet randomly generates a network topology and a set of
faulty lines. These are turned into the various PDDL encodings above
by a tool called net2pddl, implemented by Piergiorgio Bertoli and
Sylvie Thi\'ebaux. net2pddl computes the set of all lines that can
be supplied, and makes this the goal.

The instances we generated make use of randomnet default settings,
with two exceptions to create problems of increasing difficulty. The
first is that the maximal depth of the trees takes a range of values
up to twice the default. The larger this value, the harder the
problem.  The second is that the percentage of faulty lines ranges
from 0.1 to 0.7.  Problems at the middle of the range are harder on
average, those at the bottom of the range are more realistic.

Each instance suite contains 50 instances. The small instances feature
between 1 to 6 sources, the middle instances feature up to 10 sources,
and the large instances feature up to 100 sources. The large instances
are of a size typical for real-world instances, or even larger. The
example in Figure~\ref{fig:psr} is representative of a difficult
instance in the middle set.

\subsubsection{Future Work}
\label{psr:future}

While PSR has been around for some time as a benchmark for planning
under uncertainty, we expect that the work done in the framework of
IPC-4 will facilitate its acceptance as one of the standard benchmarks
for planning.  To this end, we have developed a PSR resource web page
giving access to the relevant papers, data, and tools (net2pddl,
randomnet, \ldots).\footnote{The page is available at
  http://rsise.anu.edu.au/$\sim$thiebaux/benchmarks/pds} One aspect of
future work is to complete and maintain this website, making available
a number of already existing tools, such as SyDRe
\cite{thiebaux:etal:uai-96}, a domain-specific system for the full PSR
problem, and Matt Gray's net2jpeg which graphically displays networks
generated by randomnet.

Considering future IPCs, there is potential for extending the PDDL
encoding to take the numerical and optimization aspects of the
benchmark into account. PDDL-like encodings of the partially
observable version of the benchmark exist
\cite{bonet:thiebaux:icaps-03}\ and are ready to be used in a future
edition of the probabilistic part of the IPC.\footnote{The
  probabilistic part of IPC-4 did not feature partially observable
  domains.}

%%% Local Variables: 
%%% mode: latex
%%% TeX-master: t
%%% TeX-master: t
%%% End: 

\subsection{Satellite}
\label{satellite}

% REVISION: JOERG: revise formulations, insert ``snippets''?

The {\em Satellite} domain was introduced in IPC-3 by
\citeA{long:fox:jair-03}. It is motivated by a NASA space application:
a number of satellites have to take images of a number of spatial
phenomena, obeying constraints such as data storage space and fuel
usage. In IPC-3, there were 5 versions of the domain, corresponding to
different levels of the language PDDL2.1: {\em Strips}, {\em Numeric},
{\em SimpleTime} (action durations are constants), {\em Time} (action
durations are expressions in static variables), and {\em Complex}
(durations {\em and} numerics, i.e. the ``union'' of Numeric and
Time).

The adaptation of the Satellite domain for IPC-4 was done by J\"org
Hoffmann. All IPC-3 domain versions and example instances were
re-used, except SimpleTime -- like in the other IPC-4 domains, we
didn't want to introduce an extra version distinction just for the
difference between constant durations and static durations. On top of
the IPC-3 versions, 4 new domain versions were added. The idea was to
make the domain more realistic by additionally introducing time
windows for the sending of the image data to earth, i.e. to antennas
that are visible for satellites only during certain periods of time --
according to Derek Long, the lack of such time windows was the main
shortcoming of the IPC-3 domain.\footnote{We have learned in the
  meantime that the lack of time windows for {\em the gathering of
    data} is also, or even more, essential: often, due to occlusion by
  other objects or due to the rotation of the earth, targets are
  visible only during very restricted periods of time. This probably
  constitutes one of the most important future directions for this
  domain.}

We extended the IPC-3 Time domain version to two IPC-4 domain
versions, {\em Time-timewindows} and {\em Time-timewindows-compiled}.
We extended the IPC-3 Complex domain version to the two \mbox{IPC-4}
domain versions {\em Complex-timewindows} and {\em
  Complex-timewindows-compiled}. In all cases, we introduced a new
action for the sending of data to an antenna. An antenna can receive
data of only a single satellite at a time, an antenna is visible for
only subsets of the satellites for certain time periods, and the
sending of an image takes time proportional to the size of the image.
The time windows were modelled using timed initial literals, and in
the ``-compiled'' domain versions, these literals were compiled into
artificial PDDL constructs. None of the domain versions uses ADL
constructs, so of all versions there is only a single (STRIPS)
formulation.

The instances were generated as follows. Our objectives were to
clearly demonstrate the effect of additional time windows, and to
produce solvable instances only. To accomplish the former, we re-used
the IPC-3 instances, so that the only difference between, e.g., Time
and Time-timewindows, lies in the additional time window constructs.
To ensure solvability, we implemented a tool that read the plans
produced by one of the IPC-3 participants, namely TLPlan, and then
arranged the time windows so that the input plan was suitable to solve
the enriched instance. It is important to note here that the time
windows were {\em not} arranged to exactly meet the times extracted
from the IPC-3 plan. Rather, we introduced one time window per each 5
``take-image'' actions, made the antenna visible during that time
window for only the respective 5 satellites, and let the image size
for each individual image be a random value within a certain range
where the time window was 5 times as long as the sending time
resulting from the maximum possible size.

Of course, the above generation process is arranged rather
arbitrarily, and the resulting instances might be a long way away from
the typical characteristics of the Satellite problem as it occurs in
the real world. While this isn't nice, it is the best we could do
without inside knowledge of the application domain, and it has the
advantage that the enriched instances are solvable, and directly
comparable to the IPC-3 ones.

In the new domain versions derived from Complex, we also introduced
utilities for the time window inside which an image is sent to earth.
For each image, the utility is either the same for all windows, or it
decreases monotonically with the start time of the window, or it is
random within a certain interval. Each image was put randomly into one
of these classes, and the optimization requirement is to minimize a
linear combination of makespan, fuel usage, and summed up negated
image utility.

%REVISION: STEFAN: insert table and text 

%%% Local Variables: 
%%% mode: latex
%%% TeX-master: t
%%% TeX-master: t
%%% TeX-master: t
%%% End: 

\subsection{Settlers}
\label{settlers}

% REVISION: JOERG: revise formulations, insert ``snippets''?

The {\em Settlers} domain was introduced in IPC-3 by
\citeA{long:fox:jair-03}. It makes extensive use of numeric variables.
These variables carry most of the domain semantics, which is about
building up an infrastructure in an unsettled area, involving the
building of housing, railway tracks, sawmills, etc. The domain was
included into IPC-4 in order to pose a challenge for the numeric
planners -- the other domains mostly do not make much use of numeric
variables, other than computing the (static) durations of
actions.\footnote{Note that, to some extent, this is just because the
  numeric values were abstracted away in the PDDL encoding, mostly (in
  Airport and Pipesworld, c.f.\ Sections~\ref{airport:future}
  and~\ref{pipesworld:future}) in order to obtain a discrete encoding
  suitable for PDDL2.2-style actions.} We used the exact same domain
file and example instances as in IPC-3, except that we removed some
universally quantified preconditions to improve accessibility for
planners. The quantifiers ranged over domain constants only so they
could easily be replaced by conjunctions of atoms.

%%% Local Variables: 
%%% mode: latex
%%% TeX-master: t
%%% TeX-master: t
%%% TeX-master: t
%%% End: 

\subsection{UMTS}
\label{umts}

% REVISION: STEFAN: short description up front, shorten, revise
% formulations, and insert ``snippets''

Roman Englert has been working in this application area for several
years.  The domain was adapted for IPC-4 by Stefan Edelkamp and Roman
Englert.

\subsubsection{Application Domain}
\label{umts:application}

Probably the best known feature of UMTS (Universal Mobile
Telecommunication Standard) is higher bit rate \cite{Holma:UMTS}:
packet-switched connections can reach up to 2 mega bit per second
(Mbps)\label{mbps} in the optimal case. Compared to existing mobile
networks, UMTS provides a new and important feature, namely the
negotiation of {\itshape Quality of Service} (QoS)\label{qos} and of
transfer properties. The attributes that define the characteristics of
the transfer are throughput, transfer delay, and data error rate.
UMTS bearers have to be generic in order to provide good support for
existing applications and the evolution of new applications.
Applications and services are divided into four traffic classes by
their QoS \cite{TS23107,Holma:UMTS}. The traffic classes, their
fundamental characteristics, and examples for applications are
summarized in Table~\ref{tab-class}.
\begin{table}
\begin{center}
\begin{tabular}{|c||l|l|l|l|}
\hline Class & Conversational & Streaming & Interactive & Background
\\ \hline & Preserve time & Preserve time & Request res- & Undefined
\\ & relation between & relation between & ponse pattern. & delay. \\
Constraints & information flow & information & Preserve data & Preserve
\\  & on the stream. & entities of the & integrity & data
\\  & Conversational & stream & & integrity \\ & pattern (low delay) & & & \\ \hline & Voice, video & Streaming & Web
browsing, & Background \\ Examples & telephony \& & multimedia & network
games & download \\  & video games & & & of e-mails \\ \hline
\end{tabular}
\end{center}
\caption{UMTS quality of service classes and their characteristics.}
\label{tab-class}
\end{table}

The main distinguishing factor between these classes is how
delay-sensitive the traffic is: the conversational class is very delay
sensitive (approximately 40 ms time preservation), and the background
class has no defined maximum delay.

The UMTS call set-up can be modularized using the perspective of
\emph{Intelligent Software Agents}
\cite{Appleby/Steward/99,Busuioc/99}, since agents are logical units
and enable a discrete perspective of the continuous signaling process.
The call set-up is partitioned into the following modules that are
executed in sequential order \cite{Roman:Journal}:

\begin{description}
\item[TRM] The initial step is the initiation of an application on the
  mobile and the determination of the required resources for the
  execution. The resources of the mobile like display and memory are
  checked by the {\itshape Terminal Resource Management}
  (TRM)\label{trm} and allocated, if possible. Otherwise, the
  execution is aborted.
\item[CT] The wireless connection to the radio network is initiated
  via the dedicated control channel of GSM \cite{Holma:UMTS}. In
  case of success, the transmission of "Ready for service" is
  transferred via the node B to the mobile in order to ensure the
  {\itshape Connection Timing} (CT)\label{ct} for bearer service
  availability.
\item[AM] The information of the mobile like location and data
  handling capabilities is sent to the application server in the
  Internet (cf.~AEEI). The transmission can be done comfortably by a
  so-called service agent \cite{Farjami/etal/00} that is controlled by
  the {\itshape Agent Management} (AM)\label{am} in the CND. The
  advantage of a service agent is, that in case of failure, e.g.,
  network resources are not sufficiently available, the agent can
  negotiate with the terminal's agent about another QoS class or
  different quality parameters.
\item[AEEM] A service agent with the required QoS class for the
  execution of the application and with parameters of the mobile
  application is sent from the mobile's {\itshape Agent Execution
    Environment Mobile} (AEEM)\label{aeem} to the application server
  in the Internet (cf.~AEEI).
\item[RRC] The {\itshape Radio Resource Controller} (RRC)\label{rrc}
  provisions/allocates the required QoS by logical resources from 
  the MAC level in the radio bearer \cite{Holma:UMTS}.
\item[RAB] Then, the bearer resources are supplied on the physical
  level from the {\itshape Radio Access Bearer} (RAB) from the CND and
  the call flow is set-up by mapping the logical QoS parameters and
  the physical QoS resources together.
\item[AEEI] The {\itshape Agent Execution Environment Internet}
  (AEEI)\label{aeei} establishes the data transfer from the core
  network to a PDN (e.g., Internet) and sends a service agent
  (controlled by AM) to the application in the PDN in order to ensure
  the QoS for the application.
\item[BS] Finally, the {\itshape Bearer Service} (BS)\label{bs} for
  the execution of the mobile application is established with the
  required radio bearer resources with QoS. Messages are sent to the
  modules TRM and AEEI to start the execution of the application.
\end{description}

%% \begin{figure}[t]
%%   \includegraphics[width=11cm,angle=270]{PICS/Folie4.eps}
%%   \caption{UMTS access and core network structure adorned with the
%%     modules for the discrete call set-up.}
%%   \label{fig-umts-discrete}
%% \end{figure}

These modules are executed in sequential order to set-up a call for
the execution of mobile applications. Two modules (AEEM and AEEI) have
to be executed in time windows in order to ensure that the agents are
life in the network. However, two constraints have been added: First,
the intra-application constraint, where modules from one application
are ordered. Second, the inter-application constraint, where modules
with same names from different applications cannot be executed in
parallel in order to ensure that the required resources are available.

\subsubsection{IPC-4 PDDL Adaptation}
\label{umts:adaptation}

% renewable resources

Besides action duration, the domain encodes scheduling types of
resources\footnote{The terminology for \emph{resources} in 
planning and scheduling varies. In job-shop scheduling, a machine 
is resource, while in planning such a machine would be a domain 
object. In PDDL, \emph{renewable} and \emph{consumable} resources
are both modeled using numerical fluents and are not per se distinguished.}, 
consuming some amount at action initialization time and
releasing the same amount at action ending time. Scheduling types of
resources have not been used in planning benchmarks before, and the
good news is that temporal PDDL2.1 (Level 3) is capable
of expressing them.  In fact we used a similar encoding to the one
that we found for \emph{Job-} and \emph{Flow-Shop} problems.  As one
feature, actions are defined to temporarily \emph{produce} rather than to
temporarily \emph{consume} resources.  As current PDDL has no way of stating
such resource constraints explicitly, planners that want to exploit
that knowledge have to look for a certain patterns of
\emph{increase}/\emph{decrease} effects to recognize them.
Additionally, the resource modeling of our UMTS adaptation is
constrained to the most important parameters (in total 15). In real
networks several hundred parameters are applied.

In UMTS, two subsequent actions can both check and update the value of
some resources (e.g., \emph{has-mobile-cpu}) at their starting (resp.
ending) time points as far as the start (resp. ending) events are
separated by $\epsilon$ time steps, where $\epsilon$ is minimum slack
time required between two dependent events.  When modeling renewable
resources with an \emph{over all} construct the invariant condition of
the action has to check, what the \emph{at start} event did change. We
decided that this is not the best choice for a proper temporal action.
Consequently, the temporal actions require resources to be available
\emph{before} adding the amount used.

Finally, the time windows for the two agent-based modules are defined
using the average execution times of the modules. The average times
are estimated based on signaling durations of the UMTS network
\cite{Holma:UMTS}.

\begin{table}[t]
\begin{center}
\begin{tabular}{|ll|}
\hline
mobile-cpu\hspace*{1cm} & used with $x$ per cent per application \\
d-available                     & partition of the display, e.g., ticker and chess \\
e-balance                       & energy balance of mobile accumulator \\
mobile-channels         & used for data transfer \\
-available                      & \\
num-mobiles                     & number of mobiles which are tractable \\
                                & by a node B \\
num-calls                       & mobile network load for a node B \\
mobile-storage          & memory on S(IM)AT card \\
logical-channels                & number of logical channels available in the CN \\
cell-update                     & report UE location into RNC \\
handover                        & handover required to get a higher bit rate\\
active-set-up                   & update connection \\
ggsn-bitrate                    & capacity (kbit/s) from GGSN to PDN \\
max-no-pdp                      & max.~no.~of packet data protocols per mobile \\
max-no-apn                      & max.~access point names (APN) per mobile \\
\hline
\end{tabular}
\end{center}
\caption{Scheduling types of resources in the UMTS call set-up.}
\label{tab-resources}
\end{table}

Resources may be renewable or consumable: an example for a renewable
resource is the keyboard of the mobile. It can be used to input data
for several applications. Consumable resources are released after
action execution. The resources that are realized in the experiments
are summarized in Table~\ref{tab-resources} (see \citeR{3GPP} for a
complete list of resources for the UMTS call set-up).

%% Figure~\ref{fig-umts-discrete} depicts, that a mobile application can be
%% executed, when the radio bearer with the required QoS is established.

The PDDL representation of the planning domain is based on the eight
modules for the UMTS call set-up. There are eight operators
corresponding to these eight modules. Let us consider, as an example,
the BS action, that is, the final action that can be used to establish
the predicate \texttt{bs-ok}. It is defined as follows:

{\small
\begin{tabbing}
(\=:durative-action BS\\
\> :parameters\\
\> (?A-new - application ?M - mobile ?L - list ?MS1 ?MS2 - message ?a - agent)\\
\> :duration\\
\> (= ?duration (time-bs ?A-new))\\
\> :condition\\ 
\>     (and \=  (at start (initiated ?A-new ?M))\\ 
\> \>         (at start (aeei-ok ?A-new ?M ?L ?a))\\ 
\> \>         (at start (qos-params ?A-new ?L))\\ 
\> \>         (at start (message-trm ?M ?MS1))\\ 
\> \>         (at start (message-aeei ?A-new ?MS2)))\\ 
\> :effect \\ 
\>    (and (at end (iu-bearer ?A-new ?M ?L)) (at end (bs-ok ?A-new ?M ?L ?a)))))\\ 
\end{tabbing}
}

The action has as preconditions the successful execution of the module
AEEI during the call set-up, the satisfaction of the required QoS
class parameters (denoted as list $L$), and the transfered messages of
the set-up status to the application in the mobile and the PDN. The
resources are already allocated by the preceding modules. As effect
the bearer and the network connection for the mobile application are
set up.

The initiation of an application starts in the mobile with the TRM.
Afterwards, the CT in the AND is asked for a ready-for-service signal.
In the core of the call set-up is the radio access bearer procedure in
the CND. Let us consider the latter in more detail. As first step the
logical resources must be allocated (RRC), e.g., the required number of
channels must be provided by the logical level in the radio bearer and
later these logical resources are mapped to the physical channels. The
PDDL RRC action looks as follows:

{\small
\begin{tabbing}
(\=:durative-action RRC\\
\> :parameters\\
\> (?A-new - application ?M - mobile ?L - list ?a - agent)\\
\> :duration\\
\> (= ?duration (time-rrc ?A-new))\\
\> :condition\\
\>   (and \= (at start (ct-ok ?A-new ?M ?L))\\
\> \>         (at start (aeem-ok ?A-new ?M ?L ?a))\\
\> \>         (at start ($<$= \= (has-logical-channels) \\
\> \> \>                      (- (max-logical-channels) (app-channels ?A-new ?m))))\\
\> \>         (at start ($<$= (has-cell-update) (- (max-cell-update) 2)))\\
\> \>         (at start ($<$ (has-handover) (max-handover)))\\
\> \>         (at start ($<$ (has-active-set-up) (max-active-set-up))))\\
\> :effect\\ 
\>   (and \> (at start (increase (has-logical-channels) (app-channels ?A-new ?M)))\\
\> \>         (at end (decrease (has-logical-channels) (app-channels ?A-new ?M)))\\
\> \>         (at start (increase (has-cell-update) 2)) \\
\> \>         (at end (decrease (has-cell-update) 2)) \\
\> \>         (at start (increase (has-handover) 1))\\
\> \>         (at end (decrease (has-handover) 1))\\
\> \>         (at start (increase (has-active-set-up) 1))\\
\> \>         (at end (decrease (has-active-set-up) 1))\\
\> \>         (at end (rrc-ok ?A-new ?M ?L ?a))))\\
\end{tabbing}
}

If the requested QoS class is not available, then the fact
\texttt{rab-ok} is not true and a service agent must be sent to the
mobile in order to negotiate with the application or user for weaker
QoS requirements. In case of success the predicate \texttt{rab-ok} is
true and the connection to the PDN must be checked. Finally, the goal
predicate BS can be fulfilled if all resources are available.

\subsubsection{IPC-4 Domain Structure}
\label{umts:structure}

As used in IPC-4, the UMTS domain has six versions. The first three
are: \emph{temporal}, a domain version with no timing constraints,
\emph{temporal-timewindows}, a domain version with PDDL2.2 timed
initial facts, and \emph{temporal-timewindows-compiled}, a domain
version with a PDDL2.1 wrapper encoding for the timed initial
literals. The second domain version set \emph{flaw-temporal},
\emph{flaw-temporal-timewindows}, and
\emph{flaw-temporal-timewindows-compiled}, includes the following
``{\em flaw}'' action:

{\small
\begin{tabbing}
(\=:durative-action FLAW\\
\> parameters\\
\> (?A-new - application ?M - mobile ?L - list ?a - agent)\\
\> :duration (= ?duration 4)\\
\> :condition\\
\> (and \= (at start (initiated ?A-new ?M))\\
\> \> (at start (qos-params ?A-new ?L))\\
\> \> (at start (trm-ok ?A-new ?M ?L)))\\
\> :effect\\
\> (and \> (at end (rab-ok ?A-new ?M ?L ?a))\\
\> \> (at start (not (initiated ?A-new ?M)))))        
\end{tabbing}
}

This action offers a shortcut to the \texttt{rab-ok} predicate, but
can not be used in a real solution because it deletes the
\texttt{initiated} predicate. But the action {\em can} be used in
heuristic functions based on ignoring the negative effects. In that
sense, the action encodes a flaw that may disturb the heuristic
techniques used in modern planners. To determine that the action is
not useful, negative interactions have to be considered. The idea of
flaw is practically motivated in order to see how heuristic planners
react to it. In its standard form, the domain is not a big challenge
to such planners, as we have seen in Section~\ref{new}. All domain
versions have one formulation, namely \emph{strips-fluents-temporal},
where numerical fluents, but - except typing - no ADL constructs are
used. In all instances, the plan objective is to minimize
\emph{makespan}.

\begin{table}[t]
\centering
\begin{tabular}{|l|l||c|c|}
\hline  
version          & formulation &   $max\mbox{-}\#op$ &    $max\mbox{-}\#act$\\\hline\hline 
temporal              & STRIPS-TEMPORAL      &             8     &    (5120) 80  \\\hline 
temporal-tw           & STRIPS-TEMPORAL-TW   &             8     &    (5120) 80  \\\hline 
temporal-twc          & STRIPS-TEMPORAL      &            13     &    (5125) 85  \\\hline 
flaw-temporal         & STRIPS-TEMPORAL      &             9     &    (5310) 90  \\\hline 
flaw-temporal-tw      & STRIPS-TEMPORAL-TW   &             9     &    (5310) 90  \\\hline 
flaw-temporal-twc     & STRIPS-TEMPORAL      &            14     &    (5315) 95  \\\hline 
\end{tabular}
\caption{\label{umts:versionstable} Overview over the different 
  domain versions of UMTS. Abbreviations used: 
  ``temporal-tw'' for ``temporal-timewindows'', ``temporal-twc'' for 
  temporal-timewindows-compiled; $max\mbox{-}\#op$ is the maximum number of 
  (parameterized) PDDL operators for any instance, $max\mbox{-}\#act$ is the 
  maximum number of ground actions for any instance. Data in 
  parentheses are collected before FF's ``reachability'' pre-process 
  (see text).}
\end{table}

The domain versions and numbers of operators and ground actions are
overviewed in Table~\ref{umts:versionstable}. As with many of the
empirical data for UMTS that we have seen before, the data are quite
exceptional, and at the same time easy to interpret. First, similar to
what we have seen in Section~\ref{new:connectivity}, the data are
actually constant across all instances within each domain version,
which is once again due to the fact that the instances scale only in
their specification of what applications need actually be started.
Second, the numbers of operators and actions do not differ between the
versions with and without time windows; they increase somewhat,
through the additional artificial actions, if we compile timed initial
literals away (c.f.  Section~\ref{compilations}); they also increase
somewhat, of course, if we introduce the ``flaw'' action. Third, the
most striking observation is the {\em huge} effect of FF's
reachability pre-processor, building a relaxed planning graph for the
initial state and removing all actions that do not appear in that
graph. This is due to the technical subtleties of the encoding, where
the restrictions on feasible action instantiations are, partly,
implicit in the possible action sequences, rather than explicit in the
static predicates.

\subsubsection{IPC-4 Example Instances}
\label{umts:examples}

The UMTS call set-up domain has the following challenges for the
planning task \cite{Englert/Cremers/01}:
\begin{description}
\item[Real-time:] Can plans for the execution of mobile applications
  be generated in an appropriate time? Planning has to be done with a
  maximum duration that does not exceed the UMTS call set-up time.
\item[Completeness:] Is it possible to generate the plan, i.e.~does
  planning result in an (optimal) plan for the required applications
  that minimizes the waiting period until all applications are
  started?
\end{description}

The PDDL structure of the basic problem for the discrete UMTS call
set-up (DUCS) domain is the following:

{\small
\begin{tabbing}
(define (problem DUCS DOMAIN BASIC VERSION)\\
(:domain DUCS\_DOMAIN\_BASIC\_VERSION\\
(:objects \= MS1 MS2 - message\\
\>        A1 A2 A3 A4 A5 A6 A7 A8 A9 A10 - application\\
\>        M1 M2 M3 M4 M5 M6 M7 M8 M9 M10 - mobile\\
\>        L1 L2 L3 L4 L5 L6 L7 L8 L9 L10 - list\\
\>        ae - agent)\\
(:init \= (= (time-trm A1) 76)        (= (time-ct A1) 48) \\
\>        (= (time-am A1) 74)         (= (time-aeem A1) 66) \\
\>        (= (time-rrc A1) 202)       (= (time-rab A1) 67) \\
\>        (= (time-aeei A1) 36)       (= (time-bs A1) 28)\\
\> [...]       \\
\>        (location M1)          ;; types \\
\>        (authentification M1)\\
\> [...]\\
\>        (= (has-mobile-cpu) 0) ;; current status\\
\> [...] )\\
(:goal (and (bs-ok A1 M1 L1 ae) [...] )))\\
\end{tabbing}
}

First in this PDDL description come the objects for the applications
and the mobiles. Then come the durations of the modules depending on
the applications, e.g., the module TRM requires less time for a news
ticker than for a chess game, since the latter requires more terminal
resources than the ticker. The current status of the resources is
initialized. Finally, the goal is defined: the bearer establishment
for the execution start of the initiated mobile applications. The
total execution time should be minimized.

For IPC-4 the time windows are varied with small perturbations in
order to generate different instances. The perturbations are motivated
by the average execution times of the modules in a radio network
according to the load. Furthermore, the number of applications to be
set up is varied from 1 up to 10. The domains assume that the
applications run on one mobile terminal. However, they can also be
distributed to several mobile terminals. There are 50 different
instances per domain version.

\subsubsection{Future Work}
\label{umts:future}

The UMTS domain is not a big challenge for modern heuristic, i.e.
HSP/FF/LPG-style, planners because these planners are satisficing
(potentially return sub-optimal plans). The objective in UMTS is to
minimize the execution time, and if one ignores that objective then
the task trivializes. To the optimal planners, UMTS is a realistic
challenge. The domain is already relatively realistically modelled,
except for the left-out additional constraints on the (many) less
important resources. It remains to be seen if, when introducing all
these resources, planner (in particular optimal planner) performance
gets degraded. An option in this case may be to introduce explicit
language constructs for the different types (renewable and consumable)
of resources.

In the future the following two challenges shall be investigated.
%First, the inter-application constraint is examined. This constraint
%orders modules with same names of different applications. The effect
%is that renewable resources are available. As an example consider the
%constraint $(TRM_{app1}\ BEFORE\ TRM_{app2}) \vee (TRM_{app2}\ BEFORE\ 
%TRM_{app1})$. As a consequence the disjunction of constraints is
%introduced.
First, the negotiation of UMTS Quality of Service (QoS) parameters
could be considered. Assume a video application on a mobile terminal
is initiated, but the bearer resources are not sufficiently available.
Then the QoS has to be negotiated between the terminal and the bearer.
This leads to the planning of a negotiation during the plan execution
for the already initiated applications.

Second, the approach for the optimization of the UMTS call set-up can
be applied to the Wireless LAN registration. The challenge is to
transfer the QoS parameters, since the current Wireless LAN standard
(802.11b) does not contain QoS. This demerit can be solved by applying
an additional service level that addresses QoS.

\end{appendix}

\bibliography{biblio}
\bibliographystyle{theapa}

\end{document}